\documentclass[preprint,12pt,authoryear]{elsarticle}

\usepackage[utf8]{inputenc}
\usepackage{pgfplots}
\usepackage{graphicx}
\usepackage{subcaption}
\usepackage{pstricks}
\usepackage{amsmath}

\usepackage{multirow}
\usepackage{array}
\usepackage{tabularray}
\usepackage[section]{placeins}
\usepackage{booktabs}
\usepackage{longtable}
\usepackage{tabularx}
\usepackage{graphicx}
\usepackage{amssymb}
\usepackage{svg}
\usepackage{amsthm}
\usepackage{mathrsfs}
\usepackage{physics}

\allowdisplaybreaks

\usepackage{tcolorbox}
\usepackage{tikz}
\usepackage{pgfplots}
\usepackage{wrapfig}
\usepackage{lipsum}  
\usetikzlibrary{positioning}
\usepackage{tcolorbox}
\usepackage{amsfonts,amssymb,amsthm}
\usepackage{natbib}
\setcitestyle{numbers,square}
\usepackage{mathtools}
\usepackage{commath}
\usepackage{relsize}
\usepackage{bm}
\usepackage[font={small}]{caption}
\usepackage{comment,color,soul}
\usetikzlibrary{arrows,automata}
\usepackage{amsfonts}
\usepackage{url}
\usepackage{lipsum}
\usepackage[thinlines]{easytable}
\usepackage{nicefrac}
\usepackage{algorithm}
\usepackage{algpseudocode}
\usepackage{mysymbol}
\usepackage{float}
\usepackage{multirow}
\usepackage{colortbl}
\usepackage[title]{appendix}

\usepackage[mathscr]{euscript}
\usepackage{bbm}
\usepackage{amssymb}
\usepackage{svg}
\usepackage{amsthm}
\usepackage{mathrsfs}
\usepackage{physics}
\usepackage{blindtext}
 
\usepackage{xcolor}
\usepackage[utf8]{inputenc} 

\usepackage[T1]{fontenc}    
\usepackage{hyperref}       
\usepackage{url}            
\usepackage{booktabs}   
\usepackage{color}
\usepackage{amsfonts}       
\usepackage{nicefrac}       
\usepackage{microtype}      
\usepackage{lipsum}
\usepackage{fancyhdr}       
\usepackage{graphicx}       
\usepackage{enumitem}

\usepackage[utf8]{inputenc} 
\usepackage[T1]{fontenc}    
\usepackage{url}            
\usepackage{booktabs}       
\usepackage{amsfonts}       
\usepackage{nicefrac}       
\usepackage{microtype}      

\usepackage{algorithm}
\usepackage{longtable}

\usepackage{amsthm}
\usepackage{dsfont}

\usepackage{tabularray}

\newcommand{\E}{\mathbb{E}}

\newcommand{\sA}{{\mathscr A}}

\newcommand{\sD}{{\mathscr D}}

\newcommand{\sR}{{\mathscr R}}

\newcommand{\sT}{{\mathscr T}}

\newcommand{\sX}{{\mathscr X}}
\newcommand{\sY}{{\mathscr Y}}
\newcommand{\sZ}{{\mathscr Z}}

\newcommand{\DG}{\sD_G(S)}

\newcommand{\bz}{{\mathbf z}}

\let\E\undefined
\newcommand{\R}{\mathbb R}
\newcommand{\E}{\mathbb E}

\newcommand{\Saug}{\widetilde{S}}
\newcommand{\Daug}{\widetilde{\sD}}

\newcommand\Sdrop[1]{S^{\setminus #1}}
\newcommand\Schange[1]{S^{#1}}

\begin{document}

\begin{frontmatter}



\title{A Comprehensive Survey for Generative Data Augmentation}


\author{Yunhao Chen; Zihui Yan;  Yunjie Zhu}

\affiliation{
            country={China}}

\begin{abstract}
Generative data augmentation (GDA) has emerged as a promising technique to alleviate data scarcity in machine learning applications. This thesis presents a comprehensive survey and unified framework of the GDA landscape. We first provide an overview of GDA, discussing its motivation, taxonomy, and key distinctions from synthetic data generation. We then systematically analyze the critical aspects of GDA - selection of generative models, techniques to utilize them, data selection methodologies, validation approaches, and diverse applications. Our proposed unified framework categorizes the extensive GDA literature, revealing gaps such as the lack of universal benchmarks.  The thesis summarises promising research directions, including , effective data selection, theoretical development for large-scale models' application in GDA and establishing a benchmark for GDA. By laying a structured foundation, this thesis aims to nurture more cohesive development and accelerate progress in the vital arena of generative data augmentation.
\end{abstract}




\end{frontmatter}

\clearpage
\tableofcontents
\clearpage
\section{Introduction}
\label{Introduction}
\subsection{Introduction}
In the contemporary realm of machine learning, deep learning algorithms have emerged as powerful tools for a myriad of tasks, demonstrating unprecedented accuracy and capability.\cite{1_deep_learning,2_deep_learning,3_deep_learning} However, a critical cornerstone of their efficacy resides in their access to vast amounts of data\cite{4_high_data,5_high_data,6_high_data}. The process of gathering such expansive and pristine datasets in today's digital landscape, paradoxically, proves to be intricate and can be prohibitively costly. This intricate nature of data collection stems from multifaceted challenges, including but not limited to privacy concerns\cite{7_privacy,8_privacy}, diverse data sources, and the need for labouring labelling\cite{9_labelling_diffusion}. 

Consequently, the scientific community has pivoted towards data augmentation techniques as a pragmatic solution to counteract the dearth of available data. Data augmentation refers to a suite of techniques employed to artificially expand the volume and diversity of a dataset by introducing controlled modifications to its existing entries, without altering their inherent semantic interpretations.\cite{14_data_augmentation,15_data_augmentation} Traditional data augmentation\cite{10_data_augmentation} methodologies, while good, predominantly revolve around linear transformations or rudimentary non-linear modifications. Such methods, albeit beneficial, have exhibited limitations, particularly in their capacity to significantly enhance model performance, especially when the underlying data manifolds are complex. 

Recognizing these limitations, there has been a burgeoning interest among researchers to explore more sophisticated avenues of data augmentation. A notable direction in this pursuit is the integration of generative models for data augmentation, namely the generative data augmentation(GDA). The allure of generative models lies in their inherent ability to model intricate probability distributions of data\cite{11_data_distribution,12_data_distribution,13_data_distribution}, thus offering a more nuanced and expansive augmentation landscape. By tapping into this potential, generative data augmentation presents a promising horizon for improving deep learning model performance, especially in scenarios characterized by limited data availability.

This paper provides a comprehensive survey of Generative Data Augmentation (GDA). While the field has witnessed a multitude of approaches and methodologies, a structured and unified understanding of these techniques is often elusive. Many of the works apply GDA to a certain dataset with little novelty and contributions to the development of GDA. A unified framework can help to address this issue by offering a clear structure and categorization, making it easier for researchers to identify gaps and build upon existing methods.

Addressing this gap, we propose a unified framework that systematically categorizes the vast landscape of GDA. This framework serves as a roadmap, guiding readers through the multifaceted aspects of GDA: from the choice of generative models, to the techniques of utilizing them, the strategies for selecting high-quality synthetic data, methods to validate such data, and finally, the diverse applications where GDA proves instrumental.

\subsection{Organization of Survey}
\textbf{Section 2}: Preliminaries - This section offers foundational knowledge on Generative Data Augmentation (GDA). The core concepts, key terminologies, notations, and concepts used throughout the paper will be presented here.

\textbf{Section 3}: Choice of Generative Models - Here, we dive deep into the varied generative architectures at our disposal. From the conventional VAEs\cite{16_VAE,17_VAE,18_VAE,19_VAE,20_VAE,21_VAE} and GANs\cite{22_GAN,23_GAN,24_GAN,25_GAN,26_GAN,27_GAN,28_GAN,29_GAN,30_GAN} to the burgeoning GPT-based\cite{31_GPT,32_GPT,33_GPT,34_GPT,35_GPT,36_GPT,37_GPT,38_GPT,39_GPT,40_GPT} and diffusion-based innovations\cite{41_Diffusion,42_Diffusion,43_Diffusion,44_Diffusion,45_Diffusion,46_Diffusion,47_Diffusion}, we elucidate their mechanisms, pros, cons, and use-case scenarios.

\textbf{Section 4}: Utilization Techniques - This segment is dedicated to the effective harnessing of chosen generative models. We explore both latent space manipulations and prompt engineering, assessing their implications on data quality and relevance.

\textbf{Section 5}: Selection Strategies for Synthetic Data - Given the plethora of synthetic data generated, how do we curate the best? This section probes into the techniques, both well-established and emerging, that help in refining the quality of synthesized data.

\textbf{Section 6}: Validation of Synthetic Data - A crucial aspect of GDA is verifying the credibility of the generated samples. Here, we cover both the theoretical and empirical methods that ascertain the quality and relevance of synthetic datasets.

\textbf{Section 7}: Applications of GDA - GDA's versatility is front and center in this section. We chart out its transformative potential across diverse realms, from medical imaging in healthcare to applications in fields such as agriculture.

\textbf{Section 8}: Unified Framework for Generative Data Augmentation - Building on the insights from the preceding sections, we introduce and detail our proposed unified framework. This section elucidates the rationale behind the framework's structure and how it streamlines the process of GDA from model choice to application.

\textbf{Section 9}: Current Challenges and Future Directions - As we near the culmination of the survey, this section provides a reflective look at the ongoing challenges in GDA and envisions potential breakthroughs. It serves as a guide for researchers looking to further the boundaries of GDA.

\begin{figure}
    \centering
    \centerline{\includegraphics[width=1.5\columnwidth]{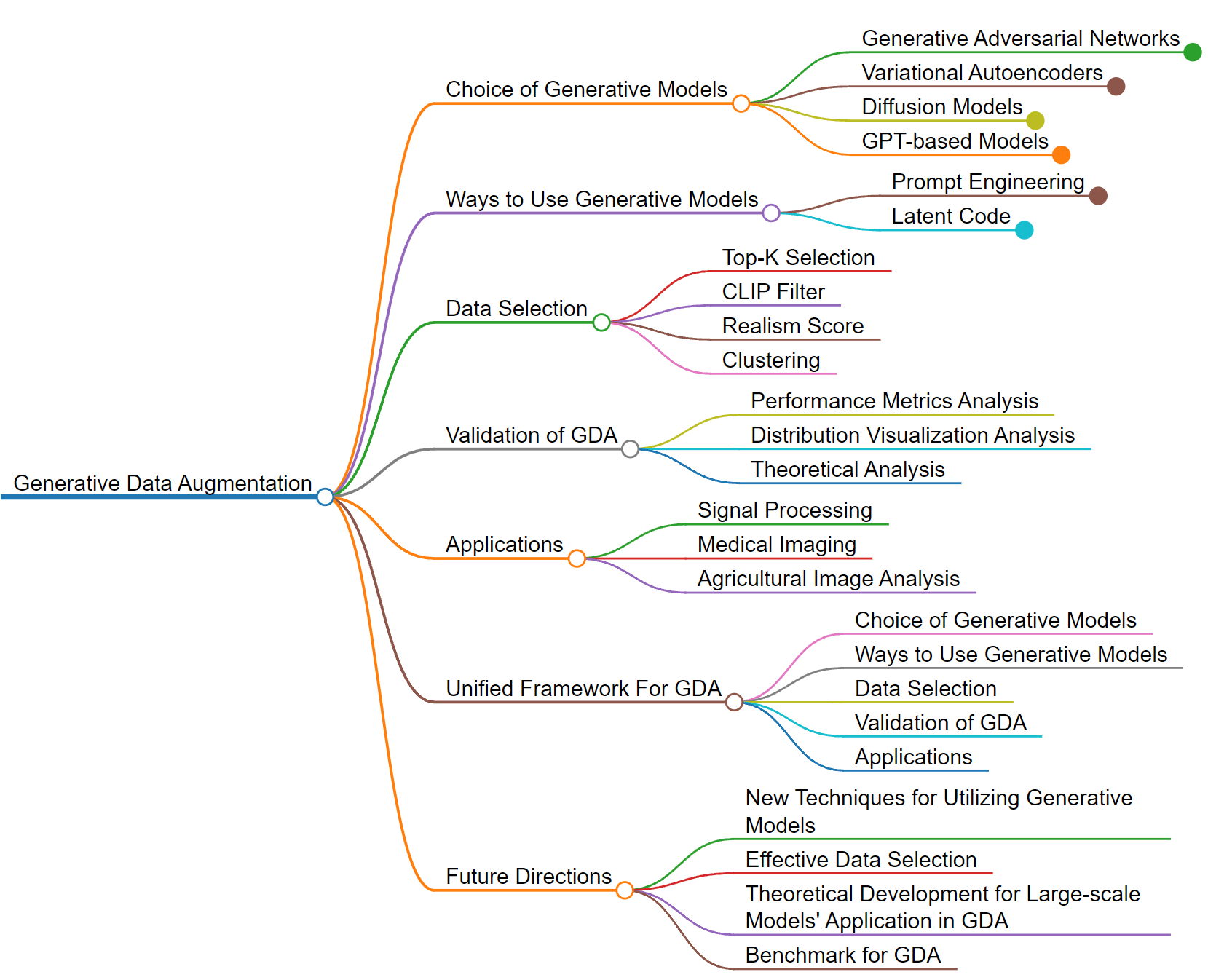}}
    \caption{Overview of the Generative Data Augmentation Survey: A structured breakdown detailing the primary components, methodologies, applications, and future prospects of leveraging generative models for data augmentation.
}
    \label{fig:enter-label}
\end{figure}

The unified framework which can also be viewed as the flow of this paper can be visualized in Fig. 1.

The main contributions of this survey are summarized as follows:

\begin{enumerate}
    \item \textbf{Extensive and Latest  Compilation:} Drawing from over 230 seminal works in the span of the last three years, this survey presents one of the most comprehensive reviews on Generative Data Augmentation (GDA), effectively capturing the rapid advancements in the domain.
    
    \item \textbf{Unified Framework Proposal:} We introduce a structured and cohesive GDA framework encompassing model selection, utilization techniques, synthetic data selection, validation, and applications. This offers researchers and practitioners a systematic guideline for improving GDA and implementing GDA in varied contexts.
    
    \item \textbf{Deep Dive into Selection \& Validation:}  Our survey delves deeply into the nuances of synthetic data selection and validation, which are given little attention on in the previous research works, emphasizing their importance in the effective deployment of GDA techniques.
    
    \item \textbf{Future Roadmap:} Benefitting from the extensive literature review, we discern and discuss the existing challenges and potential breakthrough avenues, providing a visionary roadmap for future research in GDA.
\end{enumerate}

\section{Preliminaries}

\subsection{Notations}

Define the input and label spaces as $\mathscr{X} \subseteq \R^d$ and $\mathscr{Y} \subseteq \R$, respectively. The population distribution over $\mathscr{Z} = \mathscr{X} \times \mathscr{Y}$ is represented by $\mathscr{D}$. For a random variable $X$, its $L_p$ norm is $\Vert X\Vert_p = (\E|X|^p)^{\frac{1}{p}}$.

Given a set $S = \{\bz_1, \bz_2, \dots, \bz_m\}$, $\Sdrop{i}$ removes the $i$-th data point from $S$, while $\Schange{i}$ replaces it with $\bz_i'$. For $[m] = \{1,2,\dots,m\}$, any subset $V \subseteq [n]$ results in $S_V = \{\bz_i: i \in V\}$. A function $f = f(S)$ has a conditional $L_p$ norm, $\norm{f}_p(S_V) = (\E[\ \norm{f}^p \mid S_V])^{\frac{1}{p}}$. The total variation distance and KL divergence are denoted as $d_{\mathrm{TV}}$ and $d_{\mathrm{KL}}$.

The set of all measurable functions from $\sX$ to $\sY$ is $(\sY)^{\sX}$. For a learning algorithm $\sA$, the hypothesis it formulates on dataset $S$ is $\sA(S) \in (\sY)^{\sX}$. The true error of a hypothesis $\sA(S)$, given the loss function $\ell: (\sY)^{\sX} \times \sZ \rightarrow \R_+$, is $\sR_{\sD}(\sA(S)) = \E_{\bz \sim \sD} [\ell(\sA(S), \bz)]$. Its empirical counterpart is $\widehat{\sR}_{S}(\sA(S)) = \frac{1}{m} \sum_{i=1}^m \ell(\sA(S), \bz_i)$.\cite{60_Theory}

\subsection{Data Augmentation}
Data augmentation\cite{49_TDA,50_TDA,51_TDA,52_TDA,53_TDA,54_TDA,55_TDA,56_TDA,57_TDA,58_TDA,59_TDA} stands as a cornerstone technique in machine learning and, more specifically, in deep learning. At its core, data augmentation pertains to artificially increasing a dataset's size and diversity.

This methodology serves a dual purpose:
\begin{itemize}
    \item Enhancing the volume of training data: By introducing controlled variations, data augmentation effectively bolsters the amount of available training data, thereby attenuating overfitting.
    \item Improving model generalization: The induced variability ensures that the model is exposed to a broader spectrum of data scenarios, fostering a more generalized understanding.
\end{itemize}

Traditional data augmentation techniques revolve around simple transformations, such as rotations, scaling, and cropping, especially in image data. However, as datasets grow in complexity, there's a pressing need for more sophisticated augmentation strategies.

\subsection{Generative Data Augmentation}
Generative Data Augmentation emerges as an advanced offshoot of the traditional data augmentation paradigm. The key innovation here is the utilization of generative models to fabricate synthetic data samples.

Generative Data Augmentation involves the use of generative models, trained from scratch on the target dataset or pre-trained on large-scale datasets, to fabricate synthetic data samples. 

The mathematical representation of the process is presented below:

 Let $S$ be a training set with $m_S$ i.i.d. examples from $\sD$. After training a generative model $G$, its distribution is denoted by $\DG$. They\cite{60_Theory} assume the randomness during model training is consistent throughout the paper. We express the model's expected distribution with respect to $S$ as $\sD_G = \E_S[\DG]$. 

Using the trained $G$, a dataset $S_G$ is derived with $m_G$ i.i.d. samples from $\DG$. If GDA is employed, it's common to have $m_G = \Omega(m_S)$. The augmented set, $\Saug = S \cup S_G$, contains a total of $m_T$ data points. The post-augmentation mixed distribution is $\Daug(S) = \frac{m_S}{m_T} \sD + \frac{m_G}{m_T} \DG$. Given this, a hypothesis $\sA(\Saug)$ can be formulated on the augmented dataset $\widetilde{S}$. 

\subsection{Generative Data and Synthetic Data}

Synthetic data\cite{61_synthetic_data,62_synthetic_data,63_synthetic_data,64_synthetic_data,65_synthetic_data,66_synthetic_data,67_synthetic_data,68_synthetic_data} refers to any data not generated by actual events but artificially created for various purposes, including testing, training, or simulation. Its essence lies in the replication of some properties of the original dataset while not deriving from actual events.

On the other hand, generative data is a subset of synthetic data. The distinctive characteristic of generative data is that it is produced using generative models. These models, such as Generative Adversarial Networks or Variational Autoencoders, are trained to understand and capture the underlying data distributions from real datasets. Once trained, they can generate new data samples that are statistically consistent with the original data if it is trained on sufficient data. 

For example, the study by \cite{69_synthetic_data_aerial} leveraged the game Grand Theft Auto V \cite{72_grad_theft_auto} as a platform to produce synthetic data. Similarly, \cite{70_synthetic_data_game} utilized the Unreal game engine for this purpose. On a different note, \cite{71_synthetic_data_physics} employed a physics-based system to generate their synthetic dataset. However, none of these are generative data.

To conclude, while all generative data falls under the umbrella of synthetic data, the distinction lies in the method of creation. Generative data specifically leverages the prowess of generative models to create new data samples that mirror the characteristics of real-world data. 

\begin{table}[]
\begin{tabular}{l|ll}
Name            & \begin{tabular}[c]{@{}l@{}}Differences with 
 another\end{tabular}                                       & \begin{tabular}[c]{@{}l@{}}Example ways to \\ produce it\end{tabular}              \\ \hline
Synthetic Data  & \begin{tabular}[c]{@{}l@{}}Data artificially created \\ for testing, training, or simulation\end{tabular} & \begin{tabular}[c]{@{}l@{}}Game-based system; \\ Physics-based system\end{tabular} \\ \hline
Generative Data & \begin{tabular}[c]{@{}l@{}}Subset of synthetic data;\\  producede using generative models\end{tabular}    & \begin{tabular}[c]{@{}l@{}}GAN; VAE; \\ Diffusion-based models\end{tabular}       
\end{tabular}
\caption{Comparison between Synthetic Data and Generative Data}
\end{table}

\section{Generative Models}

The landscape of generative modelling has seen a seismic shift in recent years, with a multitude of architectures pushing the boundaries of what's possible. As we delve into this realm, it becomes paramount to understand the intricacies and nuances of each model to harness their potential for specific applications effectively. This section provides a comprehensive exploration into the vast array of generative models available today. Figure 2 provides an overview of VAE, GAN and diffusion-based models. 

\begin{figure}
    \centering
    \centerline{\includegraphics[width=0.7\columnwidth]{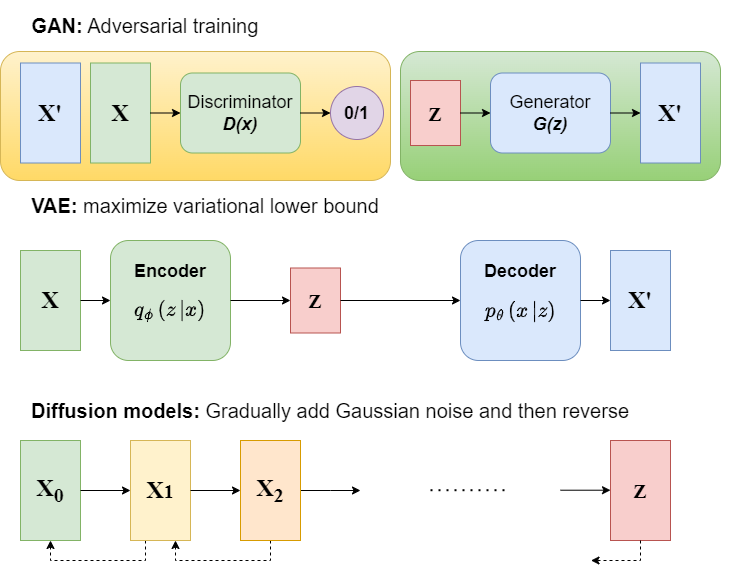}}
    \caption{An overview of GAN,VAE and Diffusion models.}
    \label{Fig 2}
\end{figure}

\subsection{VAE}
Variational inference is a powerful technique in the realm of Bayesian statistics. At its core, it aims to approximate the true posterior distribution \(p(z|x)\) by employing a more tractable or simpler distribution, denoted as \(q(z|x)\). This approximation strategy is motivated by the inherent complexity and computational challenges associated with directly calculating the posterior distribution in many Bayesian models.

The fundamental objective of the variational inference method is to minimize the Kullback--Leibler (KL) divergence between the authentic posterior \(p_\theta(z|x)\) and the approximated variational distribution \(q_\phi(z|x)\). The symbols \(\theta\) and \(\phi\) stand for the parameters governing the posterior and the variational distributions, respectively. It is imperative to note that the KL divergence serves as a measure of the difference between two probability distributions and, in this context, provides a quantification of how close our approximation \(q_\phi(z|x)\) is to the true posterior \(p_\theta(z|x)\) \cite{75_KL}.

The concept of utilizing the Kullback-Leibler divergence in the domain of variational inference and many other statistical contexts is well-established and thoroughly explored in the literature \cite{74}. 

The associated loss function, which encapsulates the difference between the true posterior and its approximation and guides the optimization procedure in variational inference, can be mathematically expressed as:

\begin{equation} 
\min_{\theta,\phi} D_{KL}(q_\phi(z|x) || p_\theta(z|x)) = \min_{\theta,\phi} \mathbb{E}_{z \sim q_\phi}[\log \frac{q_\phi(z|x)}{p_\theta(z|x)}]
\end{equation} 

$
\begin{array}{l}

D_{\mathrm{KL}}\left(q_{\phi}(z \mid x) \| p_{\theta}(z \mid x)\right) \\
=\int q_{\phi}(z \mid x) \log \frac{q_{\phi}(z \mid x)}{p_{\theta}(z \mid x)} d z \\
=\log p_{\theta}(x)+\int q_{\phi}(z \mid x) \log \frac{q_{\phi}(z \mid x)}{p_{\theta}(x \mid z) p_{\theta}(z)} d z \\
=\log p_{\theta}(x)+\mathbb{E}_{z \sim q_{\phi}(z \mid x)}\left[\log \frac{q_{\phi}(z \mid x)}{p_{\theta}(z)}-\log p_{\theta}(x \mid z)\right] \\
=\log p_{\theta}(x)+D_{\mathrm{KL}}\left(q_{\phi}(z \mid x) \| p_{\theta}(z)\right)-\mathbb{E}_{z \sim q_{\phi}(z \mid x)} \log p_{\theta}(x \mid z) \\

\end{array}
$

We can come to the following:

\begin{equation}
    \log p_{\theta}(x)-D_{\mathrm{KL}}\left(q_{\phi}(z \mid x) \| p_{\theta}(z \mid x)\right)
    \\
    =\mathbb{E}_{z \sim q_{\phi}(z \mid x)} \log p_{\theta}(x \mid z)-D_{\mathrm{KL}}\left(q_{\phi}(z \mid x) \| p_{\theta}(z)\right)
\end{equation}

Upon further refinement and using Jensen's inequality, the equation can be transformed into:

\begin{align}
    \log p_\theta(x) &= -\mathbb{E}_{z\sim q_\phi}[\log q_\phi(z|x)] + \mathbb{E}_{z \sim q_\phi}[\log p_\theta(z, x)]
    + D_{KL}(q_\phi(z|x) || p_\theta(z|x)) \\
    \log p_\theta(x) &\geq  -\mathbb{E}_{z\sim q_\phi}[\log q_\phi(z|x)] + \mathbb{E}_{z\sim q_\phi}[\log p_\theta(z, x)] \\
    &\geq \mathbb{E}_{z\sim q_\phi}[\log p_\theta(x|z)] - \mathbb{E}_{z\sim q_\phi}[\log \frac{q_\phi(z|x)}{p(z)}] = \text{ELBO}
\end{align}

In the presented equations, the term \(\log p_\theta(x)\) represents the marginal log-likelihood of the data, denoted as \(x\). This measure provides insight into the probability of observing the data under the current model parameters. On the other hand, \(p(z)\) symbolizes the prior distribution of the latent variable \(z\). In numerous applications, especially in the context of Variational Autoencoders, this distribution is often assumed to follow a Gaussian distribution.

One of the key terminologies within the realm of variational inference is the Evidence Lower Bound, commonly referred to as the ELBO. This term serves as a lower bound on the marginal log-likelihood, and maximizing the ELBO is tantamount to minimizing the Kullback--Leibler (KL) divergence between the true posterior and the variational distribution \(q_\phi(z|x)\), with the model parameters \(\theta\) being held constant during the process.

Diving deeper into the structure of ELBO, it can be compartmentalized into two pivotal components: the reconstruction term and the regularization term. The reconstruction component is primarily concerned with quantifying the difference between the original data and its reconstruction. Typically, the binary cross-entropy loss is employed to gauge this discrepancy. Conversely, the regularization term plays an instrumental role in ensuring that the latent variables conform to a predefined distribution, often Gaussian in nature. This alignment is achieved by computing the KL divergence between the latent variable distribution and the desired distribution.

Taken together, these two constituents form the backbone of the ELBO loss function, which is of paramount importance when training the Variational Autoencoder model. The architecture of a VAE is bifurcated into an encoder, represented by \(q_\phi(z|x)\), and a decoder, represented by \(p_\theta(x|z)\). The encoder, typically designed as a neural network, is tasked with the responsibility of mapping data \(x\) to the latent space, producing the latent variable \(z\). In contrast, the decoder, also a neural construct, undertakes the inverse procedure, translating the latent variables back to the data space.

The training regimen for a VAE revolves around minimizing a loss function that amalgamates the reconstruction and regularization losses:
\begin{equation}
    \mathcal{L}_{rec} = \mathbb{E}_{q_\phi(z|x)}[\log p_\theta(x|z)], \;\;
    \mathcal{L}_{reg} = \mathbb{E}_{q_\phi(z|x)}[\log \frac{q_\phi(z|x)}{p(z)}]
\end{equation}

Upon the culmination of the training phase, the VAE is equipped to generate novel data samples. This is realized by first drawing a random latent vector \(z\) from the prior distribution \(q_\phi\), which is subsequently passed through the decoder. In its essence, the decoder is a manifestation of the generative model, capable of synthesizing data samples that mirror the distribution of the training data.

\subsection{GAN} 
The Generative Adversarial Network, as introduced by \citep{23_GAN}, ushered in a paradigm shift in the realm of generative models. Unlike conventional approaches, GANs are designed to generate samples directly from the desired data distribution, bypassing the need to explicitly model the underlying probability density function. This elegant solution revolves around a tandem of neural networks, colloquially referred to as the generator (G) and the discriminator (D).

Delving into the mechanics of the GAN, the generator is initially fed an input \(z\), which constitutes random noise sampled from a predetermined prior distribution \(p(z)\). For the sake of simplicity and computational ease, this distribution often takes the form of a Gaussian or uniform distribution. Once this noise passes through the generator, the resultant output \(x_g\) should ideally resemble a real data sample, denoted \(x_r\), drawn from the genuine data distribution \(p_r(x)\). The intricate, non-linear mapping function facilitated by the generator, parameterized by \(\theta_g\), can be mathematically expressed as \(x_g = G(z;\theta_g)\).

In parallel, the discriminator is assigned the role of discerning whether a given sample is genuine or a fabrication from the generator. For any input sample, either real or crafted by the generator, the discriminator emits a singular output \(y_1\), representing the probability that the input stems from a genuine source. The mathematical representation of the mapping learned by the discriminator, parameterized by \(\theta_d\), can be depicted as \(y_1 = D(x;\theta_d)\). As the training progresses, the generator aims to fashion a distribution \(p_g(x)\) that approximates the true data distribution \(p_r(x)\). For a visual perspective on GAN's intricate structure, one may refer to the top section of Figure 2.

The intertwined objectives of the discriminator and generator, which guide the training process, are encapsulated in the following mathematical formulations:
\begin{equation}
\begin{split}
\mathcal{L}^{GAN}_{D} &= \max_{D}  \mathbb{E}_{x_r\sim p_r{(x)}}\big[\log D(x_r) \big] 
 + \mathbb{E}_{x_g\sim p_g(x)}\big[\log(1-D(x_g))\big], \\
\mathcal{L}^{GAN}_{G} &= \min_{G}   \mathbb{E}_{x_g\sim p_g(x)}\big[\log(1-D(x_g))\big].
\end{split}
\end{equation}

At its heart, the discriminator \(D\) functions as a binary classifier, employing a maximum log-likelihood objective, as elucidated by \cite{76_GAN_review}. Theoretically, if the discriminator reaches its optimal performance before the subsequent updates of the generator, then the minimization of \(\mathcal{L}^{GAN}_{G}\) converges to the minimization of the Jensen--Shannon (JS) divergence between \(p_r(x)\) and \(p_g(x)\). The epitome of successful GAN training materializes when the samples, represented by \(x_g\), approximate the true data distribution \(p_r(x)\) to a high degree.

\subsection{Diffusion Models}
Diffusion models have emerged as an incredibly potent subclass within the realm of probabilistic generative models. Their primary objective is to learn and mimic intricate data distributions. The approach that makes diffusion models particularly captivating is their two-pronged approach, delineated as the forward diffusion process and the reverse diffusion process.

\subsubsection{Variational Perspective}
Venturing into the variational domain, we are introduced to models that harness the power of variational inference to approximate their target distribution. This is generally accomplished by narrowing down the Kullback-Leibler divergence between the approximate and the target distributions. The Denoising Diffusion Probabilistic Models (DDPMs), which have been discussed in literature \cite{46_Diffusion,42_Diffusion}, are a testament to this approach. They utilize variational inference to derive the parameters of the underlying diffusion process.

\textbf{Forward Process.} 
At its core, DDPM characterizes the forward diffusion process as a Markov Chain. This chain, through successive iterations, incorporates Gaussian noise, resulting in a series of increasingly noisy samples. Let's consider the initial data distribution \(q\left(x_0\right)\) to be untarnished. When we procure a data sample \(x_0 \sim q\left(x_0\right)\), the subsequent forward noising process \(p\) establishes latents from \(x_1\) to \(x_T\) by incorporating Gaussian noise at every time point \(t\):

\begin{equation}
	q\left(x_t \mid x_{t-1}\right)=\mathcal{N}\left(x_t ; \sqrt{1-\beta_t} \cdot x_{t-1}, \beta_t \cdot \mathbf{I}\right), \forall t \in\{1, \ldots, T\},
\end{equation}

Here, \(T\) depicts the total diffusion steps, and the variables \(\beta_1, \ldots, \beta_T \in[0,1)\) embody the variance schedule spread across these steps. The identity matrix is represented as \(\mathbf{I}\), and \(\mathcal{N}(x ; \mu, \sigma)\) symbolizes the normal distribution with mean \(\mu\) and covariance \(\sigma\). For ease of interpretation and calculation, we introduce \(\alpha_t=1-\beta_t\) and \(\bar{\alpha}_t=\prod_{s=0}^t \alpha_s\). This allows us to sample directly at any noise step, keeping the input \(x_0\) as a condition:

\begin{equation}
	q\left(\mathbf{x}_t \mid \mathbf{x}_0\right)=N\left(\mathbf{x}_t ; \sqrt{\bar{\alpha}_t} \mathbf{x}_0,\left(1-\bar{\alpha}_t\right) \mathbf{I}\right)
\end{equation}

\begin{equation}
	\mathbf{x}_t=\sqrt{\bar{\alpha}_t} \mathbf{x}_0+\sqrt{1-\overline{\alpha_l}} \epsilon .
\end{equation}

\textbf{Reverse Process.} Employing the definitions constructed earlier, the reverse process can be formulated to extract a sample from \(q\left(x_0\right)\). By parameterizing this process and initializing at \(p\left(\mathbf{x}_T\right)=\mathcal{N}\left(\mathbf{x}_T ; \mathbf{0}, \mathbf{I}\right)\), we arrive at:

\begin{equation}
	p_\theta\left(\mathbf{x}_{0: T}\right)=p\left(\mathbf{x}_T\right) \prod_{t=1}^T p_\theta\left(\mathbf{x}_{t-1} \mid \mathbf{x}_t\right)
\end{equation}
\begin{equation}
	p_\theta\left(\mathbf{x}_{t-1} \mid \mathbf{x}_t\right)=\mathcal{N}\left(\mathbf{x}_{t-1} ; \mu_\theta\left(\mathbf{x}_t, t\right), \Sigma_\theta\left(\mathbf{x}_t, t\right)\right) .
\end{equation}

To mold this model such that \(p\left(x_0\right)\) emulates the true data distribution \(q\left(x_0\right)\), a variational bound on the negative log-likelihood is optimized:

\begin{equation}
	\label{eq}
	\begin{aligned}
		\mathbb{E}\left[-\log p_\theta\left(\mathbf{x}_0\right)\right] & \leq \mathbb{B}_q\left[-\log \frac{p_\theta\left(\mathbf{x}_{0: T}\right)}{q\left(\mathbf{x}_{1: T} \mid \mathbf{x}_0\right)}\right] \\
		&=\mathbb{E}_q\left[-\log p\left(\mathbf{x}_T\right)-\sum_{t \geq 1} \log \frac{p_\theta\left(\mathbf{x}_{t-1} \mid \mathbf{x}_t\right)}{q\left(\mathbf{x}_t \mid \mathbf{x}_{t-1}\right)}\right] \\
		&=-L_{\mathrm{VL} . \mathrm{B}} .
	\end{aligned}
\end{equation}

Interestingly, the study by \cite{42_Diffusion} preferred not to directly infuse the parameterization of \(\mu_\theta\left(x_t, t\right)\) using a neural network. Instead, a model, \(\epsilon_\theta\left(x_t, t\right)\), was trained to forecast the \(\epsilon\) variable. With this reparameterization of \ref{eq}, a streamlined objective was proposed:

\begin{equation}
	L_{\text {simple }}=E_{t, x_0, \epsilon}\left[\left\|\epsilon-\epsilon_\theta\left(x_t, t\right)\right\|^2\right] ,
\end{equation}
Intriguingly, this methodology draws parallels between the loss in \ref{eq} and the generative score networks discussed in the work of \cite{77_diffusion}.

\subsubsection{Score Perspective}
Score Perspective models lean on a maximum likelihood-based estimation technique, making use of the score function of the log-likelihood of the data to gauge the parameters of the diffusion process. There are key subcategories within this realm: Noise-conditioned Score Networks (NCSNs) \cite{77_diffusion} and Stochastic Differential Equations (SDEs) \cite{78_score_based}. NCSNs focus on estimating the derivative of the log density function of the perturbed data distribution at different noise levels, while SDEs are a generalization of previous approaches and encompass both DDPMs and NCSNs characteristics. \cite{84_diffusion_survey}

\textbf{Noise Conditioned Score Networks (NCSNs)}

The score function for a data distribution $p(x)$ is $\nabla_x \log p(x)$. A neural network, $\mathbf{s}_{\boldsymbol{\theta}}$, parameterized by $\boldsymbol{\theta}$, can be trained via score matching to approximate this function, i.e., $s_\theta(x) \approx \nabla_x \log p(x)$. The training objective is:

\begin{equation}
	\label{ncsn}
	\mathbb{E}_{x \sim p(x)}\left\|s_\theta(x)-\nabla_x \log p(x)\right\|_2^2 .
\end{equation}

Due to computational challenges, score matching struggles with deep networks and high-dimensional data. \cite{77_diffusion} leverages denoising and sliced score matching \cite{80_score_matching,79_sliced_score}. They note that score functions can be inaccurate in low-density regions, supporting the manifold hypothesis. This is addressed by perturbing data with Gaussian noise at varying scales, simplifying score-based generative modelling. The noise-conditioned score network (NCSN) is proposed to estimate scores across noise levels. For Gaussian noise scales $\sigma_1<\sigma_2<\cdots<\sigma_T$, the objective is:

\begin{equation}
	\frac{1}{T} \sum_{t=1}^T \lambda(\sigma_t) \mathbb{E}_{p(x)} \mathbb{E}_{x_t \sim p_{\sigma_t}(x_t \mid x)}\|s_\theta(x_t, \sigma_t)+\frac{x_t-x}{\sigma_t}\|_2^2,
\end{equation}

Langevin dynamics \cite{81_dynamics,82_dynamics} is an MCMC approach using only score functions, iteratively:

\begin{equation}
	\label{Langevin-eq}
	x_i=x_{i-1}+\frac{\gamma}{2} \nabla_x \log p(x)+\sqrt{\gamma} \cdot \omega_i,
\end{equation}

with $\omega_i \sim \mathcal{N}(0, \mathbf{I})$ and $i \in\{1, \ldots, N\}$. In the limit, the samples converge to $p(\mathbf{x})$. \cite{77_diffusion} modifies this into the annealed Langevin dynamics algorithm, where noise scale $\sigma_i$ decreases over time to improve score matching \cite{83_dynamics}.

\textbf{Stochastic Differential Equations (SDEs)}
\label{sde-eq}

score-based generative models (SGMs) \cite{78_score_based} map the data distribution $q(x_0)$ to noise. Extending noise scales to infinity allows viewing traditional probabilistic models as SGM discretizations. Many stochastic processes, e.g., diffusion, are solutions to stochastic differential equations (SDEs) like:
\begin{equation}
	\label{sde-base}
	\mathrm{d} \mathbf{x}=\mathbf{f}(\mathbf{x}, t) \mathrm{d} t+g(t) \mathrm{d} \mathbf{w} ,
\end{equation}
with $\mathbf{f}(., t)$ as the drift coefficient, $g(t)$ as the diffusion coefficient, and $\mathbf{w}$ as standard Brownian motion. For the given forward SDE, there exists a reverse time SDE running backward where, by starting out with a sample from $p_T$ and reversing this diffusion SDE, we can obtain samples from our data distribution $p_0$. For uncorrupted sample $\mathbf{x}_0$ and perturbed sample $\mathbf{x}_T$, a reverse-time SDE exists:
\begin{equation}
	\label{sdebackward}
	d \mathbf{x}=\left[\mathbf{f}(\mathbf{x}, t)-g^2(t) {\nabla_x \log p_t(x)}\right] d t+g(t) d \bar{\mathbf{w}} ,
\end{equation}
where $d t$ is the infinitesimal negative time step, and $\bar{w}$ is the Brownian motion running backward. In order to numerically solve the reverse-time SDE, one can train a neural network to approximate the actual score function via score matching \cite{77_diffusion,78_score_based} to estimate $\boldsymbol{s}_\theta(\boldsymbol{x}, t) \simeq \nabla_{\boldsymbol{x}} \log p_t(\boldsymbol{x})$.
This score model is trained using the following objective:

\begin{footnotesize}
\begin{equation}
	\mathcal{L}(\theta)=\mathbb{E}_{\mathbf{x}(t) \sim p(\mathbf{x}(t) \mid \mathbf{x}(0)), \mathbf{x}(0) \sim p_{\text {data }}}\left[\frac{\lambda(t)}{2}\left\|s_\theta(\mathbf{x}(t), t)-\nabla_{\mathbf{x}(t)} \log p_t(\mathbf{x}(t) \mid \mathbf{x}(0))\right\|_2^2\right] ,
\end{equation}
\end{footnotesize}where $\lambda$ is a weighting function and $t \sim \mathcal{U}([0, T])$. Numerical methods applied to \ref{sdebackward} can sample from SDEs.

\subsection{Generative Pre-trained Transformer}

\subsubsection{Transformer}

\begin{figure}
    \centering
    \centerline{\includegraphics[width=0.43\columnwidth]{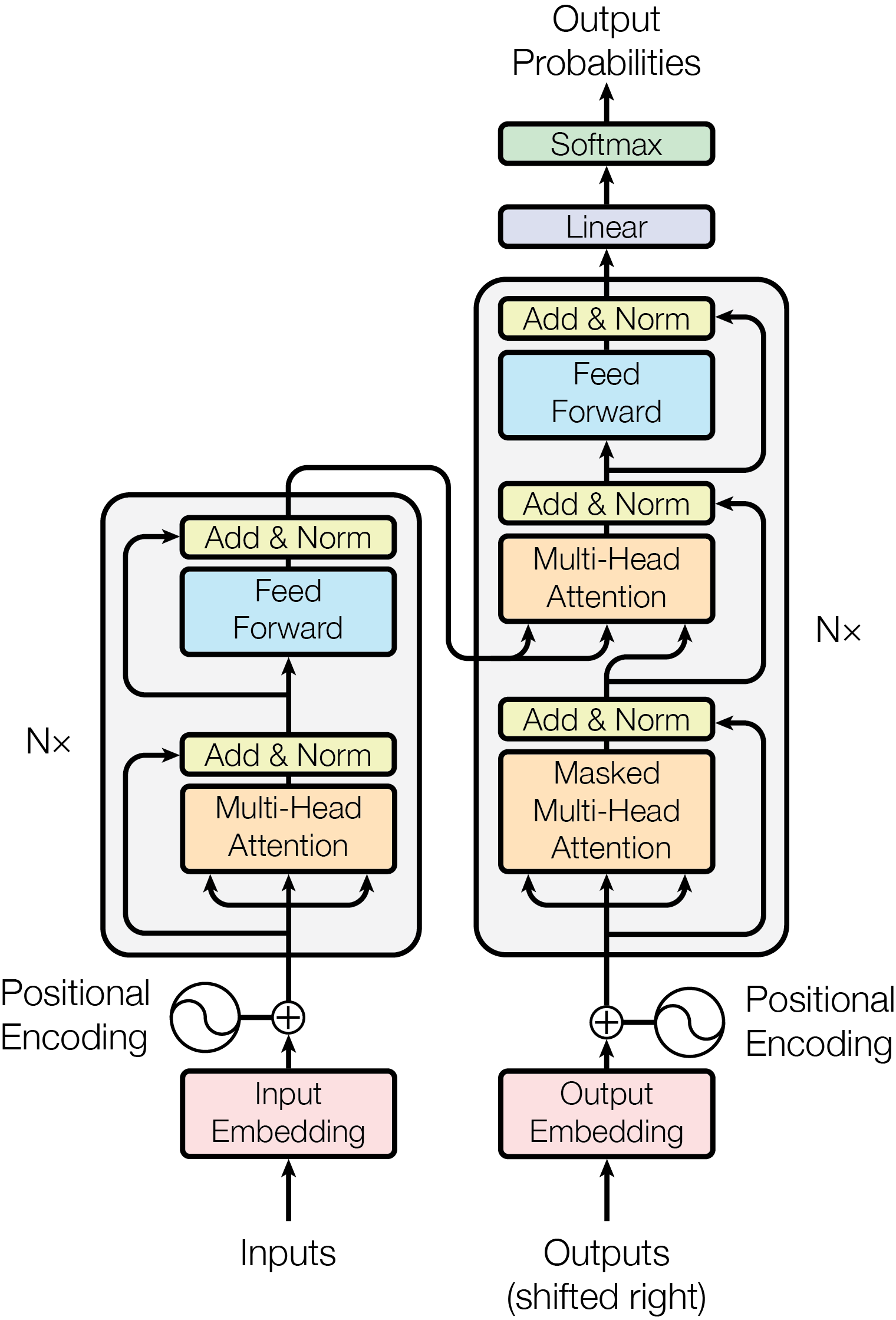}}
    \caption{Transformer Structure\cite{85_transformer} }
    \label{fig:transformer}
\end{figure}

In the realm of neural network architectures for sequence processing, the Seq2seq with attention~\cite{86_seq_atten} was a significant development. However, at that time, a novel model emerged that took a different approach. Instead of relying on the conventional seq-2seq foundation, this new model is built on the premise that attention mechanisms alone are sufficient for achieving state-of-the-art performance~\cite{85_transformer}. This specialized attention mechanism is referred to as self-attention. This innovative architecture has been named the \textit{Transformer}~\cite{85_transformer}, as depicted in Figure \ref{fig:transformer}.

Delving into the structure of a Transformer, it is primarily composed of an encoder and a decoder. Both components benefit from contemporary architectural advancements, such as residual connections \cite{87_resnet} and layer normalization techniques. Within the Transformer's framework, two primary modules stand out, excluding the Add \& Norm module. They are the multi-head attention and the feed-forward neural network( MLP ).

The attention mechanism within the Transformer is distinctive. It employs a design that incorporates multiple 'heads', and its self-attention mechanism utilizes the scaled dot-product approach, mathematically defined as:

\begin{equation}
Attention(Q, K, V) = softmax(\frac{QK^T}{\sqrt{d_k}.})V
\end{equation}

A critical comparison to make is with Recurrent Neural Networks (RNNs)\cite{89_RNN,90_RNN,91_RNN}. While RNNs inherently capture positional information through their sequential processing of data, Transformers approach the problem differently. They construct global dependencies, enabling them to model complex relationships within the data. However, a drawback is that they can overlook positional bias in the data since they process all input simultaneously. This is where positional encoding comes into play, ensuring the model recognizes the positional context of the input data.\cite{92_RNN}

There are two primary methods for positional encoding. One uses fixed position coding, which leverages sinusoids and cosines at varying frequencies to represent position. In contrast, the learnable position encoding technique uses a set of parameters that can be adjusted during training to represent positional information best.\cite{93_transformer}

Given its capabilities and versatility, the Transformer architecture has cemented its place as a cornerstone in the field of Natural Language Processing, becoming a standard choice for many tasks in the domain.\cite{88_AIGC_survey} 

\subsubsection{GPT-series models}

Autoregressive language pretraining methods have emerged as powerful techniques for few-shot or zero-shot text generation. Among these, the GPT (Generative Pre-trained Transformer) family, comprising references \cite{31_GPT,32_GPT,33_GPT}, stands out prominently. Unlike many other models that utilize a transformer encoder, the GPT series adopts a transformer decoder-centric architecture.


GPT-1 \cite{31_GPT} pioneered this approach, laying the foundational groundwork for its successors. GPT-2 \cite{32_GPT} and GPT-3 \cite{33_GPT} delved deeper into the exploration, underscoring the pivotal roles of massive datasets and expansive model sizes in enhancing transfer capabilities. The sheer scale and prowess of GPT-3 led to the development of ChatGPT\cite{94_Chatgpt}, which, owing to its unprecedented success in natural language tasks, has garnered significant attention in recent times.

The underlying mechanism of the GPT models revolves around predicting the next word in a sequence based on the preceding words, leveraging the power of deep neural networks. This autoregressive nature, combined with extensive pretraining on diverse textual data, equips the GPT models with a broad understanding of language, enabling them to generate coherent and contextually relevant text across a myriad of tasks.

\subsection{Choice of Generative Models}
\begin{figure}
    \centering
    \centerline{\includegraphics[width=0.6\columnwidth]{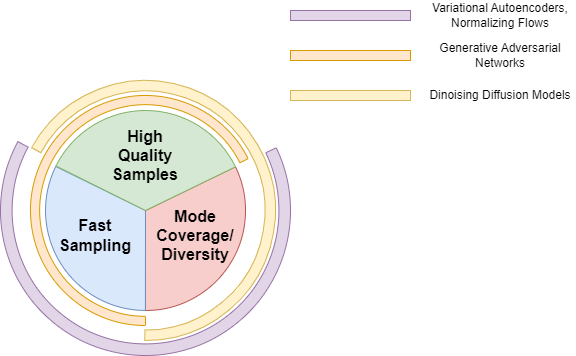}}
    \caption{Generative learning trilemma\cite{95_trilemma} }
    \label{fig:trade_off}
\end{figure}
 The choice of the right generative model depends on the specific demands of the task and the trade-offs one is willing to accept. Figure \ref{fig:trade_off} provides an illustrative overview of the capabilities and challenges faced by mainstream generative frameworks.

\begin{itemize}
    \item \textbf{Generative Adversarial Networks:} GANs are known for rapidly producing high-quality samples. This means that GDA could generate high-quality data samples. However, they suffer from mode collapse, leading to poor mode coverage~\cite {96_gan_fail, 97_gan_fail}. Consequently, they will fail to capture the entire data distribution which is vital to the GDA task.
    
    \item \textbf{Variational Autoencoders:} VAEs faithfully cover data modes. They are often chosen for tasks that require the reconstruction of input data, which is beneficial to the GDA. Nevertheless, VAEs tend to generate samples of lower visual quality compared to GANs.\cite{95_trilemma}

    \item \textbf{Diffusion Models:} Recent advancements have brought diffusion models to the forefront of generative modeling. Surpassing GANs in image generation quality~\citep{41_Diffusion,100_diffusion_good} and exhibiting good mode coverage, they are becoming a popular choice for high-fidelity image generation and restoration tasks. They also obtain good mode coverage, indicated by high likelihood.\cite{98_diffusion_good,99_diffusion_good,101_diffusion_good} Despite their capabilities, practical applications might be hindered by their computationally intensive sampling process, necessitating thousands of network evaluations. Consequently, tasks requiring real-time demand GDA are suitable for diffusion models.
    
    \item \textbf{GPT-based Models:} GPT models, while not explicitly discussed in the referenced  Figure \ref{fig:trade_off}, have gained prominence due to their prowess in natural language processing tasks. These models can be fine-tuned for various tasks, ranging from text generation to translation, code completion and even causal reasoning.\cite{102_gpt_good} GPT models, especially the larger ones, have been trained on vast amounts of data, giving them the ability to generate diverse and contextually rich samples, which can be particularly valuable for tasks like text augmentation. However, the GPT models, especially the larger GPT models, can be computationally intensive, making real-time data augmentation or augmentation on resource-constrained devices challenging.
\end{itemize}

In conclusion, the choice of a generative model for GDA applications should align with the specific requirements of the task, keeping in mind the inherent trade-offs in quality, mode coverage, and computational demands.
This structure provides an organized view of the capabilities and trade-offs associated with each generative model. You can further expand or refine based on specific needs and additional details as you progress with your thesis.

\section{Ways to use Generative Models}

Generative models, with their inherent potential to fabricate data, have found significant applications in the realm of GDA. These models, when employed judiciously, can not only augment the data pool but also enhance its quality. Primarily, there are two methodologies in which these generative models can be wielded: the art of prompt engineering and the science of harnessing latent codes. While both techniques have their merits, they cater to distinct requirements and scenarios.

\subsection{Prompt Engineering}

Prompt engineering is the strategy of crafting, refining, or selecting specific input prompts to guide a generative model, such as GPT or text-to-image models, in producing desired outputs. The essence of this technique lies in its ability to navigate the vast potential of a generative model without explicit retraining or fine-tuning. Instead, the model's behaviour is shaped and influenced by the prompts it receives. This section delves into the various facets of prompt engineering in GDA, discussing its direct application, the utilization of structured prompts, and innovative methods like ChatGPT-generated prompts.

\subsubsection{Direct Use}
\begin{figure}
    \centering
    \centerline{\includegraphics[width=0.8\columnwidth]{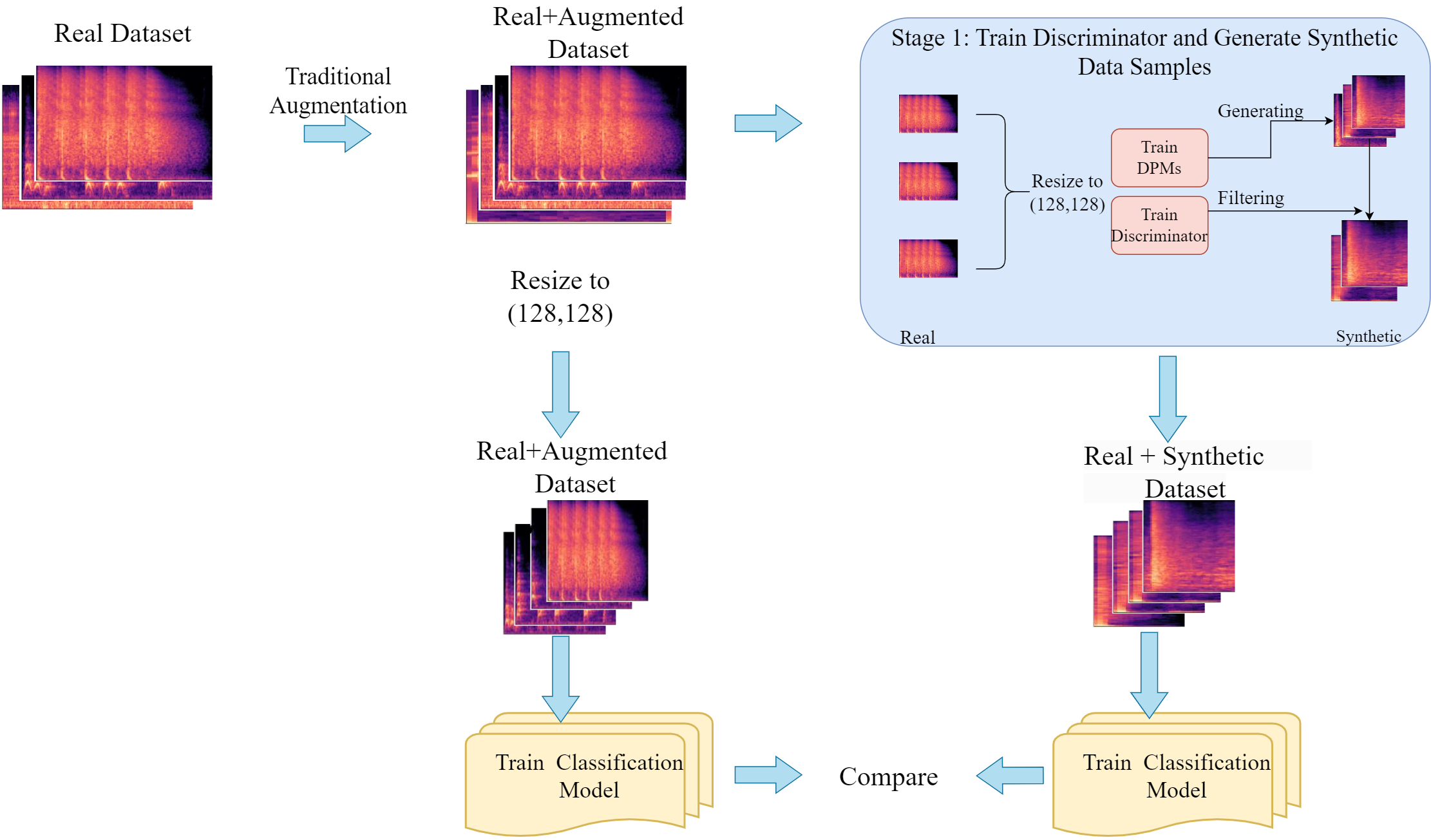}}
    \caption{Example of Direct Use in GDA\cite{103_direct_use} }
    \label{fig:direct_use}
\end{figure}

Here, the generative models are utilized by providing straightforward prompts, expecting the model's default behavior to generate appropriate data. The simplicity of direct use makes it a popular choice for quick tasks. A example of such use is illustrated in Figure \ref{fig:direct_use}. Due to the widespread use of Direct-Use, we only name a few examples in different applications.

 In the realm of environmental sound classification, \cite{103_direct_use} employed a diffusion probabilistic model combined with a top-K selection discriminator to augment data, showcasing the raw power and efficiency of generative models in this domain. Similarly, the work by ~\cite{104_direct_use} used a diffusion probabilistic model, coupled with transfer learning, to enhance weed recognition, emphasizing the adaptability and utility of such models for specific tasks. Additionally, ~\cite{105_direct_use} made strides in using deep convolutional generative adversarial networks for augmenting data related to environmental sound classification, further bolstering the case for generative models in auditory data tasks.

The medical sector, too, has leveraged the direct use of generative models. Esteban et al. adopted recurrent conditional GANs for the generation of real-valued medical time series data, underscoring the significance of generative models in healthcare applications~\cite{106_direct_use_medical}. Furthermore, ~\cite{107_direct_use_medical} tapped into the potential of GANs to create synthetic data augmentations for medical images, exemplifying the versatility of generative models in diverse imaging contexts.

In the domain of natural language processing, ~\cite{108_direct_use_QA} ventured into zero-shot multilingual synthetic question and answer generation to facilitate cross-lingual reading comprehension, demonstrating the proficiency of generative models in multilingual contexts. ~\cite{109_direct_use_few_shot} devised a methodology to guide generative language models specifically for data augmentation in few-shot text classification scenarios, elucidating the nuanced capabilities of these models in text-based tasks.

While the direct use of generative models offers a convenient and straightforward avenue for data augmentation across a range of applications, this approach is not devoid of challenges. 
First and foremost, without specific constraints or guidelines, the generated samples may deviate significantly from the true data distribution, leading to potential over-generalization or misrepresentation. Furthermore, in scenarios where precise and domain-specific augmentations are desired, the direct application might produce samples that lack the necessary nuances and details. It's also noteworthy that without tailored prompts or constraints, the generated data might include artefacts or inconsistencies, which could adversely affect the performance of models trained on such data. 

In summary, while the direct use of generative models holds promise, careful consideration and validation of the generated samples are imperative to ensure they align with the desired objectives and do not introduce unintended biases or anomalies.

\subsubsection{Using Pre-designed Prompt Structure}
\begin{figure}
    \centering
    \centerline{\includegraphics[width=0.5\columnwidth]{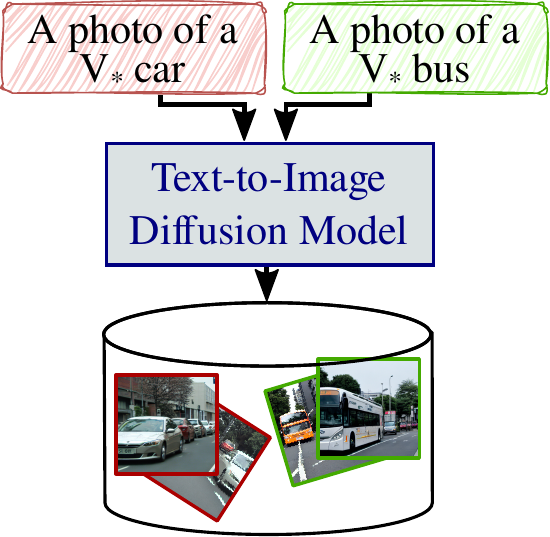}}
    \caption{Example of Prompt Structure in GDA\cite{118_prompt_structure} }
    \label{fig:prompt_structure}
\end{figure}

This strategy involves crafting prompts based on a pre-determined structure or pattern. The idea is to maintain consistency in the model's outputs, ensuring that the generated data aligns closely with specific requirements. The example of Prompt Structure is shown in Figure \ref{fig:prompt_structure}

In recent research endeavors, the role of pre-designed prompt structures in generative data augmentation (GDA) has come to the fore. Predefined manual prompts, which are often meticulously crafted to guide generative models towards generating specific outputs, have been found to be instrumental in GDA.

Notably, a substantial number of studies, including \cite{110_prompt_structure}, \cite{111_prompt_structure}, \cite{112_prompt_structure}, \cite{113_prompt_structure}, \cite{114_prompt_structure}, \cite{116_prompt_structure}, \cite{117_prompt_structure}, \cite{118_prompt_structure},\cite{122_prompt_structure} and \cite{119_prompt_structure}, employ predefined manual prompt structures. Examples of these prompt structures like "a photo of melanoma with irregular edges", "An image of [skin condition] on the [body part] of a [skin-type description] [noun]" and "The topic focuses on [theme], the label is [label]". Moreover, \cite{134}uses a more complex prompt structure by regarding the paper  describing a dataset  with some examples in it as a structure.

Diverging from this trend, Chia et al. \cite{116_prompt_structure} introduced an innovative approach that leverages literature-based prompts. Their methodology initiates with a search on the PubMed literature platform using an established vocabulary centered on drugs and cell lines. Following this, the retrieved sentences undergo clustering, and the medoids of each cluster are harnessed to craft prompts specifically tailored for PubMedBERT\cite{120_pubbert}. The cloze-prompt filling technique is then employed, and the tokens filled within the drug and cell line masks contribute to expanding the initial vocabulary.


Pre-designed prompt structures have emerged as a potent tool in GDA, offering a structured means to harness generative models for desired outputs. The advantages include a targeted approach, a potential for higher accuracy in specific applications, and adaptability across various domains. On the downside, reliance on manual prompts may introduce biases, limit the diversity of generated outputs, and might not scale efficiently. 

In addition, relying heavily on manual prompts can inadvertently introduce biases, which may not truly represent the underlying data distribution or the diversity inherent in real-world scenarios. Moreover, this manual intervention makes the approach labour-intensive and less scalable, especially when adapting to newer tasks or datasets. Given these constraints, there's a growing interest in more dynamic and adaptable prompt-generation methods. The emergence of large language models(LLMs) \cite{123_llm,124_llm}like ChatGPT hints at the next evolution in this domain, providing a potential solution to some of these challenges by automating and diversifying the prompting process.

\subsubsection{LLMs Generated Prompt}
A novel approach, this method leverages large language models(LLMs) like ChatGPT or T5\cite{150_T5} to generate prompts for the primary model dynamically. This recursive technique can lead to more diverse and nuanced data outputs, tapping into the collaborative potential of multiple models.

\begin{figure}
    \centering    \centerline{\includegraphics[width=1.1\columnwidth]{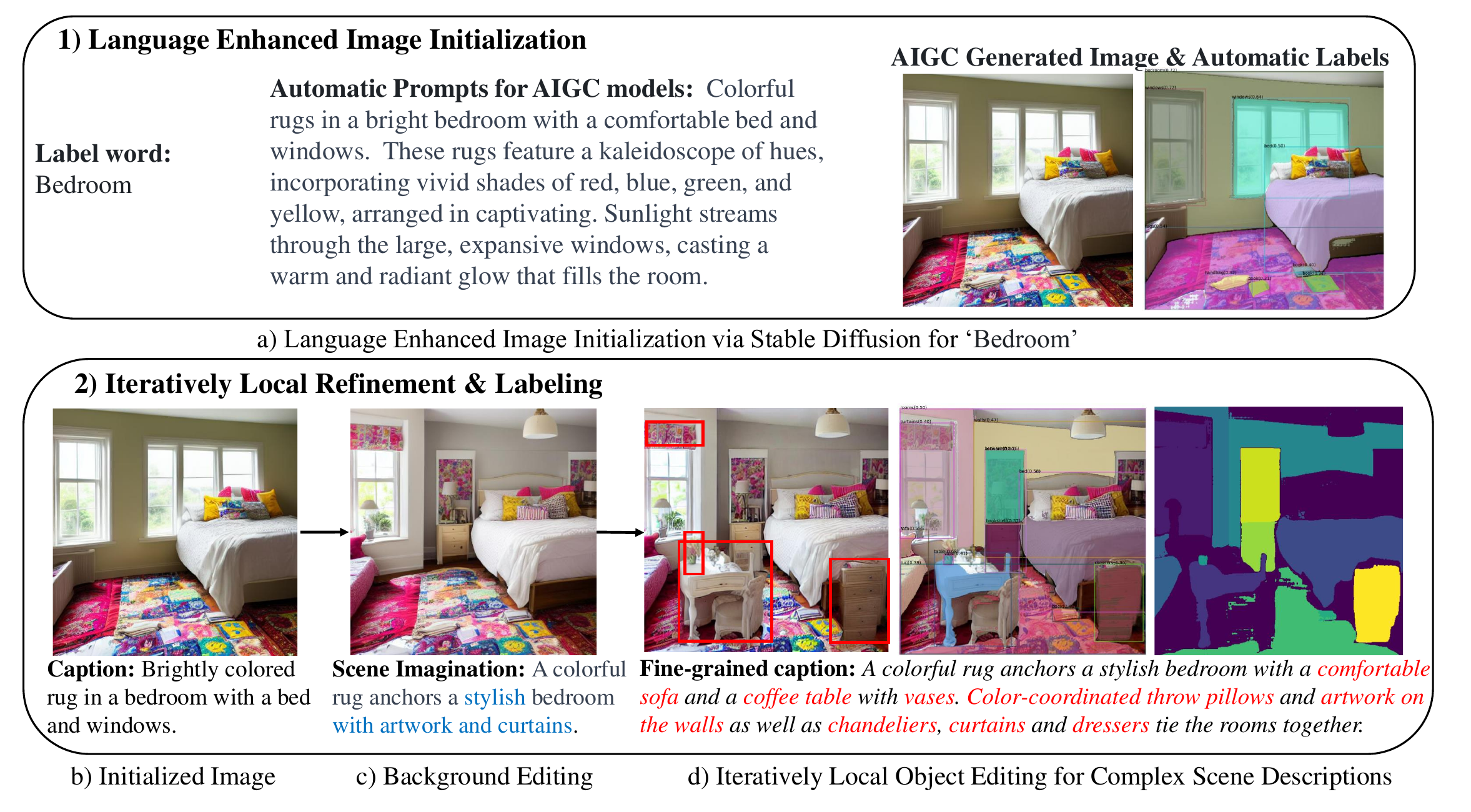}}
    \caption{Process of ChatGPT Generated Prompt\cite{121_chatgpt_prompt} }
    \label{fig:chatgpt_prompt}
\end{figure}

Firstly, recent research, specifically \cite{121_chatgpt_prompt}, \cite{126_chatgpt_prompt}, and \cite{127_chatgpt_prompt}, have illuminated the remarkable capabilities of large language models like ChatGPT. These studies emphasize that such models encompass an extensive reservoir of conceptual knowledge. This capability enables them to conceive and articulate rich, detailed descriptions from minimal input, a noteworthy example being: "A dog playing in a lush green park with a frisbee in its mouth. The dog should have a shiny coat of fur." Despite the achievements of ChatGPT in textual conceptualization, there is a noticeable limitation when we delve into the domain of existing AI-generated contents (AIGC). As per \cite{129_AIGC,131_AIGC,132_AIGC}, these AIGC models often generate images that, while accurate, lack complexity. They predominantly render simplistic images with a limited number of objects and backgrounds, thereby offering limited diversity. Such a limitation is a hindrance, especially when aiming for domain generalization, as highlighted by \cite{128_chatgpt_prompt}. Recognizing these challenges, \cite{121_chatgpt_prompt} proposes a novel approach called ChatGenImage. They architect an iterative pipeline which, at its core, seeks to mend missing intricacies and refine the output of generated images. By utilizing label foundation toolkits in conjunction with local editing prompts, this methodology iteratively enhances the richness and accuracy of AI-generated images. The culmination of this research offers a promising avenue: a systematic method to generate vast quantities of high-fidelity synthetic data. This data, replete with detailed labels, is scalable and presents a valuable asset for data augmentation, especially in situations marred by data scarcity.  The ChatGenImage pipeline(\ref{fig:chatgpt_prompt}) is schematically depicted through several stages. In stage (a), there's a symbiotic collaboration between LLMs and AIGC models, for instance, Stable Diffusion\cite{133_stable_diffusion}. Here, the user provides a specific label word, say "Bedroom". The system, leveraging the combined strength of LLMs and AIGC models, generates intricate images accompanied by detailed annotations. The subsequent stages, namely (b), (c), and (d), elucidate the image refinement process and the automated generation of fine-grained labels. This iterative procedure employs localized prompts, ensuring the image's alignment with intricate scene descriptions. The process is broken down further, involving both background and object editing, to ensure the synthetic image's fidelity and detail are progressively enhanced.

In addition, \cite{151_synthetic_ready} also used a pretrained word-to-sentenct T5 model to generate prompts for the text-to-image models.

LLMs generated prompts undoubtedly present a breakthrough in the realm of ways to use generative models in GDA, offering an intuitive means to conceive intricate, detailed descriptions from sparse input. The primary advantage of this approach lies in its ability to tap into a vast repository of conceptual knowledge encapsulated within the model. Consequently, it can imagine and produce rich descriptions that can guide the generation of sophisticated synthetic data.

However, as with all technologies, there are inherent limitations. While LLMs are adept at creating vivid textual descriptions, the conversion of these descriptions to high-quality synthetic visual data is not straightforward. Reliance solely on LLMs' textual output might lead to potential misinterpretations\cite{151_synthetic_ready} or overly generalized representations, especially when paired with AI-generated content models that are inherently limited in their diversity and complexity. Furthermore, the iterative pipeline required to refine and repair the generated images adds an additional layer of complexity to the process.

In light of the aforementioned, the LLMs generated prompt approach stands as a promising avenue for generative data augmentation, especially in data-scarce scenarios. However, users must approach with caution, balancing the model's capabilities with its constraints, and leveraging complementary tools to bridge the gap. This paves the way for future research endeavors, exploring the fusion of sophisticated LLMs like ChatGPT with advanced synthetic data generation techniques.

\subsection{Latent Code}

\begin{figure}
    \centering    \centerline{\includegraphics[width=1.1\columnwidth]{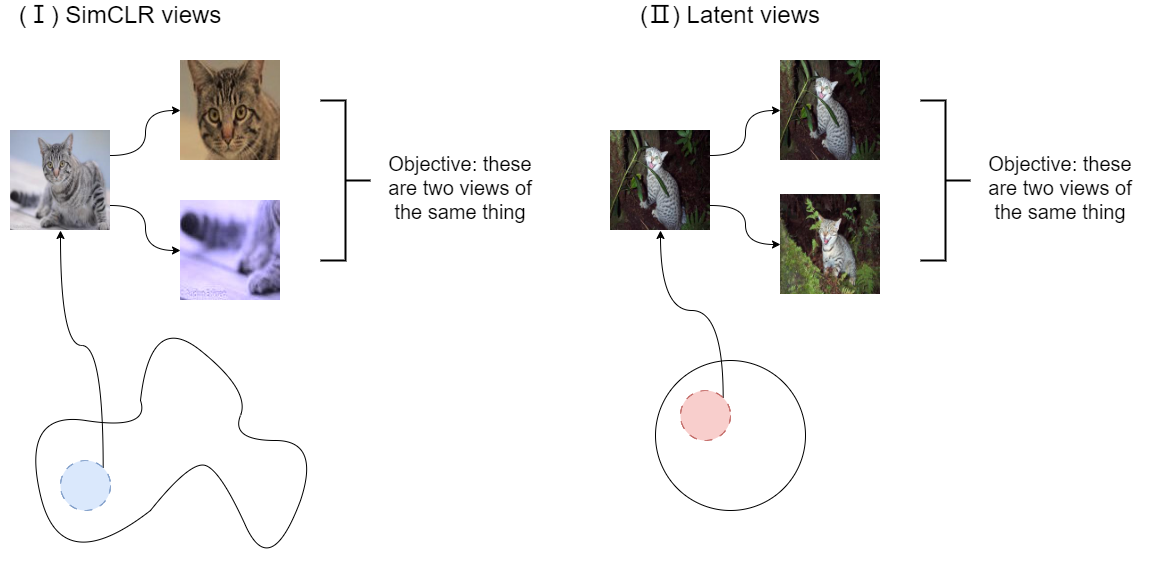}}
    \caption{Different ways of generating multiple views of the same “content"\cite{143_representation} }
    \label{fig:latent_transformation}
\end{figure}

\subsubsection{Latent Code Transformation}

The concept of latent code transformation pertains to the modulation of the latent code that serves as the input for generative models. This latent representation encapsulates an intricate and rich encoding, thereby enabling a plethora of diverse image transformations.

Generative models are imbued with the capability to manifest multifaceted perspectives of an image. Steering the generative models through alterations in their latent code allows for a myriad of transformations, such as camera adjustments, color modulations, and other intricate changes \cite{144_representation}. Indeed, as corroborated by further studies, the scope of these transformations is vast and profound \cite{145_representation,146_representation}. Taking a cue from these revelations, the work presented in \cite{143_representation} pivots from conventional methodologies. Instead of resorting to the transformation of an input image using canonical pixel-space ($\mathcal{X}$) data augmentations, they generate diverse views by sampling proximate points within the latent space, denoted by $\mathcal{Z}$.(\ref{fig:latent_transformation}) Such an approach is premised on the observation that points in close proximity within the latent space often generate images that retain the semantic essence of the same object. 

The pioneering work in \cite{143_representation} ventures beyond the mere elucidation of this latent code transformation technique. Their research delves deep into the exploration of how Generative Data Augmentation (GDA) can be harnessed to elevate the efficacy of contrastive learning \cite{148_representation,149_representation}. By leveraging the latent space of generative models, GDA unfurls the possibility of engendering multiple views of analogous image content. Such views can then be channelled for contrastive learning or can serve as inputs for other avant-garde representation learning algorithms. These algorithms traditionally depend on augmentations to engender multiple views tailored for contrastive learning. The empirical findings from their study underscore the significant advantage of utilizing latent-space views, positing that they offer a performance edge in comparison to relying solely on pixel-space transformations.

While the latent code transformation offers a nuanced approach to image generation and augmentation, it does not come without its own set of limitations. Manipulating the latent space can often lead to unpredictable or non-intuitive results, due to the highly intricate and abstract nature of these representations. There's also the challenge of understanding the precise relationship between minute changes in the latent space and their high-dimensional output effects. This lack of transparency can make it challenging to achieve very specific transformations or to guarantee consistent results across different generative models. Moreover, the reliance on latent codes for generating diverse views necessitates a deep understanding of the underlying generative model, which might not always be feasible for every application. Given these challenges, there emerges a need for more controlled and interpretable methods of image transformation and augmentation.

\begin{figure}
    \centering    \centerline{\includegraphics[width=0.8\columnwidth]{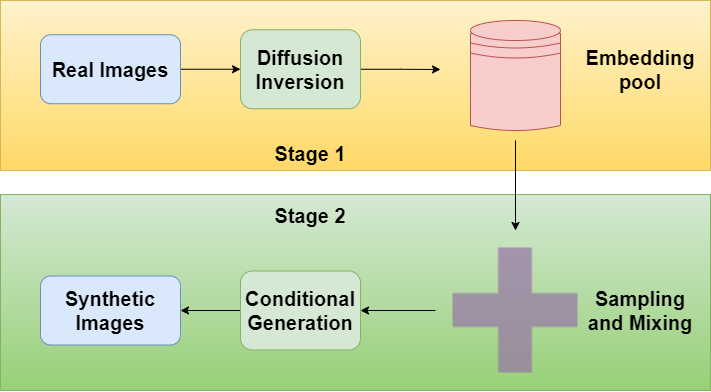}}
    \caption{Process of Diffusion Inversion\cite{121_chatgpt_prompt} }
    \label{fig:Inveresion}
\end{figure}

\subsubsection{Inversion}

Building on the challenges posed by latent code transformations, Inversion offers a potential solution. Instead of relying purely on latent space manipulations, inversion methods attempt to harness the initial data distributions more directly. In their pursuit to ensure comprehensive data coverage, \cite{137_latent_code} initially acquires a collection of embedding vectors. This is achieved by inverting each training image to the output domain of the text encoder, a process that has been extensively detailed in previous works \cite{138_inversion,139_inversion,140_Inversion,141_Inversion}.  

Subsequent to this inversion, they condition the Stable Diffusion model on a perturbed version of these vectors. Such a methodology facilitates the synthesis of a broad spectrum of novel training images, which not only resonate with the original dataset but also venture beyond its confines. This innovative approach ensures that the resultant synthetic images are imbued with profound semantic significance. At the same time, they exhibit a dynamic variability, a trait that can be attributed to the expansive knowledge reservoir inherent in the pre-trained image generator. The process in shown in Figure\ref{fig:Inveresion}. Their two-stage approach utilizes Stable Diffusion’s generalizable knowledge for targeted classification tasks by transforming real images into latent space and generating novel variants through inverse
diffusion with perturbed embeddings.

Diffusion Inversion stands as a testament to the advancements in synthetic data generation. This technique, while deceptively simple in its design, proves profoundly effective in generating synthetic data of superior quality, thereby elevating image classification capabilities, especially when leveraged alongside pre-trained generative models.

However, in the spirit of comprehensive reflection, it's pivotal to acknowledge certain constraints and broader implications associated with the method. Predominant issues encompass the staggering storage demands and the sub-optimal sampling efficiency of Stable Diffusion. Future trajectories might well hinge on integrating rapid sampling methodologies \cite{142_Inversion}, amplifying the dividends reaped from diffusion-centric data synthesis for discriminative models. A word of caution—implementing Diffusion Inversion in real-world paradigms warrants meticulous scrutiny and prudence, ensuring that potential pitfalls are pre-empted and the broader societal implications are thoroughly assessed.

\section{Data Selection}

Advancements in synthetic sample generation, driven by increasingly sophisticated generative models, have revolutionized numerous fields. However, a persistent challenge remains the propensity of these models to occasionally produce subpar, unrepresentative samples (\cite{152_data_selection}; \cite{153_data_selection}). 

\subsection{Top-k Selection}
To counteract the challenges of GDA, several studies \cite{9_labelling_diffusion}, \cite{155_top_selection},\cite{173_segmentation} have advocated for the deployment of pretrained discriminative networks or evaluation of a sample. The primary function of these networks is to sift through the pool of generated samples, filtering out the ones of low quality. The fundamental criterion driving this selection process hinges on the top-k evaluation rank. Concretely, a generated sample is decided for further use only if its genuine label ranks amongst the evaluation method's top-k ones. If it doesn't, the sample is discarded. The mathematical representation for the number of accepted samples is as follows:

\begin{equation} 
\label{eq6} 
G=\sum_{i=1}^N{\mathbb{I} \left( f_k\left( \boldsymbol{x}_i,c_i \right) =c_i \right)} 
\end{equation}
Where, $c_i$ denotes the label of the $i$-th generated sample, $\boldsymbol{x}_i$ symbolizes the said sample, $f_k$ represents the discriminator network equipped with top-k prediction, and $N$ stands for the number of generation epochs.

\subsection{Clip Filter}
The Clip Filter methodology \cite{151_synthetic_ready} presents another data sieving technique. Utilizing the zero-shot classification confidence of CLIP \cite{156_CLIP}, this method evaluates the quality of generated data, ruthlessly filtering out samples that fail to meet the confidence threshold.

\subsection{Realism Score-Based Approach}
Furthermore, \cite{104_direct_use} postulates a post-processing data selection paradigm, which kicks into action after the training process. This method leverages a realism score-based approach \citep{154_realism}, where the realism score ($R$) quantifies the proximity of a synthetic sample to genuine data:

\begin{equation} 
\label{eq:r}
    R(\phi_g, \Phi_r) = \max_{\phi_r} \Biggl\{ \frac{\| \phi_r - NN_k (\phi_r, \Phi_r)\|_2}{\| \phi_g - \phi_r\|_2} \Biggl\},
\end{equation}
In the above equation, $\phi_g$ and $\phi_r$ depict the extracted feature vectors of synthetic and real images, respectively. When the computed $R \geq 1$, it signifies that the feature vector $\phi_g$ nestles within the neighboring hypersphere of at least a single $\phi_r$. To filter the chaff from the grain, the realism score $R$ is computed for every individually synthesized image. Samples with an $R$ value less than 1 are systematically discarded, ensuring that only the crème-de-la-crème synthetic samples are fed into deep learning models.

\cite{117_prompt_structure} also proposes a selection method similar to the realism sore-based approach:

Given:
\begin{equation}
G_n=\{G_{n}^{i}\}_{i=1}^{N},\  R_n=\{R_{n}^{i}\}_{i=1}^{M}
\end{equation}
$G_n$ denotes the generated samples for category $n$, with $G_{n}^{i}$ representing the $i$-th generated instance. $R_n$ signifies the real samples for category $n$ from the few-shot dataset. $N$ and $M$ indicate the total counts of generated and real samples, respectively.

The sample-based selection utilizes the CLIP score or its image features:
\begin{equation}
\label{equ:2}
	R_s = \left\{
	\begin{array}{l}
		Se(CLIP(G_n,T)) \\
		Se(cos(CLIP(G_n),CLIP(R_n)))
	\end{array}
	\right.
\end{equation}
Here, $T$ stands for input text, and $CLIP$ produces a similarity score between text and image. In the absence of $T$, $CLIP$ refers to its image encoder, processing generated and real samples to fetch image features. The selection module $Se$ employs uniform sampling to identify representative samples, calculating distances via cosine similarity. $R_s$ is the outcome of this sample-based method, representing the selected samples.

\subsection{Cluster-based Method}
\cite{117_prompt_structure} also uses a clustering algorithm  to find the representative samples. 


\begin{equation}
C_n = Clu(CLIP(I_n),k_n)
\end{equation}
where $C_n$ is the class center of category $n$. $Clu$ is the clustering method. $k_n$ is the number of centers of category $n$. They use k-means or spectral clustering to compute the class centre. They find that using the spectral clustering algorithm achieves the best performance.

\begin{equation}
R_c = \arg \underset{i}{min}\ cos(C_n,G_{n}^{i})_{i=1}^{N}
\end{equation}

They select the sample nearest to the $k_n$ cluster centroids $C_n$ as the representative samples for category $n$. The distance is computed by the cosine similarity. $R_c$ is the samples selected by the cluster-based methods.

Generative models sometimes falter by generating samples that don't adequately mirror real data. To mitigate this, the research community has championed various selection methodologies, prominently relying on discriminators, CLIP scores, and clustering techniques. 

On the brighter side, these methods are adept at enhancing data quality, streamlining the integration of GDA into practical applications, especially deep learning models. They cater to the demand for high-quality, diverse, and representative synthetic samples, making them invaluable tools in the synthetic data arsenal.

However, it's essential to tread with caution. The reliance on discriminative networks or specific scoring models brings its baggage – these techniques are only as proficient as their foundational models. If these models have biases or are not accurate enough to reflect the data distribution, the selection could be flawed. Additionally, the use of a realism score or similar metrics to gauge sample quality is subjective and can vary based on the domain or specific application. Clustering-based methods, like those described by \cite{117_prompt_structure}, depend on the efficiency and relevance of the clustering algorithm. Clustering can sometimes group data in ways that aren't meaningful or miss nuances in the data.

In summation, while the current methodologies present an impressive arsenal to refine synthetic data, they're not without their idiosyncratic challenges. Balancing the merits with the inherent limitations will be crucial as we march forward in the evolving landscape of GDA.

\section{Validation of GDA}
In this section, we aim to list ways to validate the effectiveness of GDA through a multi-pronged approach. We will first delve into empirical performance metrics, a well-acknowledged method that has found common ground in many related research endeavours. Next, we will employ a decomposition visualization analysis to ascertain GDA's impact visually. Lastly, we shall adopt a theoretical analysis to provide a more profound understanding of GDA's generalization capabilities.
\subsection{Empirical Performance Analysis}
It is one of the most commonly used methods to validate the effectiveness of GDA. These research works use the common evaluation method to validate the GDA's effectiveness. For example, \cite{143_representation} uses Top-1 accuracy and average precision to validate the effectiveness in contrastive learning,  \cite{104_direct_use}, \cite{119_prompt_structure} use FID\cite{172_FID} to demonstrate that the generative models can generate high-quality data samples, \cite{173_segmentation} uses mIOU to show GDA's effect in segmentation, \cite{9_labelling_diffusion,105_direct_use} use the classification accuracy as validation for GDA. This validation method is commonly used and can be easily used to validate the effectiveness of GDA. However, this method can not shed light on the inner reasons why GDA can improve the generalization of deep learning models because the method can only show the results.
\subsection{Decomposition Visualization Analysis}
In pursuit of determining the presence of semantically meaningful features in synthetic images, \cite{119_prompt_structure} utilized a ResNet50 model \cite{174_resnet} pre-trained on ImageNet weights \cite{175_imagenet}. This model was fine-tuned only on the classifier head using the real CIFAR10 dataset. The feature spaces of both the real images and the synthetic CIFAR10 images generated by \cite{119_prompt_structure} were subsequently visualized using a t-SNE plot \cite{176_tsn}. If a notable difference existed in the semantic features of synthetic images as opposed to real images, it would lead to model misclassification.

Their resultant visualization revealed that the features of real and synthetic images intertwined, suggesting a degree of similarity. Additionally, the clusters formed by various classes showed that the model could classify 76.61\% of the real images and 63.8\% of the synthetic ones accurately. These observations attest to the synthetic images possessing adequate meaningful semantic features, rendering them suitable for training.

However, while such visualizations provide some insights, it's crucial to acknowledge that simple visualization and comparison methodologies have their limitations. They may not consistently produce reliable results, especially if the dataset is altered. Solely depending on them can be misleading, thus emphasizing the need for a more theoretical and in-depth analysis.

\subsection{Theoretical Analysis}

Theoretical analysis forms the bedrock of understanding the underlying principles that govern machine learning models and algorithms. Whereas empirical results provide us with practical insights, a theoretical perspective ensures that we have a foundational grasp of the mechanics at play, making our conclusions more robust and universally applicable. In the context of discerning the semantic authenticity of generative data samples, it's imperative to marry the observational with the theoretical, especially to grasp how the generative data samples influence learning dynamics. This section delves into the intricacies of machine learning theory relevant to GDA and what conclusions they come to.

Notations are shown in Section 2.1. 

\subsubsection{Generalization via algorithmic stability}

In the evolving domain of machine learning, one of the foremost challenges is to ensure that a model not only performs well on its training data but also generalizes effectively to unseen data. Herein lies the critical role of algorithmic stability analysis. This analytical tool is instrumental in offering definitive guarantees regarding a model's generalization capabilities. In other words, it serves as a bridge between a model's performance during training and its likely performance on real-world, unfamiliar data.

What truly sets stability analysis apart is its ability to harness the unique attributes of the algorithm at hand, and in doing so, offer algorithm-specific generalization bounds. This tailored approach ensures that the bounds are as relevant and accurate as possible for a given algorithm. Numerous notations of stability have been introduced over the years, as documented in various works~\cite{157_theory, 158_theory,159_theory,160_theory}. Among this plethora of notations, uniform stability has emerged as a predominant choice. Its applicability spans a diverse range of learning algorithms, encompassing methods like regularized empirical risk minimization (ERM) algorithms~\cite{157_theory} and even the popular stochastic gradient descent (SGD)~\cite{159_theory,161_theory, 162_theory}.

\textbf{Uniform stability:}
An algorithm, represented by $\sA$, attains the status of being uniformly $\beta_m$-stable with respect to a specific loss function, denoted by $\ell$, provided the condition mentioned below is satisfied:
\begin{equation}
\forall S \in \sZ^m, \forall \bz \in \sZ, \forall i \in [m], \sup_{\bz} \abs{\ell(\sA(S), \bz) - \ell(\sA(S^i), \bz)} \leq \beta_m.
\end{equation}

The insights from the pioneering work~\cite{157_theory} have shown that for a learning algorithm to be $\beta_m$-stable, it should meet the criterion $\beta_m = o(1/\sqrt{m})$ for the generalization bound to converge. This is a profound revelation, but it's worth noting that this stipulation might not be met in certain modern machine learning scenarios~\cite{163_theory}, thereby rendering the provided guarantees void.

In a bid to further enhance these bounds, more recent studies~\cite{164_theory, 165_theory, 166_theory} have taken innovative routes, proposing tighter concentration inequalities. Notably, the work delineated in~\cite{166_theory} introduced a moment bound, achieving a nearly unparalleled generalization guarantee. This groundbreaking approach requires just $\beta_m = o(1 / \log m)$ for convergence.

\textbf{Theorem 6.1:}(Corollary 8,~\cite{166_theory})
\label{thm: classical stability bound}
\textbf{Assume that $\sA$ is a $\beta_m$-stable learning algorithm and the loss function $\ell$ is bounded by $M$. Given a training set $S$ with $m$ i.i.d. examples sampled from the distribution $\sD$, then for any $\delta \in (0,1)$, with probability at least $1-\delta$, it holds that}
\begin{equation}
\abs{\sR_{\sD}(\sA(S)) - \widehat{\sR}_{S}(A(S))} \lesssim \log (m) \beta_m \log \left(\frac{1}{\delta}\right)+ M \sqrt{\frac{1}{m}\log \left(\frac{1}{\delta}\right)}.
\end{equation}

\cite{171_theory} note that all generalization bounds mentioned above require a primary condition: data points are drawn i.i.d. according to the population distribution $\sD$. However, it no longer holds in the setting of GDA. On the one hand, the distribution $\DG$ learned by the generative model is generally not the same as the true distribution $\sD$. On the other hand, the learned $\DG$ is heavily dependent on the sampled dataset $S$. This property brings obstacles to the derivation of the generalization bound for GDA. Furthermore, though there exists some non-i.i.d. stability bounds~\cite{167_theory,168_theory,169_theory}, it is still unclear whether these techniques are suitable in the GDA setting.

\subsubsection{General generalization bound}
\label{sec: General generalization bound}

Understanding the generalization capabilities of GDA necessitates an investigation into the generalization error of the hypothesis $\sA(\Saug)$, which is learned from the augmented dataset $\Saug$. This error, termed \textit{Gen-error} in Section 2, denotes the discrepancy between the empirical error and the true error. A pivotal step in \cite{171_theory}'s analysis involved decomposing this error as shown below:
\begin{equation}
\abs{\textit{{Gen-error}}} = \underbrace{\abs{\sR_{\sD}(\sA(\Saug)) - \sR_{\Daug(S)}(\sA(\Saug))}}_{\text{Divergence due to Distributions}} + \underbrace{\abs{ \sR_{\Daug(S)}(\sA(\Saug)) - \widehat{\sR}_{\Saug}(\sA(\Saug))}}_{\text{Generalization error with mixed distribution}}.
\end{equation}

The first term encapsulates the divergence between the augmented distribution, $\Daug(S)$, and the genuine distribution, $\sD$. This divergence, which can be quantified using metrics such as $d_{\mathrm{TV}}, d_{\mathrm{KL}}$, offers insights into the efficacy of the selected generative model. On the other hand, the second term poses challenges. Directly applying classical stability bounds (like Theorem~\ref{thm: classical stability bound}) is unfeasible since $\Saug$ consists of non-i.i.d. samples. A breakthrough came when \cite{171_theory} observed that while $S$ upholds the i.i.d. assumption, $S_G$ adheres to the conditional i.i.d. premise given a fixed $S$. This insight paved the way for further decomposition and enabled the utilization of sharp moment inequalities~\cite{166_theory} to set an upper bound.

\cite{171_theory} solidified this analysis with the subsequent theorem:

\textbf{Theorem 6.1.2: Comprehensive Generalization Bounds for GDA}
\label{thm: comprehensive generalization bound}
\begin{align*}
\abs{\text{Gen-error}} 
&\lesssim \underbrace{\frac{m_G}{m_T}M d_{\mathrm{TV}} \left(\sD , \DG \right)}_{\text{Divergence among Distributions}} \\
&+ \frac{M(\sqrt{m_S} + \sqrt{m_G}) + m_S\sqrt{m_G}\beta_{m_T} }{m_T} \sqrt{\log \left(\frac{1}{\delta}\right)}\\
&+\frac{\beta_{m_T} \left(m_S \log m_S + m_G \log m_G\right) + m_S \log m_S M \sT(m_S, m_G) }{m_T} \log \left(\frac{1}{\delta}\right),
\end{align*}
\textbf{where $\sT(m_S, m_G)$ assesses the stability of the learned distribution by examining how a single change in the training dataset impacts the generative model's outcome.}

\cite{171_theory} comes to the following conclusions:

\textbf{Examining the Bound's Precision.} A noteworthy attribute of \cite{171_theory} generalization bound is its adaptability. Specifically, for $m_G = 0$, Theorem~\ref{thm: comprehensive generalization bound} seamlessly reverts to the classical stability bound, Theorem~\ref{thm: classical stability bound}. This elasticity ensures that \cite{171_theory} bound not only caters to non-i.i.d. scenarios but also encapsulates the traditional i.i.d. setting, inheriting its near-optimality as illustrated by~\cite{166_theory}. Further refinements for instances where $m_G > 0$ are intriguing areas for future exploration.

\textbf{Stability Dynamics of the Learned Distribution $\DG$.} The term $\sT(m_S, m_G)$ in \cite{171_theory} theorem embodies the stability dynamics inherent to the distribution learned by the generative model. \cite{171_theory} findings underscore the relationship between the model's stability and GDA's performance—the more stable the learned distribution, the more robust the GDA's outcomes. Notably, while earlier works, such as \cite{170_theory}, delved into the uniform stability of generative algorithms, the nuances associated with $\sT(m_S, m_G)$ in \cite{171_theory}'s bound represent a pioneering effort in the realm of GDA.

\textbf{Optimal Augmentation Size: A Balancing Act.} When scrutinizing Theorem~\ref{thm: comprehensive generalization bound}, it's evident that while increasing $m_G$ nullifies the generalization error associated with mixed distribution, it doesn't necessarily temper the divergence among distributions. This dynamic underscores a nuanced balance between rapid convergence and augmented data consumption. Depending on the divergence's order, enhancing $m_G$ can either bolster convergence or amplify data usage, necessitating the establishment of an efficient augmentation size, denoted as $m_{G, \mathrm{order}}^*$. 

\begin{equation*}
    m_{G, \mathrm{order}}^* = \inf_{m_G}\left\{ \text{generalization error w.r.t. mixed distribution} \\ \lesssim \text{distributions' divergence} \right\}.
\end{equation*}

Yet, pinning down explicit forms for both $m_{G, \mathrm{order}}^*$ and its optimal counterpart, $m_G^*$, remains challenging due to the intricacies involving $\beta_{m_T}$ and $\sT(m_S, m_G)$.

\textbf{GDA's Learning Rate Dynamics.} Delving deeper into the nuances of learning rates, especially with respect to $m_S$, reveals that the divergence $d_{\mathrm{TV}} \left(\sD, \DG \right)$ plays a pivotal role in discerning whether GDA can harness accelerated learning rates. By juxtaposing Theorem~\ref{thm: comprehensive generalization bound} with Theorem~\ref{thm: classical stability bound}, \cite{171_theory} can elucidate the conditions under which GDA can either benefit from or be deprived of enhanced learning rates.

Given the bound on the loss function $\ell$:

\begin{itemize}
    \item When $d_{\mathrm{TV}} \left(\sD, \DG \right)$ is insignificantly small compared to the max of $\log(m)\beta_m$ and $1 / \sqrt{m}$, GDA capitalizes on accelerated learning rates.
    \item Conversely, if $d_{\mathrm{TV}} \left(\sD, \DG \right)$ mirrors or overshadows the max of $\log(m)\beta_m$ and $1 / \sqrt{m}$, GDA's potential for rapid learning diminishes.
\end{itemize}

\section{Applications}
The ultimate goal of generative data augmentation is to increase the performance in downstream tasks of all kinds of areas. GDA has been widely applied to Natural Language Processing, Computer vision and Signal Processing in the past three years since GAN, VAE and Diffusion models were accepted as reasonable and efficient ways to improve performance. Especially, in tasks where data could not meet the growing demands of models' size, amazing achievements have been achieved. 

We collated articles on the application of generative data augmentation since 2020 and quantified the improvements made by GDA in some specific areas. In particular, we did not specifically review the use of VAE-based GDA methods. VAE-based GDA methods are not as widely used as GAN-based GDA and diffusion-based GDA. So there are not many articles related to VAE-based GDA in a specific area. We only cite few articles related to VAE-based GDA in the areas we explore if needed.
\subsection{GDA in Medical Image Analysis}
With the presentation of several large datasets, the restriction of limited data for some computer vision tasks has been gradually overcome. Just as COCO was used for detection and classification, ImageNet and Open Images were used for detection. GDA is effective, though not spectacular, for such tasks. In addition, there are still some research areas that are limited by the lack of available public datasets, e.g. in the area of medical image analysis, where most tasks are publicly available only in terms of code and test results, but the private datasets used for training are not available due to patient privacy. For independent researchers in the field of medical image analysis, limited data is still the biggest limitation to trying innovative approaches. We will present in detail how GDA has been applied to medical image analysis and the achievements obtained.

\subsubsection{Diffusion-based Methods}
Diffusion Models are the primary data augmentation ways for Medical Image Analysis. Akrout, Mohamed et al. apply Diffusion-based Data Augmentation for Skin Disease Classification. They explore different accuracies from original Medical Datasets to fully synthetic images\cite{apl60}. As is shown in Table 2, Hybrid data achieves the best performance in the restriction of the same dataset size.
\renewcommand{\arraystretch}{1.5}
\begin{table}[]
\begin{tabular}{cccccccc}
\hline
\multirow{2}{*}{\textbf{Dataset}} & \multicolumn{2}{c}{\textbf{\# of images}} & \multicolumn{5}{c}{\textbf{Accuracy}}                                      \\ \cline{2-8} 
           & Real & stnthetic & Top-1   & Top-2            & Top-3   & Top-4            & Top-5            \\ \hline
Real-small & 250  & 0         & 53.41\% & 73.51\%          & 83.22\% & 89.75\%          & 96.45\%          \\ \hline
Real       & 500  & 0         & 54.05\% & \textbf{73.95\%} & 84.84\% & 91.49\%          & \textbf{96.96\%} \\ \hline
Hybrid                            & 250                 & 250                 & \textbf{54.13\%} & 73.23\% & \textbf{85.01\%} & \textbf{92,16\%} & 96.65\% \\ \hline
Synthetic  & 0    & 500       & 47.29\% & 70.71\%          & 84.09\% & \textbf{92.16\%} & 96.85\%          \\ \hline
\end{tabular}
\caption{Top-1 to top-5 skin disease classification accuracy of Akrout's model}
\end{table}
Sagers, Luke et al. use DALL·E 2 and generate synthetic images, improving the classification of skin disease overall, especially for underrepresented groups\cite{apl61}. Wilde, Bram De et al. increase the diagnostic accuracy for detecting prostate cancer on MRI, which is a challenging multi-modal imaging modality, from 0.78 to 0.80 across a Diffusion-based model for text-to-image generation\cite{apl62}. Pinaya, Tudosiu et al. propose a diffusion-based method which could be conditioned on covariates such as age, sex and brain structure volumes to generate synthetic images from high-resolution 3D brain images\cite{apl63}. They created a synthetic dataset with 100,000 brain images called LDM 100K together with the conditioning information and made it openly available to the scientific community\cite{apl64}. Moghadam, Van Dalen et al. investigate the potential use of diffusion probabilistic models along with prioritized morphology weighting and color normalization to synthesize high-quality histopathology images of brain cancer for the first time\cite{apl65}. The bottom row in Figure 12 shows the images generated by their diffusion model. Dorjsembe, Odonchimed et al. present a novel 3D-DDPM trained on a small set of training images with only 250 steps\cite{apl80}. The samples generated by their model have similar scores to that of the real data. Kim, Ye et al. explore the area of temporal medical image generation such as 4D cardiac volume data. They propose a diffusion deformable model to generate the 4D cardiac MR image and achieve high performance on temporal volume generation\cite{apl66}. Waibel, Ernst et al. introduce DISPR, a diffusion-based model for solving the inverse problem of three-dimensional cell shape prediction from two-dimensional single cell microscopy images. They improve the macro F1 score from 55.2±4.6\% to 72.2±4.9\% by using DISPR to oversample training data of single red blood cell classification\cite{apl67}. The improvement of their work is shown in Figure 13. Packhäuser, Folle et al. use latent diffusion models for train1191200221ing thoracic abnormality classification systems, achieving competitive results with a performance gap of only 3.5\% in the area under the receiver operating characteristic curve\cite{apl68}. The comparison of the abnormality identification performance of CheXNet when using either real or exclusively synthetic training sets can be seen in Table 3. Jiang, Mao et al. design the Conditioned Latent Diffusion Model(CoLa-Diff), the first diffusion-based multi-modality MRI synthesis model\cite{apl69}. They get a better score than many GAN-based methods and the Latent diffusion models(LDM). The visualization of synthesized images, detail enlargements (row 1 and 3) and corresponding error maps (row 2 and 4) can be seen in Figure 14.

\begin{figure}
    \centering
    \centerline{\includegraphics[width=1\columnwidth]{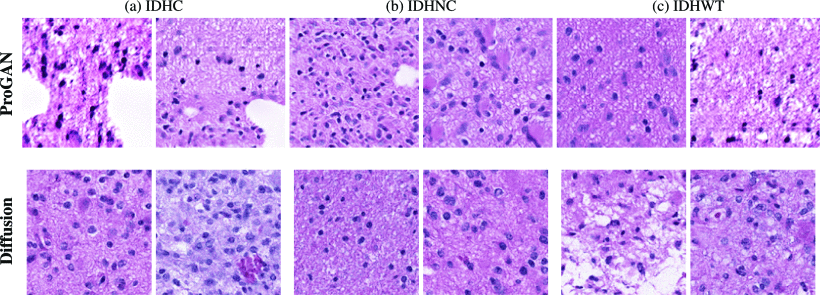}}
    \caption{High-quality brain cancer images generated by Moghadam et al.}
    \label{fig:enter-label}
\end{figure}

\begin{figure}
    \centering
    \centerline{\includegraphics[width=1\columnwidth]{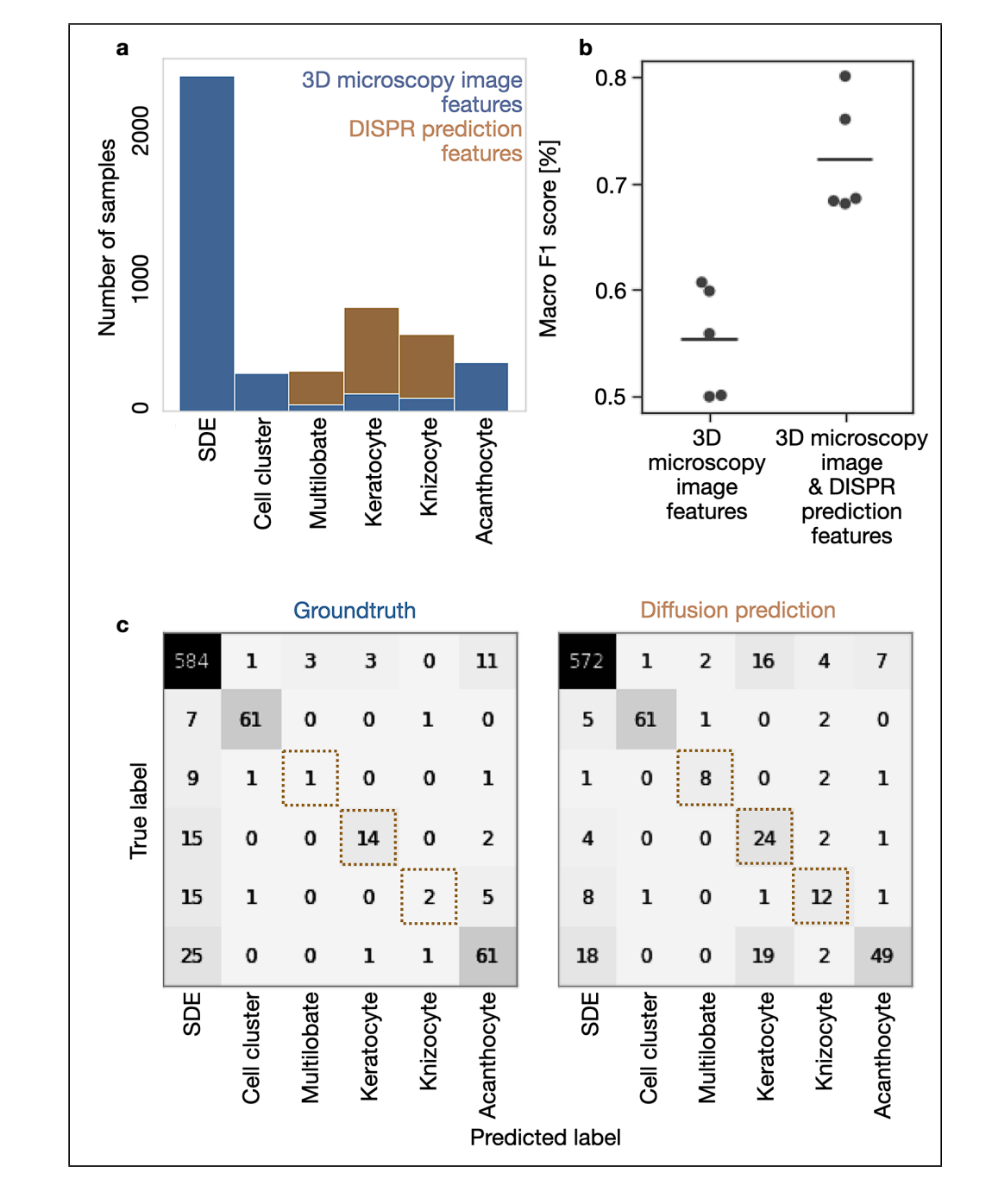}}
    \caption{The performance of Waibel's model}
    \label{fig:enter-label}
\end{figure}

\begin{figure}
    \centering
    \centerline{\includegraphics[width=1\columnwidth]{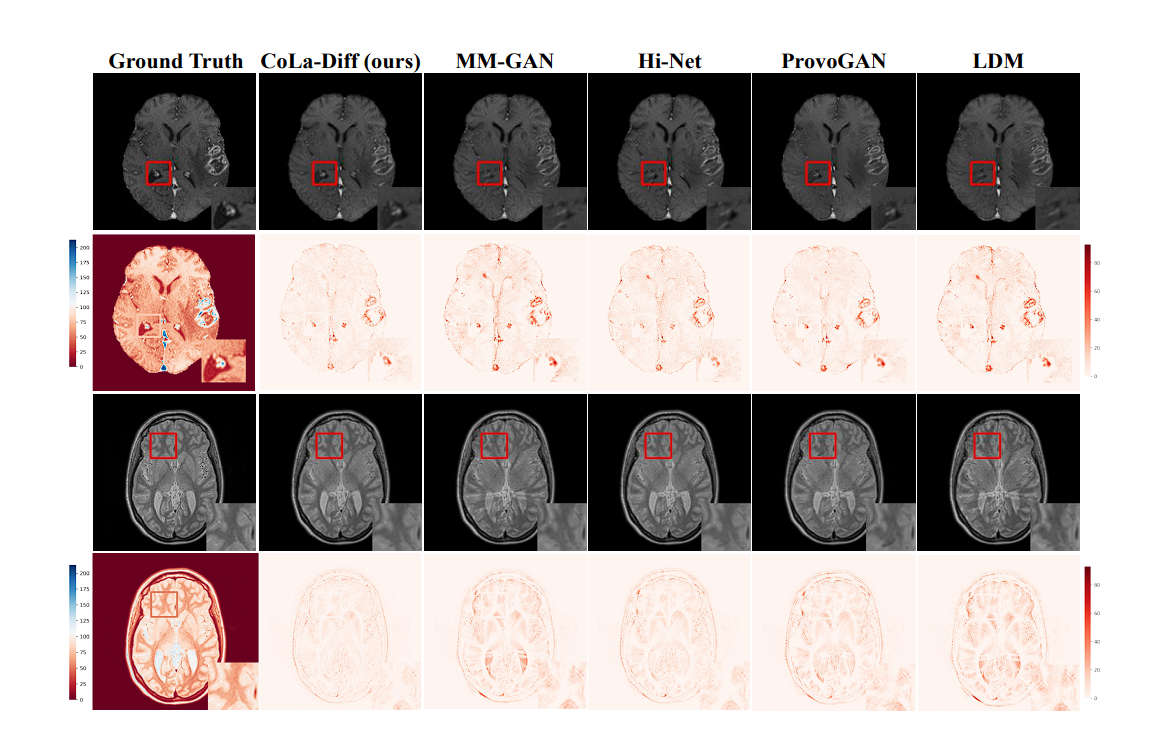}}
    \caption{The visualization for Jiang's model Copyright (c) 2023, Lilah}
    \label{fig:enter-label}
\end{figure}

\begin{table}[]
\begin{tabularx}{\textwidth}{lXXX}
\hline
\textbf{Training set}       & \textbf{Real} & Syn(PGGAN) & Syn(LDM)    \\ \hline
\textbf{Atelectasis}        & 81.3±0.8      & 70.1±1.2   & 76.2±0.4 \\
\textbf{Cardiomegaly}       & 92.9±0.6      & 86.4±1.4   & 88.6±0.8    \\
\textbf{Consolidation}      & 74.8±1.0      & 68.0±2.8   & 75.5±1.1    \\
\textbf{Edema}              & 92.8±0.8      & 84.4±2.8   & 87.5±1.7    \\
\textbf{Effusion}           & 90.7±0.4      & 83.2±0.9   & 85.9±0.9    \\
\textbf{Emphysema}          & 88.1±0.8      & 76.5±1.6   & 83.9±1.0    \\
\textbf{Fibrosis}           & 80.8±1.0      & 69.4±2.9   & 77.3±0.6    \\
\textbf{Hernia}             & 93.5±1.5      & 80.6±3.5   & 93.7±1.4    \\
\textbf{Infiltration}       & 68.7±0.2      & 59.1±0.6   & 63.4±0.5    \\
\textbf{Mass}               & 81.0±1.3      & 67.7±0.9   & 76.9±0.8    \\
\textbf{Nodule}             & 71.4±1.0      & 60.8±1.2   & 67.8±0.7    \\
\textbf{Pleural Thickening} & 75.5±1.2      & 68.4±1.3   & 73.4±0.8    \\
\textbf{Pneumonia}          & 70.8±3.3      & 61.5±5.1   & 65.8±2.0    \\
\textbf{Pneumothorax}       & 79.9±0.7      & 70.1±1.5   & 77.0±0.5    \\ \hline
\textbf{Mean}               & 81.6±0.4      & 71.9±0.8   & 78.1±0.3    \\ \hline
\end{tabularx}
\caption{The comparison of the abnormality identification performance for Packhäuser's model}
\end{table}

\subsubsection{GAN-based Methods}
GAN-based models gain attention as the first generative data augmentation model is GAN-based. GANs generate high-quality samples across the use of discriminators. Though the mainstay of GAN-based models is being replaced by diffusion-based models for the mode collapse in training, there are still many amazing GAN-based works. 

We tabulate the applications of GAN in papers from 2020 onwards in the Medical Image Analysis area and then plot Table 4.
\renewcommand{\arraystretch}{1.7}
{\small
\begin{longtable}{|p{2cm}|p{3cm}|p{2cm}|p{2cm}|p{2cm}|}
\hline
\multicolumn{1}{|c|}{\textbf{Publication}} & \multicolumn{1}{c|}{\textbf{Task}}                         & \multicolumn{1}{c|}{\textbf{Dataset}}                              & \multicolumn{1}{p{2cm}|}{\textbf{Eval. Metrics}}               & \multicolumn{1}{c|}{\textbf{Architecture}} \\ \hline
Waheed, Goyal et al.(2020)\cite{apl1}         & covid-19 detection                                        & IEEE CCX                                                          & Acc., Sens., Spec.                                       & ACGAN                                     \\ \hline
Sun, Wang et al.(2020)\cite{apl2}             & lesion detection                                          & Brats2018                                                         & Acc.                                                     & ANT-GAN                                   \\ \hline
Wang, Zhang et al.(2020)\cite{apl3}           & lung nodule synthesis                                     & LIDC-IDRI                                                         & Acc., F1                                                 & MG-CGAN                                   \\ \hline
Geng, Yao et al.(2020)\cite{apl4}             & fMRI disease classification                               & ADHD, ABIDE                                                       & Acc., Sens., Spec., AUC                                  & FC-GAN                                    \\ \hline
Pang, Wong et al.(2021)\cite{apl5}            & Breast ultrasound mass classification                     & Private dataset                                                   & Acc., Sens., Spec.                                       & TGAN                                      \\ \hline
Barile, Marzullo et al.(2021)\cite{apl6}      & Brain structural connectivity in multiple sclerosis       & Private dataset                                                   & Prec., Recall, F1                                        & AAE                                       \\ \hline
Shen, Hao et al.(2021)\cite{apl7}             & Mammogram with contextual information                     & DDSM                                                              & LPIPS, Recall                                            & DCGAN, InfillingGAN                       \\ \hline
Ambita, Boquio et al.(2021)\cite{apl8}        & Covid-19 detection                                        & COVID-CT, SARS-COV2                                               & Acc.                                                     & SAGAN                                     \\ \hline
Hirte, Platscher et al.(2021)\cite{apl9}      & MR brain images generation                                & Private dataset                                                   & -                                                        & StyleGAN                                  \\ \hline
Kaur, Aggarwal et al.(2021)\cite{apl10}       & Parkinson’s disease classification                        & PPMI                                                              & Acc., Spec., Sens.                                       & DCGAN                                     \\ \hline
Guan, Chen et al.(2022)\cite{apl11}           & lesion detection                                          & CBIS-DDMS, Private dataset                                        & Prec., Recall, F1, AUC                                   & TMP-GAN                                   \\ \hline
Ahmad, Sun et al.(2022)\cite{apl12}           & Brain Tumor Classification                                & Private dataset                                                   & Acc., Sens., Spec.                                       & VAE-GAN                                   \\ \hline
Pombo, Gray et al.(2022)\cite{apl13}          & Equitable modelling of brain imaging                      & UK Biobank, OASIS                                                 & Acc., MSE, SSIM, MAE                                     & CounterSynth                              \\ \hline
Jiang, Chen et al.(2020)\cite{apl14}          & COVID-19 CT image segmentation                            & COVID-CT                                                          & FID, PSNR, SSIM, RMSE                                    & CGAN                                      \\ \hline
Qasim, Ezhov et al.(2020)\cite{apl15}         & Attacking class imbalance                                 & Brats2015, ISIC                                                   & Dice                                                     & Red-GAN                                   \\ \hline
Platscher, Zopes et al.(2020)\cite{apl16}     & Image translation                                         & Private dataset                                                   & Dice                                                     & Pix2Pix, SPADE, CycleGAN                  \\ \hline
Shi, Lu et al.(2020)\cite{apl17}              & robust pulmonary nodule segmentation                      & LIDC-IDRI                                                         & Dice, Pres., Sens.                                       & StyleGAN                                  \\ \hline
Shen, Ouyang et al.(2022)\cite{apl18}         & chest X-ray nodule augmentation and detection             & Private dataset                                                   & MAE, PSNR, SSIM, FID, AUC                                & DCGAN, GatedConv                          \\ \hline
Yurt, Lu et al.(2021)\cite{apl19}             & MR image synthesis                                        & IXI, ISLES                                                        & SSIM, PSNR                                               & mustGAN                                   \\ \hline
Yang, Lu et al.(2021)\cite{apl20}             & MR image synthesis                                        & Private data                                                      & PSNR, SSIM, MAE                                          & CAE-ACGAN                                 \\ \hline
Sikka, Skand et al.(2021)\cite{apl21}         & MRI to PET Cross-Modality Translation                     & ADNI                                                              & SSIM, PSNR, MAE, Acc., F1                                & GLA-GAN                                   \\ \hline
Amirrajab, Lorenz et al.(2022)\cite{apl22}    & Pathology Synthesis of 3D Consistent Cardiac MR Images    & ACDC                                                              & -                                                        & VAE-CGAN                                  \\ \hline
Chen, Gao et al.(2020)\cite{apl23}            & Cervical Cancer Single Cell Image Data Augmentation       & Pap-smear                                                         & FID                                                      & WGAN-GP                                   \\ \hline
Almezhghwi, Serte et al(2020)\cite{apl24}     & Classification of White Blood Cells                       & LISC                                                              & Acc.                                                     & Vanilla GAN                               \\ \hline
Teramoto, Tsukamoto et al.(2020)\cite{apl25}  & classification of lung cytological images                 & Private dataset                                                   & Acc., Sens., Spec.                                       & WGAN                                      \\ \hline
Murali, Lutnick et al.(2020)\cite{apl26}      & renal microanatomy                                        & Private dataset                                                   & FID, Qualitative assessment                              & Vanilla GAN                               \\ \hline
Quiros, Murray-Smith et al.(2020)\cite{apl27} & Learning a low dimensional manifold of real cancer tissue & Netherlands Cancer Institute, Vancouver General Hospital database & Qualitative assessment                                   & RGAN                                      \\ \hline
Mirzazadeh, Mohseni et al.(2021)\cite{apl28}  & Heart Transplant Rejection Classification                 & DNA-based transplant rejection, Children’s Hospital of Atlanta    & Mattthews correlation coefficient (MCC) in the classfier & WGAN-GP                                   \\ \hline
Zhao, Jin et al.(2021)\cite{apl29}            & White Blood Cell Classification                           & Private dataset                                                   & FID                                                      & WGAN                                      \\ \hline
Yu, Zhang et al.(2021)\cite{apl30}            & cervical cell classification model                        & Private dataset                                                   & Acc.                                                     & Vanilla GAN                               \\ \hline
Liu, Shuai et al.(2021)\cite{apl31}           & Cells image generation                                    & BCCD                                                              & IS                                                       & WGAN-GP                                   \\ \hline
Pandya, Patel et al.(2022)\cite{apl32}        & White Blood Cell Image Generation                         & Leukemia                                                          & -                                                        & Vanilla GAN                               \\ \hline
Kunzmann, Öttl et al.(2022)\cite{apl33}       & medical image annotation                                  & Asthma Equidae dataset                                            & Qualitative assessment                                   & Conditional Vanilla GAN                   \\ \hline
Dee, Ibrahim et al(2023)\cite{apl34}          & Histopathological Domain Adaptation                       & Tharun and Thompson dataset, Niki-TCGA                            & FID                                                      & Conditional WGAN-GP   \\ \hline   
\caption{GAN-based GDA methods in Medical image analysis}
\end{longtable}}

\subsection{GDA in Agricultural Image Analysis}
\subsubsection{GAN-based Methods}
Agricultural image analysis is also plagued by limited data, and the challenges of biological variability and environmental change in agricultural image analysis make it costly to collect and annotate such data. We will present in detail how GDA has been applied to medical and agricultural image analysis and the results obtained.

Different from the Medical Image Analysis area, GAN-based GDA networks are still the main methods in Agricultural Image Analysis as even the emergence of GANs in this area can be dated to recent years\cite{apl35}. Lu, Chen et al. reviewed the application of GANs in agriculture in September 2022. Our work will only focus on GAN-based GDA methods after that and diffusion-based works from 2020 to 2023.

\begin{figure}
    \centering
    \centerline{\includegraphics[width=1\columnwidth]{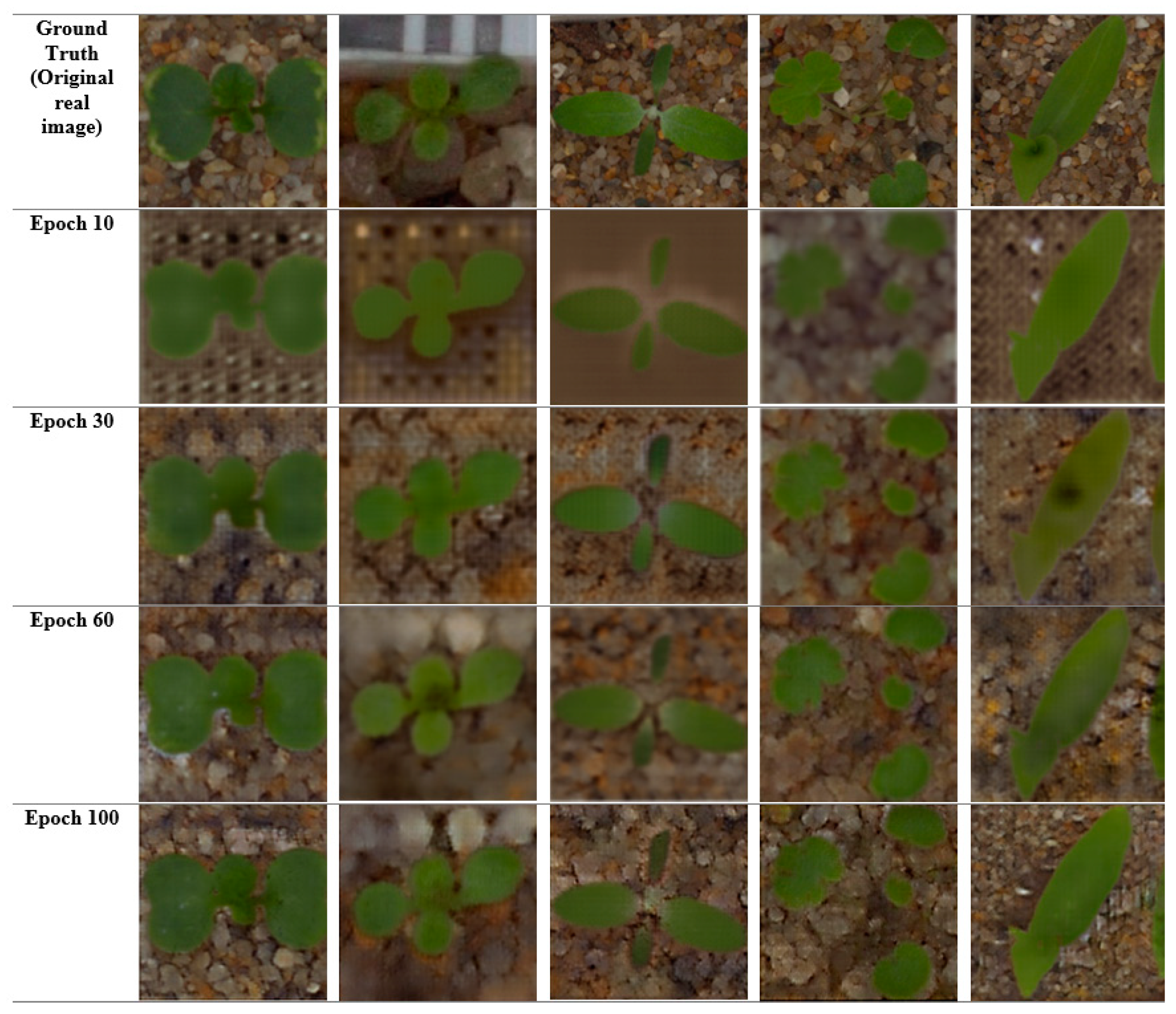}}
    \caption{The real and generated images for each class in Divyanth's work}
    \label{fig:enter-label}
\end{figure}

Divyanth, Guru et al. create artificial images of maize and four common weeds through conditional Generative Adversarial Networks\cite{apl36}. The real and generated images for each class are shown in Figure 15. Donald, Gabriel et al. use GAN to help the date fruit classification and accuracies of 100\%, 100\%, and 99\% were achieved for the hard, soft, and semi-hard classes\cite{apl37}. Bhakta, Phadikar et al. introduce both affine transformation and GAN to generate thermal images for agricultural disease prediction. GAN works better than traditional methods, achieving 97\% classification accuracy\cite{apl38}. Sharma, Tripathi et al. propose the WeedGAN for weed classification\cite{apl39}. The performance has surpassed DCGAN, WGAN, InfoGAN and VITGAN. A rice leaf disease image data augmentation based on a dual GAN has been designed by Zhang, Gao et al.\cite{apl40}. It is used for the recognition of three representative diseases. Haruna, Qin et al. adopt SG2-ADA in rice leaf plant disease detection\cite{apl46}. They generate high-quality images with fewer training datasets compared to other GAN methods. Deshpande, Patidar et al. reduce the issue of class imbalance in tomato plant disease detection with GAN\cite{apl41}. There is a CoffeeGAN designed for coffee plant disease recognition as well\cite{apl42}. Kulkarni, Deepa et al. generate images with segmented images with the Region of Interest. Plant disease recognition is a popular issue, Sharma, Kaur et al. implement three types of GAN(WGAN, DCGAN and Vanilla GAN) in their study\cite{apl43}. Khulud, Reem et al. improve pear disease classification with CycleGAN\cite{apl44}. They balance the number of records in each class. Jamadar, Sharma et al. combine GAN with transfer learning for pomegranate plant disease detection\cite{apl45}. 

We summarise the above papers that use generative data augmentation to do disease detection and classification and plot table Z to represent the increase they achieve in their tasks.

{\small
\begin{longtable}{|p{2.1cm}|p{2cm}|p{2cm}|p{1.5cm}|p{1.5cm}|p{2cm}|}
\hline
\multicolumn{1}{|p{2.5cm}|}{\textbf{Publication}} &
  \multicolumn{1}{c|}{\textbf{Task}} &
  \multicolumn{1}{c|}{\textbf{Dataset}} &
  \multicolumn{1}{p{1.5cm}|}{\textbf{Eval. Metrics}} &
  \multicolumn{1}{p{1.8cm}|}{\textbf{Arch.}} &
  \multicolumn{1}{c|}{\textbf{Increase}} \\ \hline
\endfirsthead
\endhead
\textbf{Divyanth, Guru et al.(2022)}\cite{apl36} &
  Crop/Weed classification &
  Giselsson et al.\cite{apl47} &
  F1 &
  cGAN &
  SVM: F1 0.93 to 0.96 LDA: F1 0.94 to 0.96 \\ \hline
\textbf{Donald, Gabriel et al.(2022)}\cite{apl37} &
  Date fruit classification &
  - &
  - &
  - &
  - \\ \hline
\textbf{Bhakta, Phadikar et al.(2022)}\cite{apl38} &
  Rice leaves recognition &
  Private dataset &
  Acc. &
  GAN &
  0.98 to 0.99 \\ \hline
\textbf{Sharma, Tripathi et al.(2022)}\cite{apl39} &
  Cotton weed identification &
  CottonWeed ID15 &
  FID, Acc., Recall, Spec. &
  WeedGAN &
  DenseNet121: Acc. 0.9568 to 0.9782 \\ \hline
\textbf{Zhang, Gao et al.(2023)}\cite{apl40} &
  Rice leaf disease detection &
  Private dataset &
  Acc., Prec., Recall, F1 &
  WGAN-GP and EESRGAN &
  ResNet18: Acc. 0.8708 to 0.9165 \\ \hline
\textbf{Haruna, Qin et al.(2023)}\cite{apl46} &
  Rice leaf disease detection &
  Mendeley Data\cite{apl48} &
  FID, KID, Prec., Recall &
  SG2-ADA &
  - \\ \hline
\textbf{Deshpande, Patidar et al.(2023)}\cite{apl41} &
  Tomato plant leaf disease detection &
  PlantVillage dataset\cite{apl49} &
  Acc., Recall, Prec., F1 &
  GAN &
  Acc. 0.9506 to 0.9881 \\ \hline
\textbf{Kulkarni, Deepa et al.(2023)}\cite{apl42} &
  Coffee plant disease detection &
  Private dataset &
  Acc., Prec., Recall, F1 &
  Coffee GAN &
  Acc. 0.920(CycleGAN) to 0.934(CoffeeGan) \\ \hline
\textbf{Sharma, Kaur et al.(2023)}\cite{apl43} &
  Mango plant disease detection &
  Private dataset &
  - &
  WGAN, DCGAN, Vanilla GAN &
  - \\ \hline
\textbf{Khulud, Reem et al.(2023)}\cite{apl44} &
  Pear disease classification &
  DiaMOS dataset\cite{apl50} &
  Acc., Prec., Recall, F1 &
  Cycle GAN &
  VGG19, ResNet50, EfficientNetB0: Acc. 7\% enhancement on average 
  \\ \hline
\textbf{Jamadar, Sharma et al.(2023)}\cite{apl45} &
  Pomegranate plant leaf disease detection &
  Private data &
  Acc. &
  GAN &
  DenseNet: Acc. 0.9627 to 0.9945 \\ \hline
\caption{GAN-based GDA in Agricultural image analysis}
\end{longtable}}

Plant disease detection and classification is the main task where GAN-based methods are used. However, the types of GANs used are similar and the tasks are narrow, though it seems that GANs have achieved success in many precise tasks. We are excited to find some works that break out the research boundaries of GDA in agriculture. Tai, Wang et al. use Time-Series GAN to synthesize sensing data for pest incidence forecasting\cite{apl51}. They reduce the dimensionality of the dataset from 22 to 2 with the PCA and t-SNE to visualize the data distribution. The visualization is shown in Figure 16. Another use for Gan-based GDA in agriculture is number assessment. Salvador, Javier et al. apply evolutionary conditional GANs to generate grape berry cluster images. It is critical in the grape and wine industries\cite{apl53} \cite{apl54}. Their model indeed generates a better dataset than the original one(Table 6).

\begin{table}[h]
\centering
\begin{tabular}{lllll}
\hline
\textbf{Model} & \textbf{Training} & \textbf{Validation} & \multicolumn{2}{l}{\textbf{Test}} \\ \cline{4-5} 
          & \textbf{RMSE} & \textbf{RMSE} & \textbf{RMSE} & \textbf{R²} \\
Original  & 9.781         & 64.739        & 56.515        & 0.65        \\
Augmented & 7.674         & 40.169        & 38.774        & 0.75        \\ \hline
\end{tabular}
\caption{Metrics for the deep learning modeling and testing on external dataset. RMSE: (lower is better).}
\label{tab:my-table}
\end{table}

\begin{figure}
    \centering
    \centerline{\includegraphics[width=1\columnwidth]{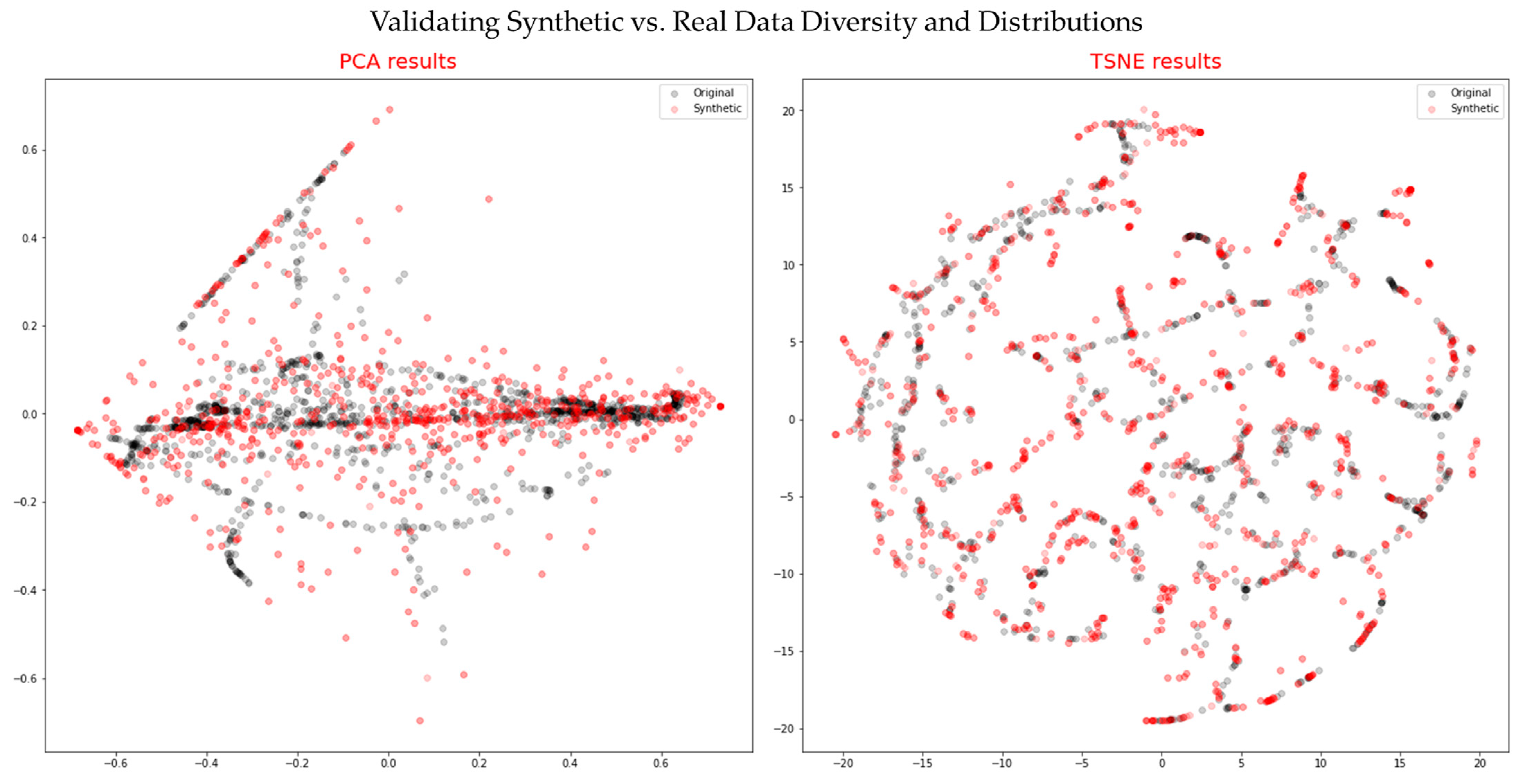}}
    \caption{Reduction of the dimensionality for the work of Tai, Wang et al.}
    \label{fig:enter-label}
\end{figure}

\subsubsection{Diffusion-based Methods}

Diffusion-based GDA in agriculture is more sparsely used compared to GAN-based GDA. Even in the previous section, we showed that GAN-based GDA still has a lot of room for development. We find only a very limited number of relevant papers. Trabucco, Doherty et al. use the Leafy Spurge in the training of stable diffusion for the first time\cite{apl55}. Their work will directly benefit efforts to restore natural ecosystems. Weed recognition served as a difficult and valuable task in Agricultural Image Analysis, diffusion-based methods have achieved better performance than GAN-based GDA methods used in this area. Diffusion models achieved the best trade-off between sample fidelity and diversity and obtained the highest Fréchet Inception Distance, compared to GANs\cite{apl56}. Dong, Xinda et al. apply diffusion models for generating weed images to enhance weed identification for the first time. The dataset expanded with synthetic weed images improved (up to 1.20\%, 1.23\%, and 2.30\% on the accuracy, precision, and recall, respectively) weed classification accuracy by four deep learning models (i.e., VGG16, Inception-v3, Inception-v3, and ResNet50)\cite{apl58}. Moreno, Gómez et al. pave the way for applying Stable Diffusion in various aspects of precision agriculture beyond crop/weed detection\cite{apl57}. They increase the performance of detection models by 6\% - 9\%. Nikolaos, Elleftheria et al. propose a Generate-Paste-Blend-Detect method[59]. They achieve a mean average precision of 0.66 for whiteflies detection with the YOLOv8 detection model.

\subsection{GDA in Signal processing}
\subsubsection{GDA in Sound Processing}
Sound signal processing is another undeveloped area for GDA. Abayomi-Alli, Damaševičius et al. did a systematic review of data augmentation and deep learning methods in sound classification in November 2022\cite{apl70}. Their work shows that there is still a lag in the application of advanced methods for generating synthetic sound data. Bakır, Çayır et al. made a comprehensive experimental study for data augmentation on voice classification in August 2023\cite{apl71}, but traditional transformation methods are their focus. As a result, we decided to have a review of the progress of GDA methods in sound classification. We focus mainly on works published after the review in 2022.

Diffusion-based methods are becoming the new trend in GDA for sound processing. Chen, Yan et al. propose an environmental sound classification (ESC) augmentation technique based on the diffusion probabilistic model (DPM) with DPM-Solver ++ for fast sampling\cite{apl72}. A Top-k selection technique is proposed to select high-quality samples. They mark around 6.3\% and 7.6\% improvements with Inception-v3 and ResNet-50 respectively compared with the baseline models without data augmentation. Pascual, Bhattacharya et al. design the Diffusion Audio Generator based on score-matching with variance exploding diffusion\cite{apl73}. They achieve a better performance on UrbanSound8K and TUT datasets(Table 7). These works are impressive, but the GAN-based methods still account for the mainstream. We make a list of the GAN-based GDA methods used in sound processing(Table 8).
\begin{table}[t]
\begin{tabular}{p{2cm}rcllll}
\hline
\multirow{2}{*}{\textbf{MODEL}} &
  \multirow{2}{*}{\textbf{Samp. rate}} &
  \multirow{2}{*}{\textbf{Num.param.}} &
  \multicolumn{2}{l}{\textbf{US8K}} &
  \multicolumn{2}{l}{\textbf{TUT}} \\ \cline{4-7} 
                        &       &      & \textbf{FD↓} & \textbf{LS↑} & \textbf{FD↓} & \textbf{LS↑} \\ \hline
SampleRNN               & 22kHz & 41M  & 237.0        & 2.80         & 138.8        & 2.07         \\
PixelSNAIL              & 22kHz & 119M & 115.1        & 3.27         & 107.3        & 4.65         \\
PixelSNAILw or TUT-HifiGAN & 22kHz & 119M & n/a          & n/a          & 114.3        & 3.33         \\ \hline
DAG48 and Downsample        & 22kHz & 22M  & 91.4         & 5.54         & 78.8         & 5.13         \\
DAG22                   & 22kHz & 22M  & 87.4         & 5.45         & 64.8         & 5.74         \\
DAG48                   & 48kHz & 22M  & 47.9         & 6.21         & 37.7         & 6.59         \\ \hline
Real data               & 22kHz & -    & 72.6         & 5.26         & 79.0         & 5.53         \\
Real data(test set)     & 48kHz & -    & 30.9         & 6.33         & 3.7          & 9.31         \\ \hline
\end{tabular}
\caption{Performance for Pascual's work}
\label{tab:my-table}
\end{table}

{\small
\begin{longtable}{|p{2cm}|p{3cm}|p{2cm}|p{2cm}|p{2cm}|}
\hline
\multicolumn{1}{|c|}{\textbf{Publication}} &
  \multicolumn{1}{c|}{\textbf{Task}} &
  \multicolumn{1}{c|}{\textbf{Dataset}} &
  \multicolumn{1}{c|}{\textbf{Architecture}} &
  \multicolumn{1}{c|}{\textbf{Eval. Metrics}} \\ \hline
\endfirsthead
\endhead
Margaryan, Seibold et al.(2022)\cite{apl74} &
  Surgical actions sound classification &
  Dataset of Mane et al. &
  WGAN-GP &
  FID, F1 \\ \hline
Jang, Kim et al.(2023)\cite{apl75} &
  Urban sound classification &
  Urban Sound8K &
  Mel-GAN &
  Prec., Recall, F1 \\ \hline
Madhu A, K S. (2022)\cite{apl79} &
  Environment  sound classification &
  ESC-10, UrbanSound8K, TUT &
  EnvGAN &
  FID, F1, Acc. et al. \\ \hline
Kim, Moon et al.(2023)\cite{apl76} &
  Animal sound classification &
  NA birds, SK frogs &
  DualDisc WaveGAN &
  FID, NDB \\ \hline
Takezaki, Kishida et al.(2022)\cite{apl78} &
  Heart sound classification &
  PCG &
  Synthetic Spectrogram based GANs &
  Acc., Sens., Spec., p-value \\ \hline
Saldanha, Chakraborty et al.(2022)\cite{apl77} &
  Respiratory disease classification &
  ICBHI &
  MLP-VAE, Conv VAE, conditional VAE &
  Spec., Sens., Prec., F1 \\ \hline
\caption{GAN-based Methods used in sound classification}
\label{tab:my-table}\\
\end{longtable}}

Margaryan, Seibold et al.\cite{apl74} obtain a relative improvement of 2.84\% in F1-Score by applying their data augmentation methods to the public THA sounds dataset. Takezaki, Kishida et al.\cite{apl78} get the classification accuracy for heart sound improved by 1\% with Synthetic Spectrogram-based GANs(SSG). But they get a better score with Window Slicing with Spectrogram, approximately 1\% higher than the score achieved by SSG. Jang, Kim et al.\cite{apl75} and Madhu A, K S.\cite{apl79} all apply their GAN-based methods to the classification of UrbanSound8K. Jang, Kim et al. increase the F1 score by 3.5\% and Madhu A, K S. achieve a better performance by increasing F1 score by 6.4\%. Chen, Yan et al. achieve 6.3\% and 7.6\% improvements with Inception-v3 and ResNet-50 respectively by diffusion-based GDA. They base their work on UrbanSound8K as well. It's arbitrary to conclude that diffusion-based GDA methods perform better than GAN-based methods since there are no direct comparisons between these two types of methods in the same tasks with the same evaluation metrics. However, it is undisputed that diffusion-based methods are more stable during training. We hope that further work could explore the ability of diffusion-based methods not only in sound processing, but in all signal processing.

\section{A Unified Framework for GDA}

\begin{figure}
    \centering
    \centerline{\includegraphics[width=1.15\columnwidth]{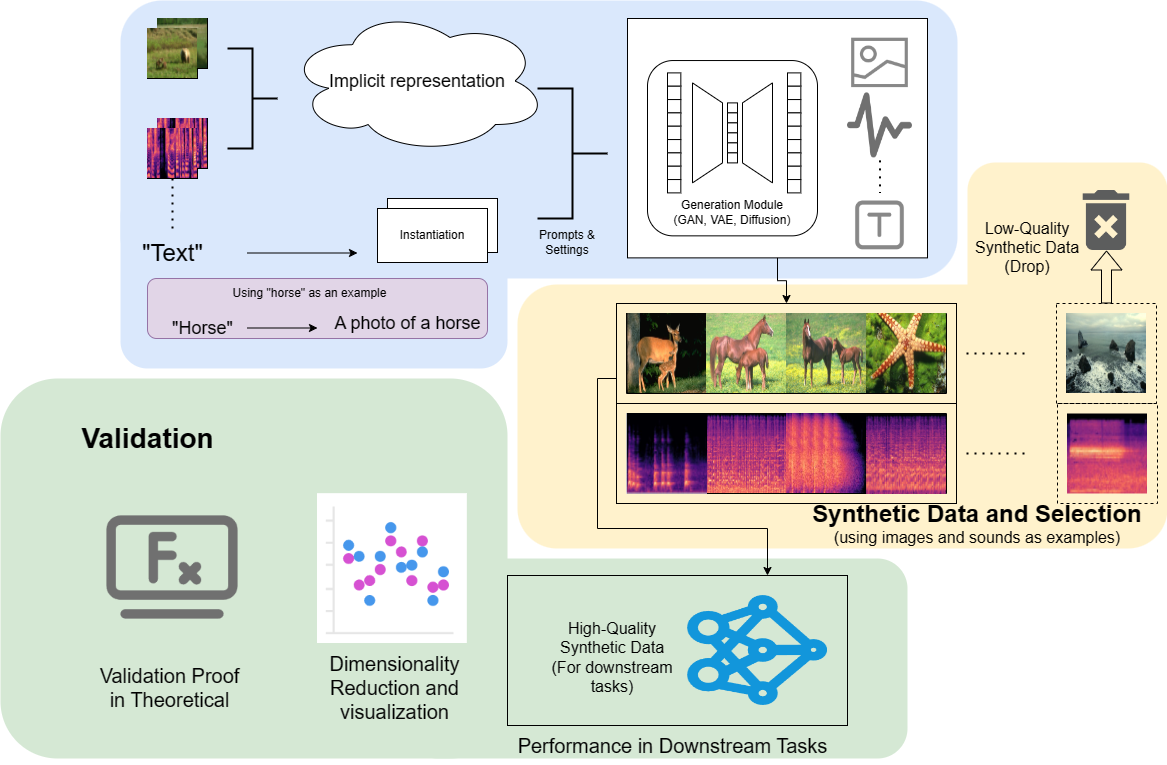}}
    \caption{A schematic representation of the proposed Unified Framework for Generative Data Augmentation (GDA). This structure synthesizes the multifaceted elements of GDA into a systematic approach, emphasizing model choice, utilization techniques, data selection, validation, and practical applications. Designed to address the current research gaps and lack of comprehensive benchmarking, the framework serves as a guiding beacon for both present applications and future advancements in GDA.}
    \label{fig:unified}
\end{figure}

The proliferation of generative models and their application in various domains has necessitated a structured and systematic approach towards Generative Data Augmentation (GDA). While much of the existing research has tended towards applying generative models to specific datasets with a dearth of innovation, a salient gap remains: the lack of a comprehensive benchmark for GDA. Addressing this, we propose a unified framework, rooted in systematic processes, that not only provides clarity in implementation but also paves the way for future benchmarking. The importance of such a structured approach cannot be overstated, as it stands to substantially benefit subsequent advancements in GDA. The unified structure is shown in Figure \ref{fig:unified}.

\subsection{Choice of Generative Models}
The landscape of Generative Data Augmentation (GDA) is rapidly evolving, and while our unified framework provides a comprehensive overview of the current state of the art, the horizon of GDA is filled with opportunities and challenges that demand exploration. Based on our analysis and the gaps identified in the current literature, we outline the following future directions.

\subsection{Utilization of Generative Models}
Transitioning from choice to application, this module is dedicated to harnessing the potential of the chosen generative model. It delves into sophisticated techniques, such as prompt engineering and latent space manipulation, ensuring that data generation is both purposeful and controlled, aligned with the augmentation objectives.

\subsection{Selection of Augmented Data}
Quality trumps quantity in the realm of synthetic samples. Recognizing this, this module offers a roadmap to discern and select the most semantically coherent and high-quality samples from the ocean of generated data. With tools like discriminators and clustering at the helm, it ensures that only the most pertinent data augments the original dataset.

\subsection{Validation of Augmentation}
Trust but verify—before integrating synthetic data into real-world applications, its veracity and utility demand rigorous validation. This module furnishes a spectrum of tools, spanning empirical metrics, visual assessments, and theoretical analyses, which together authenticate the value proposition of the generated data, ensuring it serves to bolster model generalization.

\subsection{Applications of GDA}
Synthesizing the theory with practice, this module explores the multifaceted domains where GDA shines. From the intricate patterns in computer vision to the nuances of computational biology and the intricacies of Natural Language Processing (NLP), it underscores the transformative potential of GDA across diverse arenas.

In conclusion, our unified framework serves as a beacon, illuminating the expansive world of GDA. By systematizing the rich literature, highlighting existing gaps, and offering a structured foundation, we believe that this framework stands poised to foster cohesive development in GDA. In doing so, it paves the way for benchmarking and standardized assessment, catalyzing the evolution and progress in data augmentation methodologies.

\section{Future Directions in Generative Data Augmentation}

The landscape of Generative Data Augmentation (GDA) is rapidly evolving, and while our unified framework provides a comprehensive overview of the current state of the art, the horizon of GDA is filled with opportunities and challenges that demand exploration. Based on our analysis and the gaps identified in the current literature, we outline the following future directions that promise to shape the next frontier of GDA:

\subsection{Advanced Techniques for Utilizing Generative Models}
Generative models, though powerful, often require a nudge in the right direction to yield the desired outputs. Techniques like prompt engineering have showcased potential in guiding models, but the surface has merely been scratched. The future could witness the emergence of innovative prompt engineering strategies, alongside novel methods for latent space manipulation, that could offer unprecedented control over data generation. These advancements would empower users to generate data that is more aligned with their specific requirements, ensuring that the augmented data complements the original dataset seamlessly.

\subsection{Effective Methods for Synthetic Data Selection}
While our framework touches upon the significance of cherry-picking high-quality synthetic data, the methodologies dedicated to this purpose remain relatively nascent. Future research should focus on devising robust and scalable techniques that can sift through massive volumes of generated data to identify and retain only those samples that truly enhance the dataset's diversity and richness. Automated, adaptive methods could significantly expedite this process, making GDA more efficient and impactful.

\subsection{Development for Generalization Bound with Large-Scale Models}
One of the primary challenges in applying GDA lies in understanding and quantifying the generalization capabilities of trained models using synthetic photos generated by generative models in GDA, especially when scaled up. Large-scale models, like Stable Diffusion, promise impressive results but also introduce complexities in ensuring consistent performance across diverse datasets. There is a pressing need to establish theoretical bounds for generalization when employing these behemoths in the GDA landscape. Such bounds would not only enhance our understanding of the model's behavior but also guide practitioners in model selection and tuning for specific augmentation tasks.

\subsection{Establishing a Benchmark for Generative Data Augmentation}
Perhaps one of the most pressing needs, as underscored in our unified framework, is the establishment of a comprehensive benchmark for GDA. The current state of GDA research, albeit rich, lacks a standardized yardstick for performance assessment. A universally accepted benchmark would facilitate comparisons across different GDA techniques, fostering a more collaborative and cohesive research environment. Moreover, it would enable researchers and practitioners to evaluate and select methods best suited for their tasks objectively.

In summation, the vistas of Generative Data Augmentation are vast and verdant, beckoning researchers to traverse their depths. As we stand on the cusp of new discoveries and innovations, these future directions offer a guiding light, illuminating pathways that promise not just incremental enhancements but transformative breakthroughs. Building upon our unified framework and the foundational literature, the onus is now on the global research community to embark on these exciting journeys, pushing the frontiers of GDA to new zeniths.

\section{Conclusion}
This thesis presented a comprehensive survey and analysis of generative data augmentation, an emerging technique to address data scarcity in machine learning. We proposed a unified framework that systematically organizes the extensive GDA literature based on five modules - generative model selection, model utilization strategies, data selection methodologies, validation approaches, and applications. Their empirical studies and theoretical analyses demonstrate GDA's capabilities in enhancing model generalization. However, open challenges remain, including establishing universal benchmarks for standardized assessment and developing techniques tailored to limited data regimes. Promising future directions involve new techniques for utilizing generative models, effective data selection, theoretical development for large-scale models' application in GDA and establishing a benchmark for GDA. As generative models continue to advance alongside increasing computational power, GDA is poised to transform data augmentation. By laying a structured foundation, this thesis aims to nurture more cohesive development in this vital arena and accelerate progress towards more data-efficient and generalizable machine learning.



\clearpage

\begin{thebibliography}{00}


\bibitem{1_deep_learning}Ye, Mang et al. “Deep Learning for Person Re-Identification: A Survey and Outlook.” IEEE Transactions on Pattern Analysis and Machine Intelligence 44 (2020): 2872-2893.

\bibitem{2_deep_learning}Ekhande, Sonali et al. “Review on effectiveness of deep learning approach in digital forensics.” International Journal of Electrical and Computer Engineering (IJECE) (2022): n. pag.

\bibitem{3_deep_learning}Abedin, Mohammad Zoynul et al. “Deep learning-based exchange rate prediction during the COVID-19 pandemic.” Annals of Operations Research (2021): 1 - 52.

\bibitem{4_high_data}Chen, Yunhao et al. “Effective Audio Classification Network Based on Paired Inverse Pyramid Structure and Dense MLP Block.” ArXiv abs/2211.02940 (2022): n. pag.

\bibitem{5_high_data}Xuzhen He et al. "Deep learning for efficient stochastic analysis with spatial variability." Acta Geotechnica, 17 (2021): 1031 - 1051. https://doi.org/10.1007/s11440-021-01335-1.

\bibitem{6_high_data}Wan Zhu et al. "The Application of Deep Learning in Cancer Prognosis Prediction." Cancers, 12 (2020). https://doi.org/10.3390/cancers12030603.

\bibitem{7_privacy}Ding, X., Hongbiao, F., Zhang, Z., Choo, K., \& Jin, H. (2020). Privacy-Preserving Feature Extraction via Adversarial Training. IEEE Transactions on Knowledge and Data Engineering, 34, 1967-1979. https://doi.org/10.1109/tkde.2020.2997604.

\bibitem{8_privacy}Wei, W., \& Liu, L. (2021). Gradient Leakage Attack Resilient Deep Learning. IEEE Transactions on Information Forensics and Security, 17, 303-316. https://doi.org/10.1109/tifs.2021.3139777.

\bibitem{9_labelling_diffusion}Chen, Yunhao et al. “Data Augmentation for Environmental Sound Classification Using Diffusion Probabilistic Model with Top-k Selection Discriminator.” ArXiv abs/2303.15161 (2023): n. pag.

\bibitem{10_data_augmentation}Shorten, Connor and Taghi M. Khoshgoftaar. “A survey on Image Data Augmentation for Deep Learning.” Journal of Big Data 6 (2019): 1-48.

\bibitem{14_data_augmentation}Pei Liu et al. "A Survey of Text Data Augmentation." 2020 International Conference on Computer Communication and Network Security (CCNS) (2020): 191-195. https://doi.org/10.1109/CCNS50731.2020.00049.

\bibitem{15_data_augmentation}Ruoqi Shen et al. "Data Augmentation as Feature Manipulation: a story of desert cows and grass cows." ArXiv, abs/2203.01572 (2022). https://doi.org/10.48550/arXiv.2203.01572.

\bibitem{11_data_distribution}Karakas, C.E., Dirik, A., Yalçınkaya, E., Yanardag, P. (2022). FairStyle: Debiasing StyleGAN2 with Style Channel Manipulations. In: Avidan, S., Brostow, G., Cissé, M., Farinella, G.M., Hassner, T. (eds) Computer Vision – ECCV 2022. ECCV 2022. Lecture Notes in Computer Science, vol 13673. Springer, Cham. https://doi.org/10.1007/978-3-031-19778-9\_33

\bibitem{12_data_distribution}E. Pajouheshgar, T. Zhang and S. Süsstrunk, "Optimizing Latent Space Directions for Gan-Based Local Image Editing," ICASSP 2022 - 2022 IEEE International Conference on Acoustics, Speech and Signal Processing (ICASSP), Singapore, Singapore, 2022, pp. 1740-1744, doi: 10.1109/ICASSP43922.2022.9747326.

\bibitem{13_data_distribution}Fried, Daniel et al. “InCoder: A Generative Model for Code Infilling and Synthesis.” International Conference on Learning Representations (2022): n. pag.

\bibitem{16_VAE}Takida, Y., Shibuya, T., Liao, W., Lai, C., Ohmura, J., Uesaka, T., Murata, N., Takahashi, S., Kumakura, T. \&amp; Mitsufuji, Y..  SQ-VAE: Variational Bayes on Discrete Representation with Self-annealed Stochastic Quantization. <i>Proceedings of the 39th International Conference on Machine Learning</i>, in <i>Proceedings of Machine Learning Research</i> 162:20987-21012 (2022).

\bibitem{17_VAE}Ma, Changsheng and Xiangliang Zhang. “GF-VAE: A Flow-based Variational Autoencoder for Molecule Generation.” Proceedings of the 30th ACM International Conference on Information \& Knowledge Management (2021): n. pag.

\bibitem{18_VAE}Vahdat, Arash and Jan Kautz. “NVAE: A Deep Hierarchical Variational Autoencoder.”Neural Information Processing Systems (2020): n. pag.

\bibitem{19_VAE}Shao, Huajie et al. “ControlVAE: Controllable Variational Autoencoder.” International Conference on Machine Learning (2020).

\bibitem{20_VAE}Ding, Zheng et al. “Guided Variational Autoencoder for Disentanglement Learning.” 2020 IEEE/CVF Conference on Computer Vision and Pattern Recognition (CVPR) (2020): 7917-7926.

\bibitem{21_VAE}Kingma, Diederik P. and Max Welling. “Auto-Encoding Variational Bayes.” CoRR abs/1312.6114 (2013): n. pag.

\bibitem{22_GAN}Brock, Andrew et al. “Large Scale GAN Training for High Fidelity Natural Image Synthesis.” International Conference on Learning Representations (2018): n. pag.

\bibitem{23_GAN}Arjovsky, Martín et al. “Wasserstein GAN.” ArXiv abs/1701.07875 (2017): n. pag.

\bibitem{24_GAN}Yang, Tao et al. “GAN Prior Embedded Network for Blind Face Restoration in the Wild.” 2021 IEEE/CVF Conference on Computer Vision and Pattern Recognition (CVPR) (2021): 672-681.

\bibitem{25_GAN}Liu, Hongyu et al. “PD-GAN: Probabilistic Diverse GAN for Image Inpainting.” 2021 IEEE/CVF Conference on Computer Vision and Pattern Recognition (CVPR) (2021): 9367-9376.

\bibitem{26_GAN}Chan, Eric et al. “pi-GAN: Periodic Implicit Generative Adversarial Networks for 3D-Aware Image Synthesis.” 2021 IEEE/CVF Conference on Computer Vision and Pattern Recognition (CVPR) (2020): 5795-5805.

\bibitem{27_GAN}Tran, Ngoc-Trung et al. “On Data Augmentation for GAN Training.” IEEE Transactions on Image Processing 30 (2020): 1882-1897.

\bibitem{28_GAN}Härkönen, Erik et al. “GANSpace: Discovering Interpretable GAN Controls.” Neural Information Processing Systems (2020): n. pag.

\bibitem{29_GAN}Karras, Tero et al. “Training Generative Adversarial Networks with Limited Data.”  Neural Information Processing Systems (2020): n. pag.

\bibitem{30_GAN}Goodfellow, Ian J. et al. “Generative Adversarial Networks.” (2014).

\bibitem{31_GPT}Radford, Alec and Karthik Narasimhan. “Improving Language Understanding by Generative Pre-Training.” (2018).

\bibitem{32_GPT}Radford, Alec et al. “Language Models are Unsupervised Multitask Learners.” (2019).

\bibitem{33_GPT}Brown, Tom B. et al. “Language Models are Few-Shot Learners.” Neural Information Processing Systems (2020): n. pag.

\bibitem{34_GPT}Bang, Yejin et al. “A Multitask, Multilingual, Multimodal Evaluation of ChatGPT on Reasoning, Hallucination, and Interactivity.” ArXiv abs/2302.04023 (2023): n. pag.

\bibitem{35_GPT}Guo, Biyang et al. “How Close is ChatGPT to Human Experts? Comparison Corpus, Evaluation, and Detection.” ArXiv abs/2301.07597 (2023): n. pag.

\bibitem{36_GPT}Shen, Yongliang et al. “HuggingGPT: Solving AI Tasks with ChatGPT and its Friends in Hugging Face.” ArXiv abs/2303.17580 (2023): n. pag.

\bibitem{37_GPT}Pan, Shirui et al. “Unifying Large Language Models and Knowledge Graphs: A Roadmap.” ArXiv abs/2306.08302 (2023): n. pag.

\bibitem{38_GPT}OpenAI. “GPT-4 Technical Report.” ArXiv abs/2303.08774 (2023): n. pag.

\bibitem{39_GPT}Bubeck, Sébastien et al. “Sparks of Artificial General Intelligence: Early experiments with GPT-4.” ArXiv abs/2303.12712 (2023): n. pag.

\bibitem{40_GPT}Peng, Baolin et al. “Instruction Tuning with GPT-4.” ArXiv abs/2304.03277 (2023): n. pag.

\bibitem{41_Diffusion}Dhariwal, Prafulla and Alex Nichol. “Diffusion Models Beat GANs on Image Synthesis.” ArXiv abs/2105.05233 (2021): n. pag.

\bibitem{42_Diffusion}Ho, A. Jain, and P. Abbeel. Denoising diffusion probabilistic models. Advances in
Neural Information Processing Systems, 33:6840–6851, 2020
\bibitem{43_Diffusion}Song, Jiaming, Chenlin Meng and Stefano Ermon. “Denoising Diffusion Implicit
Models.” ArXiv abs/2010.02502 (2020): n. pag

\bibitem{44_Diffusion}Saharia, Chitwan, William Chan, Saurabh Saxena, Lala Li, Jay Whang, Emily L.
Denton, Seyed Kamyar Seyed Ghasemipour, Burcu Karagol Ayan, Seyedeh Sara
Mahdavi, Raphael Gontijo Lopes, Tim Salimans, Jonathan Ho, David J. Fleet and
Mohammad Norouzi. “Photorealistic Text-to-Image Diffusion Models with Deep
Language Understanding.” ArXiv abs/2205.11487 (2022): n. pag

\bibitem{45_Diffusion}onathan Ho \& Tim Salimans. “Classifier-Free Diffusion Guidance." Neural Information Processing Systems 2021
Workshop on Deep Generative Models and Downstream Applications

\bibitem{46_Diffusion}Sohl-Dickstein, Jascha Narain et al. “Deep Unsupervised Learning using Nonequilibrium Thermodynamics.” International Conference on Machine Learning: n. pag.

\bibitem{47_Diffusion}Yang, Ling et al. “Diffusion Models: A Comprehensive Survey of Methods and Applications.” ArXiv abs/2209.00796 (2022): n. pag.

\bibitem{48_Diffusion_LAION5B}Schuhmann, Christoph et al. “LAION-5B: An open large-scale dataset for training next generation image-text models.” ArXiv abs/2210.08402 (2022): n. pag.

\bibitem{49_TDA}Zhao, Tong et al. “Graph Data Augmentation for Graph Machine Learning: A Survey.” ArXiv abs/2202.08871 (2022): n. pag.
\bibitem{50_TDA}Fu, Biying et al. “Data augmentation for time series: traditional vs generative models on capacitive proximity time series.” Proceedings of the 13th ACM International Conference on PErvasive Technologies Related to Assistive Environments (2020): n. pag.
\bibitem{51_TDA}Chlap, Phillip et al. “A review of medical image data augmentation techniques for deep learning applications.” Journal of Medical Imaging and Radiation Oncology 65 (2021): n. pag.
\bibitem{52_TDA}Yang, Benyi et al. “Mask2Defect: A Prior Knowledge-Based Data Augmentation Method for Metal Surface Defect Inspection.” IEEE Transactions on Industrial Informatics 18 (2022): 6743-6755.
\bibitem{53_TDA}Athalye, Chinmayee and Rima Arnaout. “Domain-guided data augmentation for deep learning on medical imaging.” PLOS ONE 18 (2022): n. pag.
\bibitem{54_TDA}Shorten, Connor and Taghi M. Khoshgoftaar. “A survey on Image Data Augmentation for Deep Learning.” Journal of Big Data 6 (2019): 1-48.
\bibitem{55_TDA}Feng, Steven Y. et al. “A Survey of Data Augmentation Approaches for NLP.” Findings (2021).
\bibitem{56_TDA}Morris, John X. et al. “TextAttack: A Framework for Adversarial Attacks, Data Augmentation, and Adversarial Training in NLP.” Conference on Empirical Methods in Natural Language Processing (2020).
\bibitem{57_TDA}Shi, Freda et al. “Substructure Substitution: Structured Data Augmentation for NLP.” ArXiv abs/2101.00411 (2021): n. pag.
\bibitem{58_TDA}Lam, Tsz Kin et al. “Sample, Translate, Recombine: Leveraging Audio Alignments for Data Augmentation in End-to-end Speech Translation.” Annual Meeting of the Association for Computational Linguistics (2022).
\bibitem{59_TDA}Muthumari, Muthumari et al. “Data Augmentation Model for Audio Signal Extraction.” 2022 3rd International Conference on Electronics and Sustainable Communication Systems (ICESC) (2022): 334-340.

\bibitem{60_Theory}Zheng, Chenyu, et al. "Toward Understanding Generative Data Augmentation." ArXiv, 2023, /abs/2305.17476. 

\bibitem{61_synthetic_data}Vanherle, Bram et al. “Analysis of Training Object Detection Models with Synthetic Data.” British Machine Vision Conference (2022).
\bibitem{62_synthetic_data}Jordon, James et al. “Synthetic Data - what, why and how?” ArXiv abs/2205.03257 (2022): n. pag.
\bibitem{63_synthetic_data}Figueira, Alvaro, and Bruno Vaz. “Survey on Synthetic Data Generation, Evaluation Methods and GANs.” Mathematics, vol. 10, no. 15, Aug. 2022, p. 2733. Crossref, https://doi.org/10.3390/math10152733.
\bibitem{64_synthetic_data}Figueira, Alvaro, and Bruno Vaz. “Survey on Synthetic Data Generation, Evaluation Methods and GANs.” Mathematics, vol. 10, no. 15, Aug. 2022, p. 2733. Crossref, https://doi.org/10.3390/math10152733.
\bibitem{65_synthetic_data}Emam, K.; Mosquera, L.; Hoptroff, R. Chapter 1: Introducing Synthetic Data Generation. In Practical Synthetic Data Generation:
Balancing Privacy and the Broad Availability of Data; O’Reilly Media, Inc.: Sebastopol, CA, USA, 2020; pp. 1–22.
\bibitem{66_synthetic_data}Chen, Richard J. et al. “Synthetic data in machine learning for medicine and healthcare.” Nature Biomedical Engineering 5 (2021): 493 - 497.
\bibitem{67_synthetic_data}McDuff, Daniel J. et al. “Synthetic Data in Healthcare.” ArXiv abs/2304.03243 (2023): n. pag.
\bibitem{68_synthetic_data}Arora, Anmol and Ananya Arora. “Machine learning models trained on synthetic datasets of multiple sample sizes for the use of predicting blood pressure from clinical data in a national dataset.” PLOS ONE 18 (2023): n. pag. 

\bibitem{69_synthetic_data_aerial}Kiefer, Benjamin et al. “Leveraging Synthetic Data in Object Detection on Unmanned Aerial Vehicles.” 2022 26th International Conference on Pattern Recognition (ICPR) (2021): 3564-3571.

\bibitem{70_synthetic_data_game}Rasmussen, Ingeborg et al. “Development of a Novel Object Detection System Based on Synthetic Data Generated from Unreal Game Engine.” Applied Sciences (2022): n. pag.

\bibitem{71_synthetic_data_physics}Yang, Qinqin et al. “Physics-Driven Synthetic Data Learning for Biomedical Magnetic Resonance: The imaging physics-based data synthesis paradigm for artificial intelligence.” IEEE Signal Processing Magazine 40 (2022): 129-140.

\bibitem{72_grad_theft_auto}R. Games. Grand theft auto v. [Online]. Available: https://www.rockstargames.com/de/games/V
\bibitem{73}Diffusion Models vs. GANs vs. VAEs: Comparison of Deep Generative Models Ainur Gainetdinov https://towardsai.net/p/machine-learning/diffusion-models-vs-gans-vs-vaes-comparison-of-deep-generative-models
\bibitem{74}Kebaili, Aghiles et al. “Deep Learning Approaches for Data Augmentation in Medical Imaging: A Review.” Journal of Imaging 9 (2023): n. pag.
\bibitem{75_KL}Wibisono, Andre and Kaylee Yingxi Yang. “Convergence in KL Divergence of the Inexact Langevin Algorithm with Application to Score-based Generative Models.” ArXiv abs/2211.01512 (2022): n. pag.
\bibitem{76_GAN_review}Yi, Xin et al. “Generative Adversarial Network in Medical Imaging: A Review.” Medical image analysis 58 (2018): 101552 .

\bibitem{77_diffusion} Song, Yang and Stefano Ermon. “Generative Modeling by Estimating Gradients of the Data Distribution.” Neural Information Processing Systems (2019).

\bibitem{78_score_based}Song, Yang et al. “Score-Based Generative Modeling through Stochastic Differential Equations.” International Conference on Learning Representation (2020): n. pag.

\bibitem{79_sliced_score}Song, Yang et al. “Sliced Score Matching: A Scalable Approach to Density and Score Estimation.” Conference on Uncertainty in Artificial Intelligence (2019).

\bibitem{80_score_matching}Vincent, Pascal. “A Connection Between Score Matching and Denoising Autoencoders.” Neural Computation 23 (2011): 1661-1674.

\bibitem{81_dynamics}Giorgio Parisi. Correlation functions and computer simulations. Nuclear Physics B , 180(3):378–384, 1981.
\bibitem{82_dynamics}Grenander, Ulf and Michael I. Miller. “REPRESENTATIONS OF KNOWLEDGE IN COMPLEX SYSTEMS.” Journal of the royal statistical society series b-methodological 56 (1994): 549-581.

\bibitem{83_dynamics}Yang Song and Stefano Ermon. Improved techniques for training score-based generative models. Advances in neural
information processing systems, n.pag(2020).

\bibitem{84_diffusion_survey}Kazerouni, Amirhossein et al. “Diffusion models in medical imaging: A comprehensive survey.” Medical image analysis 88 (2022): 102846 .

\bibitem{85_transformer} Vaswani, Ashish, et al. "Attention Is All You Need." ArXiv, /abs/1706.03762. (2017).


\bibitem{86_seq_atten}Parikh, Ankur P. et al. “A Decomposable Attention Model for Natural Language Inference.” ArXiv abs/1606.01933 (2016): n. pag.
\bibitem{87_resnet}He, Kaiming et al. “Deep Residual Learning for Image Recognition.” 2016 IEEE Conference on Computer Vision and Pattern Recognition (CVPR) (2015): 770-778.

\bibitem{88_AIGC_survey}Zhang, Chaoning et al. “A Complete Survey on Generative AI (AIGC): Is ChatGPT from GPT-4 to GPT-5 All You Need?” ArXiv abs/2303.11717 (2023): n. pag.

\bibitem{89_RNN}Hidasi, Balázs et al. “Recurrent Neural Networks.” Computer Science Today (2021).

\bibitem{90_RNN}Yin, Bojian et al. “Accurate and efficient time-domain classification with adaptive spiking recurrent neural networks.” Nature Machine Intelligence 3 (2021): 905 - 913.

\bibitem{91_RNN}Erichson, N. Benjamin et al. “Lipschitz Recurrent Neural Networks.” International Conference on Learning Representations: n. pag.

\bibitem{92_RNN}J. Ainslie et al. "ETC: Encoding Long and Structured Data in Transformers." ArXiv, abs/2004.08483 (2020).

\bibitem{93_transformer}Jianqiao Zheng et al. "Rethinking Positional Encoding." ArXiv, abs/2107.02561 (2021).

\bibitem{94_Chatgpt}Ouyang, Long et al. “Training language models to follow instructions with human feedback.” Neural Information Processing Systems (2022): n. pag.

\bibitem{95_trilemma}Xiao, Zhisheng et al. “Tackling the Generative Learning Trilemma with Denoising Diffusion GANs.” International Conference on Learning Representation (2021): n. pag.

\bibitem{96_gan_fail}Salimans, Tim et al. “Improved Techniques for Training GANs.” NIPS (2016): n. pag.

\bibitem{97_gan_fail}Zhao, Shengjia et al. “Bias and Generalization in Deep Generative Models: An Empirical Study.” Neural Information Processing Systems (2018).

\bibitem{98_diffusion_good}Song, Yang et al. “Maximum Likelihood Training of Score-Based Diffusion Models.” Neural Information Processing Systems (2021).

\bibitem{99_diffusion_good}Huang, Chin-Wei et al. “A Variational Perspective on Diffusion-Based Generative Models and Score Matching.” ArXiv abs/2106.02808 (2021): n. pag.

\bibitem{100_diffusion_good}Ho, Jonathan et al. “Cascaded Diffusion Models for High Fidelity Image Generation.” J. Mach. Learn. Res. 23 (2021): 47:1-47:33.

\bibitem{101_diffusion_good}Kingma, Diederik P. et al. “Variational Diffusion Models.” ArXiv abs/2107.00630 (2021): n. pag.

\bibitem{102_gpt_good}Marcel Binz et al. "Using cognitive psychology to understand GPT-3." Proceedings of the National Academy of Sciences of the United States of America, 120 6 (2022): e2218523120 . https://doi.org/10.1073/pnas.2218523120.

\bibitem{103_direct_use} Yunhao Chen, Zihui Yan, Yunjie Zhu, Zhen Ren, Jianlu Shen, Yifan Huang: Data Augmentation for Environmental Sound Classification Using Diffusion Probabilistic Model with Top-K Selection Discriminator. ICIC (2) 2023: 283-295

\bibitem{104_direct_use}Chen, Dong et al. “Deep Data Augmentation for Weed Recognition Enhancement: A Diffusion Probabilistic Model and Transfer Learning Based Approach.” ArXiv abs/2210.09509 (2022): n. pag.

\bibitem{105_direct_use}Bahmei, Behnaz et al. “CNN-RNN and Data Augmentation Using Deep Convolutional Generative Adversarial Network for Environmental Sound Classification.” IEEE Signal Processing Letters 29 (2022): 682-686.

\bibitem{106_direct_use_medical}Esteban, Cristóbal et al. “Real-valued (Medical) Time Series Generation with Recurrent Conditional GANs.” ArXiv abs/1706.02633 (2017): n. pag.

\bibitem{107_direct_use_medical}Zhang, Huijuan et al. “Medical Image Synthetic Data Augmentation Using GAN.” Proceedings of the 4th International Conference on Computer Science and Application Engineering (2020): n. pag.

\bibitem{107_direct_use1}Subedi, Bharat et al. “Feature Learning-Based Generative Adversarial Network Data Augmentation for Class-Based Few-Shot Learning.” Mathematical Problems in Engineering (2022): n. pag.

\bibitem{107_direct_use2}Zhang, Xiaofeng et al. “Deep Adversarial Data Augmentation for Extremely Low Data Regimes.” IEEE Transactions on Circuits and Systems for Video Technology 31 (2021): 15-28.

\bibitem{108_direct_use_QA}Shakeri, Siamak et al. “Towards Zero-Shot Multilingual Synthetic Question and Answer Generation for Cross-Lingual Reading Comprehension.” International Conference on Natural Language Generation (2020).
 

\bibitem{109_direct_use_few_shot}Aleksandra Edwards, Asahi Ushio, Jose Camacho-collados, Helene Ribaupierre, and Alun Preece. 2022. Guiding Generative Language Models for Data Augmentation in Few-Shot Text Classification. In Proceedings of the Fourth Workshop on Data Science with Human-in-the-Loop (Language Advances), pages 51–63, Abu Dhabi, United Arab Emirates (Hybrid). Association for Computational Linguistics.

\bibitem{110_prompt_structure}Akrout, Mohamed et al. “Diffusion-based Data Augmentation for Skin Disease Classification: Impact Across Original Medical Datasets to Fully Synthetic Images.” ArXiv abs/2301.04802 (2023): n. pag.


\bibitem{111_prompt_structure}Sagers, Luke et al. “Improving dermatology classifiers across populations using images generated by large diffusion models.” ArXiv abs/2211.13352 (2022): n. pag.

\bibitem{112_prompt_structure}Li, Y. and Jiawei Yuan. “Generative Data Augmentation with Contrastive Learning for Zero-Shot Stance Detection.” Conference on Empirical Methods in Natural Language Processing (2022)


\bibitem{113_prompt_structure}Mekala, Dheeraj et al. “Leveraging QA Datasets to Improve Generative Data Augmentation.” Conference on Empirical Methods in Natural Language Processing (2022).

\bibitem{114_prompt_structure}Vu, Tu et al. “STraTA: Self-Training with Task Augmentation for Better Few-shot Learning.” Conference on Empirical Methods in Natural Language Processing (2021).

\bibitem{115_prompt_structure}Yang, Cai et al. “BLIAM: Literature-based Data Synthesis for Synergistic Drug Combination Prediction.” ArXiv abs/2302.06860 (2023): n. pag.

\bibitem{116_prompt_structure}Chia, Yew Ken et al. “RelationPrompt: Leveraging Prompts to Generate Synthetic Data for Zero-Shot Relation Triplet Extraction.” ACL2022 Findings

\bibitem{117_prompt_structure}Lin, Shaobo et al. “Explore the Power of Synthetic Data on Few-shot Object Detection.” 2023 IEEE/CVF Conference on Computer Vision and Pattern Recognition Workshops (CVPRW)

\bibitem{118_prompt_structure}Benigmim, Yasser et al. “One-shot Unsupervised Domain Adaptation with Personalized Diffusion Models.” 2023 IEEE/CVF Conference on Computer Vision and Pattern Recognition Workshops (CVPRW) (2023): 698-708.

\bibitem{119_prompt_structure}Shipard, Jordan, et al. "Diversity Is Definitely Needed: Improving Model-Agnostic Zero-shot Classification via Stable Diffusion." ArXiv, 2023, /abs/2302.03298.

\bibitem{120_pubbert}Gu, Yu, et al. "Domain-Specific Language Model Pretraining for Biomedical Natural Language Processing." ArXiv, 2020, https://doi.org/10.1145/3458754.

\bibitem{121_chatgpt_prompt}Yu, Qifan et al. “Interactive Data Synthesis for Systematic Vision Adaptation via LLMs-AIGCs Collaboration.” ArXiv abs/2305.12799 (2023): n. pag.

\bibitem{122_prompt_structure}Bansal, Hritik and Aditya Grover. “Leaving Reality to Imagination: Robust Classification via Generated Datasets.” ArXiv abs/2302.02503 (2023): n. pag.

\bibitem{123_llm}Chowdhery, Aakanksha et al. “PaLM: Scaling Language Modeling with Pathways.” ArXiv abs/2204.02311 (2022): n. pag.

\bibitem{124_llm}Touvron, Hugo et al. “LLaMA: Open and Efficient Foundation Language Models.” ArXiv abs/2302.13971 (2023): n. pag.

\bibitem{126_chatgpt_prompt}Bubeck, Sébastien et al. “Sparks of Artificial General Intelligence: Early experiments with GPT-4.” ArXiv abs/2303.12712 (2023): n. pag.

\bibitem{127_chatgpt_prompt}Touvron, Hugo et al. “LLaMA: Open and Efficient Foundation Language Models.” ArXiv abs/2302.13971 (2023): n. pag.

\bibitem{128_chatgpt_prompt}Kumar, Ananya et al. “Fine-Tuning can Distort Pretrained Features and Underperform Out-of-Distribution.” ArXiv abs/2202.10054 (2022): n. pag.

\bibitem{129_AIGC}Ramesh, Aditya et al. “Hierarchical Text-Conditional Image Generation with CLIP Latents.” ArXiv abs/2204.06125 (2022): n. pag.

\bibitem{131_AIGC}Robin Rombach, Andreas Blattmann, Dominik Lorenz,
Patrick Esser, and Björn Ommer. High-resolution image
synthesis with latent diffusion models. In Proceedings of
the IEEE/CVF Conference on Computer Vision and Pattern
Recognition, pages 10684–10695, 2022.

\bibitem{132_AIGC}Chitwan Saharia, William Chan, Saurabh Saxena, Lala
Li, Jay Whang, Emily L Denton, Kamyar Ghasemipour, Raphael Gontijo Lopes, Burcu Karagol Ayan, Tim Salimans, et al. Photorealistic text-to-image diffusion models with deep language understanding. Advances in Neural Information Processing Systems, 35:36479–36494, 2022.

\bibitem{133_stable_diffusion}Rombach, Robin et al. “High-Resolution Image Synthesis with Latent Diffusion Models.” 2022 IEEE/CVF Conference on Computer Vision and Pattern Recognition (CVPR) (2021): 10674-10685.

\bibitem{134}Whitehouse, Chenxi et al. “LLM-powered Data Augmentation for Enhanced Crosslingual Performance.” ArXiv abs/2305.14288 (2023): n. pag.

\bibitem{135_llm_generated_prompt}Zhou, Yongchao et al. “Large Language Models Are Human-Level Prompt Engineers.” International Conference on Learning Representation  (2023): n. pag.

\bibitem{136_latent_code}Zhou, Yongchao et al. “Training on Thin Air: Improve Image Classification with Generated Data.” ArXiv abs/2305.15316 (2023): n. pag.

\bibitem{137_latent_code}Zhou, Yongchao et al. “Training on Thin Air: Improve Image Classification with Generated Data.” ArXiv abs/2305.15316 (2023): n. pag.

\bibitem{138_inversion}Weihao Xia, Yulun Zhang, Yujiu Yang, Jing-Hao Xue, Bolei Zhou, and Ming-Hsuan Yang. Gan
inversion: A survey. IEEE Transactions on Pattern Analysis and Machine Intelligence, 2022.

\bibitem{139_inversion}Choi, Jooyoung et al. “ILVR: Conditioning Method for Denoising Diffusion Probabilistic Models.” 2021 IEEE/CVF International Conference on Computer Vision (ICCV) (2021): 14347-14356.

\bibitem{140_Inversion}Gal, Rinon et al. “An Image is Worth One Word: Personalizing Text-to-Image Generation using Textual Inversion.” ArXiv abs/2208.01618 (2022): n. pag.

\bibitem{141_Inversion}Kumari, Nupur et al. “Multi-Concept Customization of Text-to-Image Diffusion.” 2023 IEEE/CVF Conference on Computer Vision and Pattern Recognition (CVPR) (2022): 1931-1941.

\bibitem{142_Inversion}Morgan Klaus Scheuerman, Alex Hanna, and Emily Denton. Do datasets have politics?
disciplinary values in computer vision dataset development. Proceedings of the ACM on
Human-Computer Interaction, 5(CSCW2):1–37, 2021

\bibitem{143_representation}Jahanian, Ali et al. “Generative Models as a Data Source for Multiview Representation Learning.” International Conference on Learning Representation (2022): n. pag. 

\bibitem{144_representation}Jahanian, Ali et al. “On the "steerability" of generative adversarial networks.” International Conference on Learning Representation  (2020): n. pag.

\bibitem{145_representation}Härkönen, Erik et al. “GANSpace: Discovering Interpretable GAN Controls.” Neural Information Processing Systems (2021): n. pag.

\bibitem{146_representation}He, Kaiming et al. “Momentum Contrast for Unsupervised Visual Representation Learning.” 2020 IEEE/CVF Conference on Computer Vision and Pattern Recognition (CVPR) (2019): 9726-9735.

\bibitem{147_representation}Chen, Ting et al. “A Simple Framework for Contrastive Learning of Visual Representations.” International Conference on Machine Learning
 (2021): n. pag.

\bibitem{148_representation}Tian, Yonglong et al. “What makes for good views for contrastive learning.” Neural Information Processing Systems (2021): n. pag.

\bibitem{149_representation}He, Kaiming et al. “Momentum Contrast for Unsupervised Visual Representation Learning.” 2020 IEEE/CVF Conference on Computer Vision and Pattern Recognition (CVPR) (2019): 9726-9735.

\bibitem{150_T5}Raffel, Colin et al. “Exploring the Limits of Transfer Learning with a Unified Text-to-Text Transformer.” Journal of machine learning research(202): n. pag.

\bibitem{151_synthetic_ready}He, Ruifei et al. “Is synthetic data from generative models ready for image recognition?” International Conference on Learning Representation (2023): n. pag.

\bibitem{152_data_selection}Mo, Sangwoo et al. “Mining GOLD Samples for Conditional GANs.” Neural Information Processing Systems (2019).

\bibitem{153_data_selection}Lee, Jin Ha et al. “Self-Diagnosing GAN: Diagnosing Underrepresented Samples in Generative Adversarial Networks.” Neural Information Processing Systems (2021): n. pag.

\bibitem{154_realism} Kynkäänniemi, Tuomas et al. “Improved Precision and Recall Metric for Assessing Generative Models.” Neural Information Processing Systems (2020): n. pag.

\bibitem{155_top_selection}Vu, Tu et al. “STraTA: Self-Training with Task Augmentation for Better Few-shot Learning.” Conference on Empirical Methods in Natural Language Processing (2021).

\bibitem{156_CLIP}Radford, Alec et al. “Learning Transferable Visual Models From Natural Language Supervision.” International Conference on Machine Learning (2021).

\bibitem{157_theory}Bousquet, Olivier and André Elisseeff. “Stability and Generalization.” J. Mach. Learn. Res. 2 (2002): 499-526.

\bibitem{158_theory}Shalev-Shwartz, Shai et al. “Learnability, Stability and Uniform Convergence.” J. Mach. Learn. Res. 11 (2010): 2635-2670.

\bibitem{159_theory}Kuzborskij, Ilja and Christoph H. Lampert. “Data-Dependent Stability of Stochastic Gradient Descent.” International Conference on Machine Learning (2017).

\bibitem{160_theory}Liu, Tongliang et al. “Algorithmic Stability and Hypothesis Complexity.” International Conference on Machine Learning (2018): n. pag.

\bibitem{161_theory}Hardt, Moritz et al. “Train faster, generalize better: Stability of stochastic gradient descent.”  International Conference on Machine Learning (2016): n. pag.

\bibitem{162_theory}Zhang, Yikai et al. “Stability of SGD: Tightness Analysis and Improved Bounds.” Conference on Uncertainty in Artificial Intelligence (2021).

\bibitem{163_theory}Xing, Yue et al. “On the Algorithmic Stability of Adversarial Training.” Neural Information Processing Systems (2021).

\bibitem{164_theory}Feldman, Vitaly and Jan Vondrák. “Generalization Bounds for Uniformly Stable Algorithms.” Neural Information Processing Systems (2018).

\bibitem{165_theory}Feldman, Vitaly and Jan Vondrák. “High probability generalization bounds for uniformly stable algorithms with nearly optimal rate.” Annual Conference Computational Learning Theory (2019).

\bibitem{166_theory}Bousquet, Olivier et al. “Sharper bounds for uniformly stable algorithms.” Annual Conference Computational Learning Theory (2020): n. pag.

\bibitem{167_theory}Mohri, Mehryar and Afshin Rostamizadeh. “Stability Bounds for Non-i.i.d. Processes.” NIPS (2007).

\bibitem{168_theory}Mohri, Mehryar and Afshin Rostamizadeh. “Stability Bounds for Stationary phi-mixing and beta-mixing Processes.” J. Mach. Learn. Res. 11 (2008): 789-814.

\bibitem{169_theory}Zhang, Rui et al. “McDiarmid-Type Inequalities for Graph-Dependent Variables and Stability Bounds.” Neural Information Processing Systems (2019).

\bibitem{170_theory}Farnia, Farzan and Asuman E. Ozdaglar. “Train simultaneously, generalize better: Stability of gradient-based minimax learners.” International Conference on Machine Learning (2020).

\bibitem{171_theory}Zheng, Chenyu, et al. "Toward Understanding Generative Data Augmentation." ArXiv, /abs/2305.17476. (2023) .n.pag

\bibitem{172_FID}Heusel, Martin et al. “GANs Trained by a Two Time-Scale Update Rule Converge to a Local Nash Equilibrium.” NIPS (2017).

\bibitem{173_segmentation}Zhang, Yuxuan et al. “DatasetGAN: Efficient Labeled Data Factory with Minimal Human Effort.” 2021 IEEE/CVF Conference on Computer Vision and Pattern Recognition (CVPR) (2021): 10140-10150.

\bibitem{174_resnet}He, Kaiming et al. “Deep Residual Learning for Image Recognition.” 2016 IEEE Conference on Computer Vision and Pattern Recognition (CVPR) (2015): 770-778.

\bibitem{175_imagenet}Deng, Jia et al. “ImageNet: A large-scale hierarchical image database.” 2009 IEEE Conference on Computer Vision and Pattern Recognition (2009): 248-255.

\bibitem{176_tsn}Maaten, Laurens van der and Geoffrey E. Hinton. “Visualizing Data using t-SNE.” Journal of Machine Learning Research 9 (2008): 2579-2605.

\bibitem{apl1}Waheed A, Goyal M, Gupta D, et al. Covidgan: data augmentation using auxiliary classifier gan for improved covid-19 detection[J]. Ieee Access, 2020, 8: 91916-91923.
\bibitem{apl2}Sun L, Wang J, Huang Y, et al. An adversarial learning approach to medical image synthesis for lesion detection[J]. IEEE journal of biomedical and health informatics, 2020, 24(8): 2303-2314.
\bibitem{apl3}Wang Q, Zhang X, Chen W, et al. Class-aware multi-window adversarial lung nodule synthesis conditioned on semantic features[C]//Medical Image Computing and Computer Assisted Intervention–MICCAI 2020: 23rd International Conference, Lima, Peru, October 4–8, 2020, Proceedings, Part VI 23. Springer International Publishing, 2020: 589-598.
\bibitem{apl4}Geng X, Yao Q, Jiang K, et al. Deep neural generative adversarial model based on VAE+ GAN for disorder diagnosis[C]//2020 International Conference on Internet of Things and Intelligent Applications (ITIA). IEEE, 2020: 1-7.

\bibitem{apl5}Pang T, Wong J H D, Ng W L, et al. Semi-supervised GAN-based radiomics model for data augmentation in breast ultrasound mass classification[J]. Computer Methods and Programs in Biomedicine, 2021, 203: 106018.

\bibitem{apl6}Barile B, Marzullo A, Stamile C, et al. Data augmentation using generative adversarial neural networks on brain structural connectivity in multiple sclerosis[J]. Computer methods and programs in biomedicine, 2021, 206: 106113.

\bibitem{apl7}Shen T, Hao K, Gou C, et al. Mass image synthesis in mammogram with contextual information based on GANs[J]. Computer Methods and Programs in Biomedicine, 2021, 202: 106019.

\bibitem{apl8}Ambita A A E, Boquio E N V, Naval Jr P C. Covit-gan: vision transformer forcovid-19 detection in ct scan imageswith self-attention gan forDataAugmentation[C]//International Conference on Artificial Neural Networks. Cham: Springer International Publishing, 2021: 587-598.

\bibitem{apl9}Hirte A U, Platscher M, Joyce T, et al. Realistic generation of diffusion-weighted magnetic resonance brain images with deep generative models[J]. Magnetic Resonance Imaging, 2021, 81: 60-66.

\bibitem{apl10}Kaur S, Aggarwal H, Rani R. MR image synthesis using generative adversarial networks for Parkinson’s disease classification[C]//Proceedings of International Conference on Artificial Intelligence and Applications: ICAIA 2020. Springer Singapore, 2021: 317-327.

\bibitem{apl11}Guan Q, Chen Y, Wei Z, et al. Medical image augmentation for lesion detection using a texture-constrained multichannel progressive GAN[J]. Computers in Biology and Medicine, 2022, 145: 105444.

\bibitem{apl12}Ahmad B, Sun J, You Q, et al. Brain tumor classification using a combination of variational autoencoders and generative adversarial networks[J]. Biomedicines, 2022, 10(2): 223.

\bibitem{apl13}Pombo G, Gray R, Cardoso M J, et al. Equitable modelling of brain imaging by counterfactual augmentation with morphologically constrained 3d deep generative models[J]. Medical Image Analysis, 2023, 84: 102723.
\bibitem{apl14}Jiang Y, Chen H, Loew M, et al. COVID-19 CT image synthesis with a conditional generative adversarial network[J]. IEEE Journal of Biomedical and Health Informatics, 2020, 25(2): 441-452.
\bibitem{apl15}Qasim, A.B.; Ezhov, I.; Shit, S.; Schoppe, O.; Paetzold, J.C.; Sekuboyina, A.; Kofler, F.; Lipkova, J.; Li, H.; Menze, B. Red-GAN: Attacking class imbalance via conditioned generation. Yet another medical imaging perspective. In Proceedings of the Medical Imaging with Deep Learning, Montreal, QC, Canada, 6–9 July 2020; pp. 655–668.
\bibitem{apl16}Platscher, M.; Zopes, J.; Federau, C. Image Translation for Medical Image Generation–Ischemic Stroke Lesions. arXiv 2020, arXiv:2010.02745.

\bibitem{apl17}Shi, H.; Lu, J.; Zhou, Q. A novel data augmentation method using style-based GAN for robust pulmonary nodule segmentation. In Proceedings of the 2020 Chinese Control and Decision Conference (CCDC), Hefei, China, 22–24 August 2020; pp. 2486–2491.

\bibitem{apl18}Shen, Z.; Ouyang, X.; Xiao, B.; Cheng, J.Z.; Shen, D.; Wang, Q. Image synthesis with disentangled attributes for chest X-ray nodule augmentation and detection. Med. Image Anal. 2022, 102708.

\bibitem{apl19}Yurt, M.; Dar, S.U.; Erdem, A.; Erdem, E.; Oguz, K.K.; Çukur, T. mustGAN: Multi-stream generative adversarial networks for MR image synthesis. Med. Image Anal. 2021, 70, 101944.

\bibitem{apl20}Yang, H.; Lu, X.; Wang, S.H.; Lu, Z.; Yao, J.; Jiang, Y.; Qian, P. Synthesizing multi-contrast MR images via novel 3D conditional Variational auto-encoding GAN. Mob. Netw. Appl. 2021, 26, 415–424.

\bibitem{apl21}Sikka A, Virk J S, Bathula D R. MRI to PET Cross-Modality Translation using Globally and Locally Aware GAN (GLA-GAN) for Multi-Modal Diagnosis of Alzheimer's Disease[J]. arXiv preprint arXiv:2108.02160, 2021.

\bibitem{apl22}Amirrajab, S.; Lorenz, C.; Weese, J.; Pluim, J.; Breeuwer, M. Pathology Synthesis of 3D Consistent Cardiac MR Images Using 2D VAEs and GANs. In Proceedings of the International Workshop on Simulation and Synthesis in Medical Imaging, Singapore, 18 September 2022; pp. 34–42.
\bibitem{apl23}Chen S, Gao D, Wang L, Zhang Y. Cervical Cancer Single Cell Image Data Augmentation Using Residual Condition Generative Adversarial Networks. In: 2020 3rd International Conference on Artificial Intelligence and Big Data (ICAIBD); 2020. p. 237–241.
\bibitem{apl24}Almezhghwi K, Serte S. Improved classification of white blood cells with the generative adversarial network and deep convolutional neural network[J]. Computational Intelligence and Neuroscience, 2020, 2020.

\bibitem{apl25} Teramoto A, Tsukamoto T, Yamada A, et al. Deep learning approach to classification of lung cytological images: Two-step training using actual and synthesized images by progressive growing of generative adversarial networks[J]. PloS one, 2020, 15(3): e0229951.

\bibitem{apl26} Murali L K, Lutnick B, Ginley B, et al. Generative modeling for renal microanatomy[C]//Medical Imaging 2020: Digital Pathology. SPIE, 2020, 11320: 99-108.

\bibitem{apl27} Quiros A C, Murray-Smith R, Yuan K. Learning a low dimensional manifold of real cancer tissue with PathologyGAN[J]. arXiv preprint arXiv:2004.06517, 2020.

\bibitem{apl28}  Mirzazadeh A, Mohseni A, Ibrahim S, Giuste FO, Zhu Y, Shehata BM, et al. Improving Heart Transplant Rejection Classification Training using Progressive Generative Adversarial Networks. In: 2021 IEEE EMBS International Conference on Biomedical and Health Informatics (BHI); 2021. p. 1–4.
\bibitem{apl29} Zhao M, Jin L, Teng S, Li Z. Attention Residual Network for White Blood Cell Classification with WGAN Data Augmentation. In: 2021 11th International Conference on Information Technology in Medicine and Education (ITME); 2021. p. 336–340.

\bibitem{apl30} Yu S, Zhang S, Wang B, et al. Generative adversarial network based data augmentation to improve cervical cell classification model[J]. Math. Biosci. Eng, 2021, 18: 1740-1752.

\bibitem{apl31} Liu K, Shuai R, Ma L. Cells image generation method based on VAE-SGAN[J]. Procedia Computer Science, 2021, 183: 589-595.

\bibitem{apl32} Pandya D, Patel T, Singh Dk. White Blood Cell Image Generation using Deep Convolutional Generative Adversarial Network. In: 2022 International Conference on Augmented Intelligence and Sustainable Systems (ICAISS); 2022. p. 129–134.

\bibitem{apl33} Kunzmann S, Öttl M, Madhu P, et al. An unobtrusive quality supervision approach for medical image annotation[J]. arXiv preprint arXiv:2211.06146, 2022.

\bibitem{apl34}. Dee W, Alaaeldin Ibrahim R, Marouli E. Histopathological Domain Adaptation with Generative Adversarial Networks-Bridging the Domain Gap Between Thyroid Cancer Histopathology Datasets[J]. bioRxiv, 2023: 2023.05. 22.541691.

\bibitem{apl35} Lu Y, Chen D, Olaniyi E, et al. Generative adversarial networks (GANs) for image augmentation in agriculture: A systematic review[J]. Computers and Electronics in Agriculture, 2022, 200: 107208.

\bibitem{apl36} Divyanth L G, Guru D S, Soni P, et al. Image-to-image translation-based data augmentation for improving crop/weed classification models for precision agriculture applications[J]. Algorithms, 2022, 15(11): 401.

\bibitem{apl37} Ufuah D, Thomas G, Balocco S, et al. A Data Augmentation Approach Based on Generative Adversarial Networks for Date Fruit Classification[J]. Applied Engineering in Agriculture, 2022, 38(6): 975-982.

\bibitem{apl38} Bhakta I, Phadikar S, Majumder K. Thermal image augmentation with generative adversarial network for agricultural disease prediction[C]//International conference on computational intelligence in pattern recognition. Singapore: Springer Nature Singapore, 2022: 345-354.

\bibitem{apl39} Sharma V, Tripathi A K, Mittal H, et al. Weedgan: a novel generative adversarial network for cotton weed identification[J]. The Visual Computer, 2022: 1-17.

\bibitem{apl40} Zhang Z, Gao Q, Liu L, et al. A High-Quality Rice Leaf Disease Image Data Augmentation Method Based on a Dual GAN[J]. IEEE Access, 2023, 11: 21176-21191.

\bibitem{apl41} Deshpande R, Patidar H. Detection of Plant Leaf Disease by Generative Adversarial and Deep Convolutional Neural Network[J]. Journal of The Institution of Engineers (India): Series B, 2023: 1-10.

\bibitem{apl42} Kulkarni S, Deepa Shenoy P, Venugopal K R. CoffeeGAN: An Effective Data Augmentation Model for Coffee Plant Diseases[C]//International Conference on Data Management, Analytics \& Innovation. Singapore: Springer Nature Singapore, 2023: 431-443.

\bibitem{apl43} Sharma A, Kaur H, Prashar D. Generative Adversarial Networks Based Approach for Data Augmentation in Mango Leaf Disease Detection System[C]//2023 IEEE 12th International Conference on Communication Systems and Network Technologies (CSNT). IEEE, 2023: 816-821.

\bibitem{apl44} Alshammari K, Alshammari R, Alshammari A, et al. An Improved Pear Disease Classification Approach using Cycle Generative Adversarial Network[J]. 2023.

\bibitem{apl45} Jamadar R A, Sharma A, Wagh K. Combining Generative Adversarial Networks with Transfer Learning for Deep Learning-Based Pomegranate Plant Leaf Disease Detection[C]//International Conference on Smart Computing and Communication. Singapore: Springer Nature Singapore, 2023: 487-496.

\bibitem{apl46} Haruna Y, Qin S, Mbyamm Kiki M J. An improved approach to detection of rice leaf disease with gan-based data augmentation pipeline[J]. Applied Sciences, 2023, 13(3): 1346.

\bibitem{apl47} Giselsson T M, Jørgensen R N, Jensen P K, et al. A public image database for benchmark of plant seedling classification algorithms[J]. arXiv preprint arXiv:1711.05458, 2017.

\bibitem{apl48} Sethy P K, Barpanda N K, Rath A K, et al. Deep feature based rice leaf disease identification using support vector machine[J]. Computers and Electronics in Agriculture, 2020, 175: 105527.

\bibitem{apl49} Xu H. Plantvillage disease classification challenge-color images[J]. 2018.
\bibitem{apl50} Fenu G, Malloci F M. DiaMOS plant: A dataset for diagnosis and monitoring plant disease[J]. Agronomy, 2021, 11(11): 2107.
\bibitem{apl51} Tai C Y, Wang W J, Huang Y M. Using Time-Series Generative Adversarial Networks to Synthesize Sensing Data for Pest Incidence Forecasting on Sustainable Agriculture[J]. Sustainability, 2023, 15(10): 7834.
\bibitem{apl52} Gutiérrez S, Tardaguila J. Evolutionary conditional GANs for supervised data augmentation: The case of assessing berry number per cluster in grapevine[J]. Applied Soft Computing, 2023: 110805.

\bibitem{apl53} Taylor J A, Tisseyre B, Leroux C. A simple index to determine if within-field spatial production variation exhibits potential management effects: Application in vineyards using yield monitor data[J]. Precision Agriculture, 2019, 20: 880-895.

\bibitem{apl54} Tardaguila J, Stoll M, Gutiérrez S, et al. Smart applications and digital technologies in viticulture: A review[J]. Smart Agricultural Technology, 2021, 1: 100005.

\bibitem{apl55} Trabucco B, Doherty K, Gurinas M, et al. Effective data augmentation with diffusion models[J]. arXiv preprint arXiv:2302.07944, 2023.
\bibitem{apl56} Chen D, Qi X, Zheng Y, et al. Deep data augmentation for weed recognition enhancement: A diffusion probabilistic model and transfer learning based approach[C]//2023 ASABE Annual International Meeting. American Society of Agricultural and Biological Engineers, 2023: 1.
\bibitem{apl57} Moreno H, Gómez A, Altares-López S, et al. Analysis of Stable Diffusion-Derived Fake Weeds Performance for Training Convolutional Neural Networks[J]. Available at SSRN 4535844.
\bibitem{apl58} Chen D, Qi X, Zheng Y, et al. Synthetic Data Augmentation by Diffusion Probabilistic Models to Enhance Weed Recognition[J]. Available at SSRN 4414966.

\bibitem{apl59} Giakoumoglou N, Pechlivani E M, Tzovaras D. Generate-Paste-Blend-Detect: Synthetic dataset for object detection in the agriculture domain[J]. Smart Agricultural Technology, 2023, 5: 100258.

\bibitem{apl60} Akrout M, Gyepesi B, Holló P, et al. Diffusion-based Data Augmentation for Skin Disease Classification: Impact Across Original Medical Datasets to Fully Synthetic Images[J]. arXiv preprint arXiv:2301.04802, 2023.
\bibitem{apl61} Sagers L W, Diao J A, Groh M, et al. Improving dermatology classifiers across populations using images generated by large diffusion models[J]. arXiv preprint arXiv:2211.13352, 2022.
\bibitem{apl62} de Wilde B, Saha A, ten Broek R P G, et al. Medical diffusion on a budget: textual inversion for medical image generation[J]. arXiv preprint arXiv:2303.13430, 2023.
\bibitem{apl63} Pinaya W H L, Tudosiu P D, Dafflon J, et al. Brain imaging generation with latent diffusion models[C]//MICCAI Workshop on Deep Generative Models. Cham: Springer Nature Switzerland, 2022: 117-126.
\bibitem{apl64} Walter H. L. Pinaya and Petru-Daniel Tudosiu et al. LDM 100k Dataset
\bibitem{apl65} Moghadam P A, Van Dalen S, Martin K C, et al. A morphology focused diffusion probabilistic model for synthesis of histopathology images[C]//Proceedings of the IEEE/CVF Winter Conference on Applications of Computer Vision. 2023: 2000-2009.
\bibitem{apl66} Kim B, Ye J C. Diffusion deformable model for 4D temporal medical image generation[C]//International Conference on Medical Image Computing and Computer-Assisted Intervention. Cham: Springer Nature Switzerland, 2022: 539-548.

\bibitem{apl67} Waibel D J E, Röell E, Rieck B, et al. A diffusion model predicts 3d shapes from 2d microscopy images[C]//2023 IEEE 20th International Symposium on Biomedical Imaging (ISBI). IEEE, 2023: 1-5.

\bibitem{apl68} Packhäuser K, Folle L, Thamm F, et al. Generation of anonymous chest radiographs using latent diffusion models for training thoracic abnormality classification systems[C]//2023 IEEE 20th International Symposium on Biomedical Imaging (ISBI). IEEE, 2023: 1-5.

\bibitem{apl69} Jiang L, Mao Y, Chen X, et al. CoLa-Diff: Conditional Latent Diffusion Model for Multi-Modal MRI Synthesis[J]. arXiv preprint arXiv:2303.14081, 2023.
\bibitem{apl70} Abayomi-Alli O O, Damaševičius R, Qazi A, et al. Data augmentation and deep learning methods in sound classification: A systematic review[J]. Electronics, 2022, 11(22): 3795.

\bibitem{apl71} Bakır H, Çayır A N, Navruz T S. A comprehensive experimental study for analyzing the effects of data augmentation techniques on voice classification[J]. Multimedia Tools and Applications, 2023: 1-28.
\bibitem{apl72} Chen Y, Yan Z, Zhu Y, et al. Data Augmentation for Environmental Sound Classification Using Diffusion Probabilistic Model with Top-K Selection Discriminator[C]//International Conference on Intelligent Computing. Singapore: Springer Nature Singapore, 2023: 283-295.

\bibitem{apl73} Pascual S, Bhattacharya G, Yeh C, et al. Full-band general audio synthesis with score-based diffusion[C]//ICASSP 2023-2023 IEEE International Conference on Acoustics, Speech and Signal Processing (ICASSP). IEEE, 2023: 1-5.


\bibitem{apl74} Margaryan M, Seibold M, Joshi I, et al. Improved Techniques for the Conditional Generative Augmentation of Clinical Audio Data[J]. arXiv preprint arXiv:2211.02874, 2022.

\bibitem{apl75} Jang S, Kim Y. Dual ResNet-based Environmental Sound Classification using GAN[C]//2023 17th International Conference on Ubiquitous Information Management and Communication (IMCOM). IEEE, 2023: 1-6.
\bibitem{apl76} Kim E, Moon J, Shim J, et al. DualDiscWaveGAN-Based Data Augmentation Scheme for Animal Sound Classification[J]. Sensors, 2023, 23(4): 2024.
\bibitem{apl77} Saldanha J, Chakraborty S, Patil S, et al. Data augmentation using Variational Autoencoders for improvement of respiratory disease classification[J]. Plos one, 2022, 17(8): e0266467.
\bibitem{apl78} Takezaki S, Kishida K. Data Augmentation and the Improvement of the Performance of Convolutional Neural Networks for Heart Sound Classification[J]. IAENG Int. J. Comput. Sci, 2022, 49(4).
\bibitem{apl79} Madhu A, K S. EnvGAN: A GAN-based augmentation to improve environmental sound classification[J]. Artificial Intelligence Review, 2022, 55(8): 6301-6320.
\bibitem{apl80} Dorjsembe Z, Odonchimed S, Xiao F. Three-dimensional medical image synthesis with denoising diffusion probabilistic models[C]//Medical Imaging with Deep Learning. 2022.

\end{thebibliography}


\section{Acknowledgement}
Declaration of generative AI and AI-assisted technologies in the writing process

During the preparation of this work, the author(s) used chatgpt in order to polish up the writing. After using this tool/service, the author(s) reviewed and edited the content as needed and take(s) full responsibility for the content of the publication.

\end{document}